\documentclass{article}
\usepackage{iclr2027_conference,times}

\iclrfinalcopy

\usepackage[utf8]{inputenc} 
\usepackage[T1]{fontenc}    
\usepackage{hyperref}       
\usepackage{xurl}            
\usepackage{longtable}
\usepackage{booktabs}       
\usepackage{amsfonts}       
\usepackage{nicefrac}       
\usepackage{microtype}      
\usepackage{xcolor}         

\usepackage{graphicx}
\usepackage[most]{tcolorbox}
\usepackage{enumitem}
\usepackage{float}
\usepackage{amsmath}
\usepackage{amssymb}
\usepackage{mathtools}
\usepackage{amsthm}
\usepackage{subcaption}
\usepackage{caption}
\usepackage{listings}
\usepackage[capitalize,noabbrev]{cleveref}
\usepackage{wrapfig}
\hypersetup{hidelinks}

\newtcolorbox{summarybox}{
  colframe=black!70,   
  boxrule=1.5pt,         
  arc=4pt,                   
  left=2pt,
  right=2pt,
  top=3pt,
  bottom=3pt,
  halign=center,
}
\title{Same Error, Different Function: The Optimizer as an Implicit Prior in Financial Time Series}

\author{%
  Federico Vittorio Cortesi\thanks{Equal contribution.} \\
  Massachusetts Institute of Technology\\
  Cambridge, MA, USA \\
  \texttt{corte911@mit.edu} \\
  \And
  Giuseppe Iannone\footnotemark[1] \\
  Massachusetts Institute of Technology\\
  Cambridge, MA, USA \\
  \And
  Giulia Crippa \\
  Princeton University\\
  Princeton, NJ, USA \\
  \And
  Tomaso Poggio \\
  Massachusetts Institute of Technology\\
  Cambridge, MA, USA \\
  \And
  Pierfrancesco Beneventano \\
  Massachusetts Institute of Technology\\
  Cambridge, MA, USA \\
  \texttt{pierb@mit.edu} \\
}

\usepackage[section]{placeins}
\begin{document}

\maketitle
\lhead{Preprint}

\begin{abstract}
    Neural networks applied to financial time series operate in a regime of underspecification, where predictors achieve comparable out-of-sample error. Using large-scale volatility forecasting for S\&P 500 stocks---twelve architecture–optimizer configurations, thirteen seeds each, evaluated on 564{,}795 held-out observations---we show that comparable test loss conceals measurably different predictors. In eleven of twelve comparisons, models trained from identical initialization and identical data order diverge $1.3$ to $7.8$ times further than repeated runs of either optimizer alone, ruling out initialization noise as the explanation. On a common scale, predictions differ by 5--17\% of the typical forecast error while the models' error magnitudes differ by less than 0.7\% of it. The divergence reshapes non-linear response profiles, temporal dependence, and ultimately decisions: volatility-ranked portfolios trace a Sharpe–turnover frontier with up to 58\% turnover dispersion at comparable gross Sharpe ratios. In underspecified settings, optimization is a consequential source of inductive bias, and model evaluation must extend beyond scalar loss to functional and decision-level behavior.
\end{abstract}

\section{Introduction}
\label{sec:intro}

\paragraph{Model leaderboard ties.} 

In financial time series forecasting, substantially different predictors often achieve indistinguishable out-of-sample performance under standard loss metrics. Recent benchmark efforts and empirical studies document that deep architectures often match, and sometimes fail to exceed, linear econometric baselines (see \citealp{hu2025fintsbcomprehensivepracticalbenchmark} or Table~\ref{tab:main_results_nmse}).

When loss-based evaluation cannot distinguish among models, the practical question is no longer which model performs best on a leaderboard, but whether these models are meaningfully different. This leads to our first question:

\begin{summarybox}

    \textbf{Question 1:}
    \emph{Are models with comparable test loss actually\\ interchangeable? Should we employ deep networks at all?}
\end{summarybox}

In practice, one must still deploy \textbf{one} model. On what basis should this choice be made? 

\paragraph{Optimizer choice appears inconsequential.}

In vision and language, similar training losses can coexist with substantially different test performance, making optimizer tuning central to model selection \citep{zhao2024deconstructing,wilson2017marginal}. In financial forecasting, even validation and test losses are often close. Studies commonly prioritize architecture and data while defaulting to Adam \citep{gu2020empirical,chen2024deep}. This motivates our second question:

\begin{summarybox}
    \textbf{Question 2:}
    \emph{Does the optimizer merely affect training efficiency, or does it materially affect the learned function even when the test loss is comparable?}
\end{summarybox}

\paragraph{Overview and takeaways.}
We study the fundamental task of volatility forecasting for S\&P~500 stocks and empirically investigate the questions introduced above. Using a controlled comparison of architectures (MLP, CNN, LSTM, Transformer) and optimizers (SGD, Adam, Muon) with different initializations and hyperparameters, we establish the following:

\begin{summarybox}
\textbf{Takeaway 1:}
    \textit{Comparable test loss does not imply interchangeable predictors: the pairs \textit{(architecture, optimizer)} jointly shape the learned \emph{response surfaces}.}
\end{summarybox}

We measure prediction disagreement on real held-out observations, compare it with variability across repeated fits, and test predictive equivalence within explicit error tolerances. We further characterize the nature of these differences. We reveal simpler versus more complex behavior across optimizers (Fig.~\ref{fig:diff_cnn}); optimizer-difference surfaces exhibit highly structured patterns for sequence models (Fig.~\ref{fig:all_difference_surfaces}); and temporal attributions uncover optimizer-dependent ``effective receptive fields'' (Fig.~\ref{fig:shap_composite}).
Crucially, these differences are not merely aesthetic: when forecasts are used to rank assets by volatility, portfolio turnover varies substantially despite comparable risk-adjusted performance (Fig.~\ref{fig:sharpe_turnover_q1q5}; Appendix~\ref{app:vol_managed}).
\begin{summarybox}
\textbf{Takeaway 2:}
    \textit{Valid benchmarking and model selection should not rely on minor oscillations of the loss, but on interpretability or further financial metrics.}
\end{summarybox}
The paper proceeds as follows: Sections~\ref{sec:methodology}--\ref{sec:result1} establish the forecasting comparison; Section~\ref{sec:result2} measures and interprets functional divergence, and Section~\ref{sec:result3} examines its financial consequences.

\section{Experimental Framework}
\label{sec:methodology}

We hold data and evaluation fixed while varying architecture and optimizer configuration, measuring their effects on forecasting error and the learned function.

\subsection{Task Definition: Volatility Forecasting}
\label{subsec:task}
We focus on one-step-ahead volatility forecasting, a task characterized by a low signal-to-noise ratio but substantial temporal dependence. Unlike raw returns, daily volatility exhibits strong autocorrelation and clustering, making it an ideal testbed for examining inductive bias under weak identifiability.

We build a survivorship-bias-free dataset of S\&P 500 constituents spanning 2000--2024, constructed from CRSP (see Appendices~\ref{app:choosing_dataset} and~\ref{app:dataset_analysis} for further details).
The target variable is realized variance, proxied by the Garman--Klass estimator \citep{0d979ce1-e68e-3a44-a15a-4554e7fd21ca} computed from daily high ($H_t$), low ($L_t$), open ($O_t$), and close ($C_t$) prices:
\begin{equation}
\sigma^2_{GK,t} = \frac{1}{2} \left(\ln \frac{H_t}{L_t}\right)^2 - \left(2\ln 2 - 1\right) \left(\ln \frac{C_t}{O_t}\right)^2.
\end{equation}
Models are trained to predict the log-variance, $\ln(\sigma^2_{GK, t+1})$ using a lookback window of past daily volatilities. While the model predicts log-variance for numerical stability, we discuss results in terms of volatility, as the two are equivalent up to a monotonic transformation.

\subsection{Model--Optimizer Pairs}
We define a learning system as a tuple $\mathcal{T} = (\mathcal{A}, \mathcal{O})$, where $\mathcal{A}$ and $\mathcal{O}$ denote the network architecture and the optimizer, respectively. This formulation emphasizes that the learned function is jointly determined by representational capacity and training dynamics.

\textbf{Architectures ($\mathcal{A}$).} We consider four standard neural architectures that impose distinct inductive biases over temporal data. Specifically, we consider: (1) MLPs, (2) CNNs \citep{726791}, (3) LSTMs \citep{hochreiter1997long}, and (4) Transformer models \citep{vaswani2017attention}. All four architectures have been previously applied to financial forecasting tasks, including volatility and return prediction \citep{gu2020empirical, chen2024deep}.

\textbf{Optimizers ($\mathcal{O}$).} We contrast three optimization methods that differ in update geometry and implicit regularization. Specifically, we consider: (1) SGD, serving as a baseline for non-adaptive first-order optimization; (2) Adam \citep{kingma2017adammethodstochasticoptimization}, the default adaptive moment estimation algorithm; and (3) Muon \citep{jordan2024muon}, a recent matrix-aware optimizer designed for high-dimensional training dynamics. Across all optimizers, we fine-tuned hyperparameters as specified in Appendix \ref{app:experimental_setup}.

\paragraph{Experimental grid.}
We evaluate $4$ architectures $\times\;3$ optimizers $=\;12$ learning systems. For each system we tune learning rate and weight decay on a fixed validation split, then fit on merged training and validation data. We report test NMSE over $13$ repeated fits per configuration and use separately identified representative checkpoints for single-model diagnostics. The selected LSTM--SGD representative and its 13 repetitions share the same hyperparameters and fitting budget. Appendix~\ref{app:experimental_setup} records these distinctions. Predictive and functional evaluations cover all $564{,}795$ available observations in the test set. 
\subsection{Functional Diagnostics Beyond Error Metrics}
We complement forecasting error with prediction distances, controlled input probes and lag attributions.

\paragraph{Distance between predictions.}
We measure disagreement on identical held-out observations in a common standardized log-variance scale using empirical $L^2$ distance (Eq.~\eqref{eq:function_distance}). Its square estimates integrated squared prediction disagreement. This compares predictions directly, rather than comparing each predictor's forecasting error.

\paragraph{Impulse Response Analysis.}

\begin{figure}[tbp]\centering
\begin{minipage}[c]{.42\linewidth}\centering
\includegraphics[width=\linewidth]{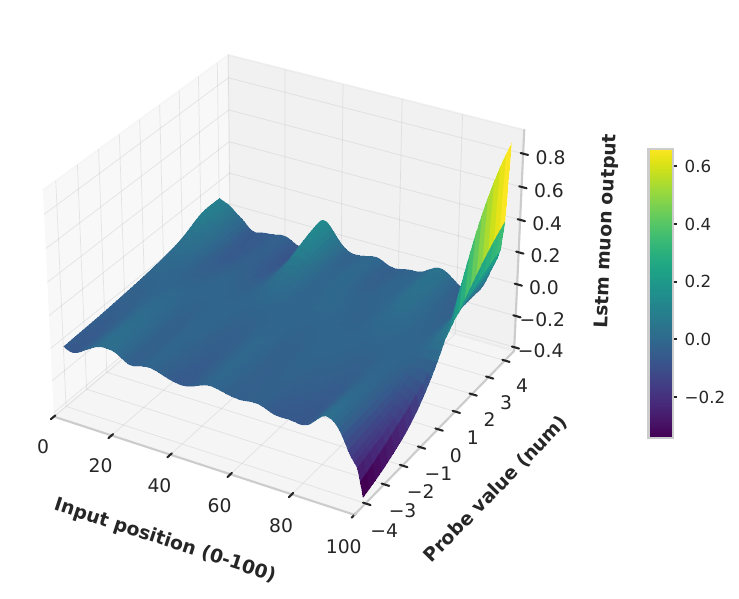}
\captionof{figure}{\textbf{LSTM Response Surface.} Visualization of the impulse response $\mathcal{R}(k, \delta)$. The LSTM (trained with Muon) exhibits a curved response surface, indicating distinct non-linear sensitivity to volatility shocks at specific lags.}\label{fig:lstm_response}
\end{minipage}\hfill
\begin{minipage}[c]{.56\linewidth}\centering
\captionof{table}{\textbf{Out-of-Sample Performance (NMSE)}. Comparison of generalization error across all Architecture-Optimizer pairs. Mean $\pm$ SD over 13 runs. Lower is better.}\label{tab:main_results_nmse}
\vspace{.6em}\resizebox{\linewidth}{!}{\begin{tabular}{lccc}
        \toprule
        \multicolumn{4}{c}{\textbf{Panel A: Linear Baselines}} \\
        \midrule
        Model & NMSE & & \\
        \midrule
        OLS   & $0.5751$ & & \\
        LASSO & $0.5771$ & & \\
        HAR   & $0.5671$ & & \\
        \midrule
        \multicolumn{4}{c}{\textbf{Panel B: Neural Models}} \\
        \midrule
        Architecture & Adam & Muon & SGD \\
        \midrule
        MLP         & $0.5752 \pm 0.0012$ & $0.5770 \pm 0.0002$ & $0.5781 \pm 0.0040$ \\
        CNN         & $0.5747 \pm 0.0015$ & $0.5758 \pm 0.0007$ & $0.5905 \pm 0.0147$ \\
        LSTM        & $0.5769 \pm 0.0026$ & $0.5783 \pm 0.0018$ & $0.5899 \pm 0.0203$ \\
        Transformer & $0.5751 \pm 0.0016$ & $0.5764 \pm 0.0032$ & $0.5761 \pm 0.0015$ \\
        \bottomrule
    \end{tabular}}
\end{minipage}
\end{figure}

To map the model's response surface, we measure its global sensitivity by constructing synthetic input vectors. We define the impulse response $\mathcal{R}(k, \delta)$ as the model output when the $k$-th lag is set to value $\delta$ and all other inputs are held at their mean (zero):
\begin{equation}
\mathcal{R}(k, \delta) = \hat{y}(\delta \cdot \mathbf{e}_k),
\end{equation}
where $\mathbf{e}_k$ is the standard basis vector corresponding to lag $k$. We vary $\delta \in [-4, 4]$ (in standard deviation units) to visualize the architecture's inherent nonlinearity, independent of specific market contexts; Figure~\ref{fig:lstm_response} shows the resulting surface for the LSTM.

\paragraph{Functional Difference Surfaces.} To compare optimizer-induced solutions, we compute output difference surfaces, for instance
\begin{equation}
\Delta_{M,A}(x) = \hat{y}_{\text{Muon}}(x) - \hat{y}_{\text{Adam}}(x), 
\end{equation}
evaluated over the same input space. Non-planar $\Delta_{M,A}$ reveals input-dependent disagreement along the probe, without identifying parameter-space minima (see Figure \ref{fig:all_difference_surfaces}).

\paragraph{Feature Attribution via SHAP.} To evaluate the marginal contribution of each input lag to the model's output, we employ SHapley Additive exPlanations (SHAP) \citep{lundberg2017unifiedapproachinterpretingmodel}. This allows us to determine whether different optimizers prioritize distinct temporal windows (e.g.,  recent momentum at $t-1$ \textit{vs} quarterly cycles at $t-60$) while achieving comparable predictive error. 

\section{The Phenomenon of Predictive Equivalence}
\label{sec:result1}

In domains characterized by low signal-to-noise ratios, such as financial time series, standard aggregate performance metrics often fail to identify a unique predictor. This phenomenon has been formalized as \textit{underspecification} \citep{damour2020underspecification}, extending Breiman's \textit{Rashomon Effect} \citep{breiman2001statistical}. Previous work has focused on high-noise settings, where models with indistinguishable test performance may assign conflicting predictions to individual inputs \citep{marx2020predictivemultiplicityclassification, black2022multiplicity}. In overparameterized regimes, training dynamics play a central role in selecting among many compatible solutions \citep{wilson2017marginal, zou2021adam}. For instance, batch size and stochasticity influence the sharpness of the selected solutions \citep{keskar2017largebatch}. See Appendix~\ref{app:further_work} for a detailed discussion.

In this section, we provide an empirical characterization of underspecification in financial volatility forecasting as described in Section \ref{subsec:task}. Specifically, we train all combinations of architectures (MLP, CNN, LSTM, Transformer) and optimizers (Adam, SGD, Muon) and document a systematic phenomenon we refer to as \textit{predictive equivalence}: many optimizer--architecture pairs have small error differences, with the extent of equivalence depending on an explicit tolerance. 

\subsection{The Leaderboard Tie}

\label{sec:leaderboardtie}

To ensure robustness, we repeat the training procedure for every architecture--optimizer pair using $N=13$ distinct random seeds for initialization and data shuffling. Table~\ref{tab:main_results_nmse} reports the mean NMSE and its standard deviation across seeds. The following observations emerge:
\begin{itemize}[leftmargin=1em,itemsep=0.25em,topsep=0.15em,parsep=0pt]
    \item \textbf{Linear baselines remain competitive.} Neural networks do not materially outperform linear benchmarks. OLS and LASSO achieve error levels comparable to even the most complex Transformer models, while HAR \citep{10.1093/jjfinec/nbp001}---which encodes hand-crafted multi-horizon aggregates (weekly/monthly/quarterly volatility)---attains the lowest NMSE in the table. This is consistent with prior evidence that parsimonious econometric models with strong domain priors are difficult to beat in low-SNR univariate volatility forecasting \citep{BRANCO2024101524, VORTELINOS2017824}.

\item \textbf{The leaderboard tie persists.} After Holm--Bonferroni correction, only 7 of the 66 comparisons reject equality of expected out-of-sample NMSE. The remaining 59 comparisons (89\%) are statistically indistinguishable under this test (Appendix~\ref{app:welch_tests}).

    \item \textbf{Equivalence within stated tolerances.} Two one-sided tests (TOST) assess whether the mean-NMSE difference lies inside $(-\Delta,\Delta)$. At exploratory margins $\Delta=0.001,0.002,0.003,0.005,0.010$, respectively $0,1,10,31,45$ of $66$ comparisons pass Holm correction at $5\%$ within each margin. The five margins provide an exploratory sensitivity analysis (Appendix~\ref{app:welch_tests}).
    \item \textbf{Small error differences, distinct predictions.} MLP--Muon has mean NMSE $0.00176$ above MLP--Adam: exact equality is rejected (Holm $p=0.0090$), yet equivalence is supported at $\Delta=0.005$. Their mean across-optimizer RMS prediction distance is $0.0539$, compared with within-optimizer distances $0.0400$ for Adam and $0.0119$ for Muon (Section~\ref{sec:repeated_functions}).
\end{itemize}

The leaderboard leaves room for markedly different predictors. Across a broad set of comparisons, error differences are either statistically unresolved or bounded within explicit tolerances; yet the corresponding functions need not agree on what comes next. We now show that this freedom has structure: optimizer configurations select different responses to the same information, and those differences persist in the forecasts and decisions they produce.

\subsection{Robustness via Hyperparameter Optimization}
A potential concern in comparing optimization algorithms is that
performance parity may arise from suboptimal tuning of one or both
methods. To address this concern, we perform optimizer-specific
hyperparameter optimization for every architecture--optimizer pair
using the Optuna framework
\citep{akiba2019optunanextgenerationhyperparameteroptimization}.
We tune the learning rate and weight decay separately for each pair,
selecting the configuration that minimizes validation loss.
Search ranges, selected hyperparameters, and implementation details
are reported in Appendix~\ref{app:experimental_setup}.

The results reported in Table~\ref{tab:main_results_nmse} reflect
these validation-selected configurations. Strikingly, even after
optimizer-specific tuning, performance remains statistically
indistinguishable across the pairwise comparisons identified in
Section~\ref{sec:leaderboardtie}. Hyperparameter optimization therefore
\textit{fails to break the broad leaderboard tie}, in contrast to
the pronounced performance differences often observed between
optimizers in vision and language tasks.

\section{Functional Divergence via Optimization}
\label{sec:result2}

Section~\ref{sec:result1} establishes a large matched-loss region. Unlike vision or NLP, where optimizers induce meaningful generalization differences, most pairs here are either statistically unresolved or equivalent within stated tolerances. This predictive equivalence masks deep heterogeneity: qualitatively distinct predictors remain equally compatible with the data (Appendix~\ref{app:further_work}).

We investigate the \emph{functions} learned by each pair $(\mathcal{A}, \mathcal{O})$, using the representative model retained for each configuration. Holding architecture fixed, the optimizer acts as a source of inductive bias, selecting qualitatively different decision rules from the same hypothesis class. We call this \emph{functional divergence}: models metrically equivalent in predictive error, yet geometrically and economically distinct in how they map past information into forecasts.

\subsection{Functional Divergence}
\label{sec:distance}
\begin{figure}[t]
\centering\includegraphics[width=\linewidth]{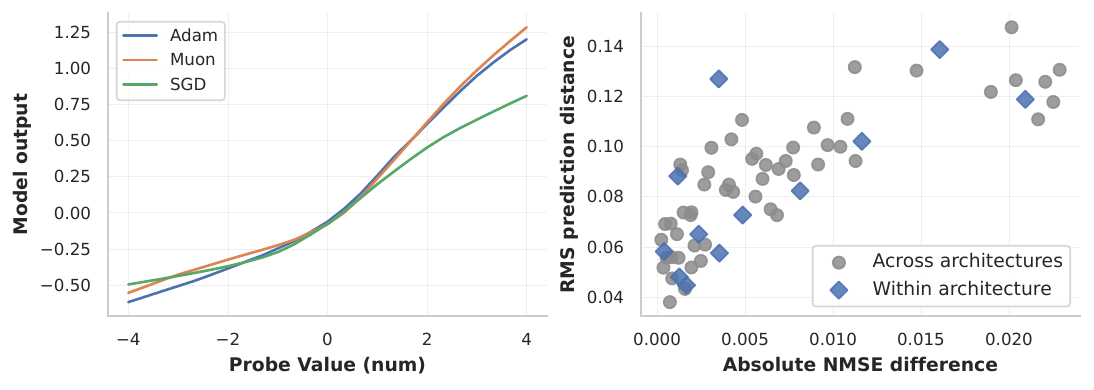}
\caption{\textbf{Similar error, different functions.} (a) CNN response to an impulse at $t-1$, with other inputs held at their standardized means. (b) Absolute NMSE difference versus RMS prediction distance for all 66 representative pairs on the same held-out observations. Diamonds identify comparisons within an architecture. Prediction distances use standardized log-variance units.}
\label{fig:functional_distance}\label{fig:difference_adam_muon_t-1_cnn}
\end{figure}

The same forecasting score can conceal markedly different responses to the same input. Figure~\ref{fig:functional_distance}a makes this visible for CNNs: changing the optimizer changes the response to the most recent observation, including how that response bends and dampens large shocks. To see how broadly this disagreement extends, we evaluate the fitted functions on the same held-out observations, rather than only on synthetic probes.

For predictions in a common output scale, their empirical $L^2$ distance is
\begin{equation}
D(f,g)=\left[\frac{1}{N}\sum_{i=1}^{N}\bigl(f(x_i)-g(x_i)\bigr)^2\right]^{1/2}.
\label{eq:function_distance}
\end{equation}
Its square estimates $\int(f-g)^2\,dP_x$. Unlike an error difference, which compares each predictor with the target, $D$ compares their predictions directly. Figure~\ref{fig:functional_distance}b evaluates all $\binom{12}{2}=66$ representative pairs on all $564{,}795$ test observations. Small differences in NMSE coexist with substantial prediction disagreement: closeness on the leaderboard does not imply closeness of the learned functions.

The response surfaces reveal the structure of this disagreement. Figure~\ref{fig:all_difference_surfaces} in Appendix~\ref{app:functional_divergence_analysis} plots $\Delta_{M,A}(x)=\hat y_{\mathrm{Muon}}(x)-\hat y_{\mathrm{Adam}}(x)$ for CNN, LSTM and Transformer. 
These single-impulse slices are structured across both lag and impulse size: the optimizers select different sensitivities to past information, even within the same architecture.

\subsection{Functional Differences across Repeated Fits}
\label{sec:repeated_functions}
\begin{figure}[tbp]
\begin{minipage}[c]{.54\linewidth}
\centering
\includegraphics[width=\linewidth]{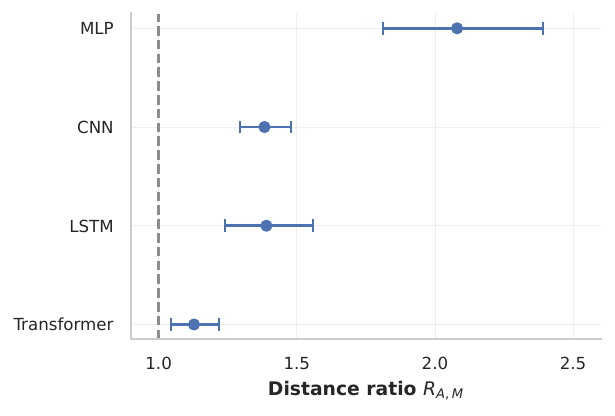}
\captionof{figure}{\textbf{Functional separation beyond repeated-fit variability.} Adam--Muon ratios $R_{A,M}$, with pointwise 95\% seed-level jackknife intervals. The dashed line marks $R=1$.}\label{fig:distance_ratios}

\end{minipage}\hfill
\begin{minipage}[c]{.43\linewidth}
\centering\small
\captionof{table}{Components of the Adam--Muon ratio, in RMS prediction units.}\label{tab:distance_components_main}
\begin{tabular}{lrrr}\toprule
 & $W_A$ & $W_M$ & $C_{A,M}$\\\midrule
MLP & 0.0400 & 0.0119 & 0.0539 \\
CNN & 0.0475 & 0.0364 & 0.0580 \\
LSTM & 0.0477 & 0.0519 & 0.0693 \\
Transf. & 0.0461 & 0.0505 & 0.0545 \\
\bottomrule\end{tabular}

\end{minipage}
\end{figure}

Could the observed differences simply reflect initialization? We compare disagreement across optimizers with disagreement among repeated fits of the same configuration. Let $f_{A,i}$ be fit $i$ from optimizer configuration $A$, with $n_A=13$ repetitions. Define
\begin{gather}
W_A=\frac{2}{n_A(n_A-1)}\sum_{i<j}D(f_{A,i},f_{A,j}),\qquad
C_{AB}=\frac{1}{n_A n_B}\sum_{i,j}D(f_{A,i},f_{B,j}),\label{eq:within_across}\\
\boxed{R_{AB}=\frac{C_{AB}}{(W_A+W_B)/2}}.\label{eq:distance_ratio}
\end{gather}
$W_A$ measures typical disagreement between repeated fits of the same configuration; $C_{AB}$ measures disagreement across the two configurations. Thus $R_{AB}=2$ means that across-optimizer disagreement is twice the average within-optimizer disagreement. Because this denominator is an average, a cohort with unusually reproducible fits can inflate $R_{AB}$; 
we therefore also report the stricter ratio $R^{\max}_{AB}=C_{AB}/\max(W_A,W_B)$, which compares across-optimizer disagreement with the larger of the two within-optimizer levels (Appendix~\ref{app:functional_distance}); Table~\ref{tab:distance_components_main} gives the components it is built from.

For Adam--Muon, $R_{A,M}$ is $2.08$ for MLP, $1.38$ for CNN, $1.39$ for LSTM and $1.13$ for Transformer (Figure~\ref{fig:distance_ratios}). Their pointwise 95\% intervals are $[1.81,2.39]$, $[1.29,1.48]$, $[1.24,1.56]$ and $[1.04,1.22]$, respectively. Under the stricter denominator the MLP, CNN and LSTM comparisons retain intervals excluding one ($R^{\max}=1.35\,[1.13,1.61]$, $1.22\,[1.05,1.42]$ and $1.33\,[1.18,1.51]$), whereas Transformer does not ($1.08\,[0.89,1.31]$): for these three architectures the optimizer moves predictions further than rerunning either optimizer on its own, not merely further than their average. This connects directly to the MLP Adam--Muon predictive comparison: a small but statistically detectable mean-error difference satisfies the exploratory equivalence margin $\Delta=0.005$ NMSE after Holm correction, while the learned functions remain well separated.

The size of the effect depends on the architecture, and we report where it is largest and where it is not: in LSTM Adam--SGD the across-optimizer distance falls between the two within-optimizer levels, and the interval for $R$ includes one. Read together, these comparisons locate the optimizer as a substantial and measurable source of functional separation, on a scale set by repeated fitting itself. Full results and the uncertainty construction appear in Appendix~\ref{app:functional_distance}.

\paragraph{Functional divergence persists under matched initialization and data order.}
The decisive test removes randomness from the comparison altogether. We refit every architecture from three shared initializations under a common data order (36 runs), so that two optimizers see identical starting weights and identical batches. \emph{In eleven of twelve comparisons the resulting predictors diverge $1.3$ to $7.8$ times further than repeated runs of either optimizer alone}---a margin the unpaired seed cohorts cannot resolve, since there the initialization variance sits on both sides of the comparison. The single exception is Transformer Adam--Muon, where the paired and repeated-fit distances coincide. Putting error and disagreement on one scale makes the gap concrete: predictions differ by $5.0$--$16.7\%$ of the typical forecast error, while the two models' error magnitudes differ by less than $0.7\%$ of it, a ratio of $21$ to $2859$. MLP SGD--Adam is the extreme case, matching to $2\times10^{-5}$ in RMSE while sitting $0.0593$ apart in prediction space. Initialization noise cannot produce this: it is the smaller effect in every single comparison.
Appendix~\ref{app:paired_initialization} reports all configurations and puts
the loss and functional gaps on a common scale.

\subsection{Optimizer-Driven Feature Attribution}
\label{sec:shap}
To understand which parts of the input history distinguish these functions, we analyze lag attribution using SHAP \citep{lundberg2017unifiedapproachinterpretingmodel}. The resulting temporal profiles give structure to the prediction distances: optimizers differ not only in the magnitude of their forecasts, but also in the past information associated with those forecasts.

\begin{figure}[tbp]\centering
\includegraphics[width=\linewidth]{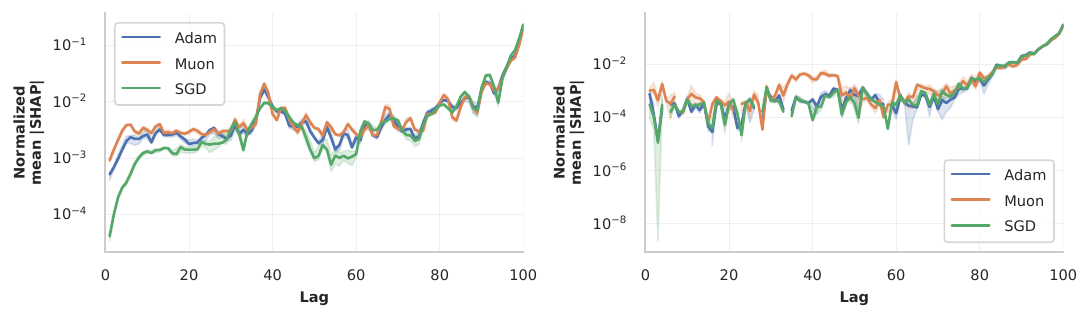}
\caption{\textbf{Optimizer-dependent lag importance.} CNN (left) and LSTM (right). Normalized mean absolute SHAP values across the 100 input lags. Lines show medians over 13 repeated fits; shaded bands show interquartile ranges.}\label{fig:shap_composite}\label{fig:shap_cnn}\label{fig:shap_lstm}
\end{figure}

For CNNs, recent lags dominate, but optimizers differ in their secondary attribution structure. For LSTMs, Muon develops a pronounced mid-horizon band, whereas Adam concentrates more strongly on recent history (Figure~\ref{fig:shap_composite}). The architecture supplies the capacity to use the full input history; the optimization procedure helps determine which temporal structure the fitted predictor actually uses. 
From a financial perspective, recent-lag emphasis is consistent with short-horizon volatility persistence, while attribution at longer horizons reflects a different temporal weighting of past information \citep{engle1982autoregressive,andersen2007real}. These describe the predictors' dependence on observed history rather than identifying market mechanisms. The finding stands regardless: at matched forecasting error, optimizers select different effective receptive fields over the past, and so different representations of volatility persistence. Appendix~\ref{app:shap_protocol} gives the estimators, shared evaluation samples and attribution checks.

\FloatBarrier
\subsection{Mechanism: Curvature Constraints}
\label{sec:mechanism}
Why do these optimizers induce functional divergence? Their update rules---plain mini-batch gradients, coordinatewise adaptive scaling, and orthogonalized matrix-valued momentum \citep{jordan2024muon}---can impose different constraints on the regions explored during fitting.

A curvature plateau need not mean that optimization has settled at a fixed minimum: it may instead be held in place by the update dynamics. The two accounts make opposite predictions when the step size is cut. Around a fixed quadratic minimum, a smaller learning rate contracts fluctuations and leaves the Hessian alone \citep{mandt2017}; under a dynamical constraint, it releases renewed sharpening \citep{andreyev2025edgestochasticstabilityrevisiting}. We therefore continue each architecture--optimizer configuration from a common fitted checkpoint and its saved optimizer state, comparing an unchanged control against a tenfold learning-rate reduction and, separately, a doubling of batch size.

\begin{figure}[t]
\centering\includegraphics[width=\linewidth]{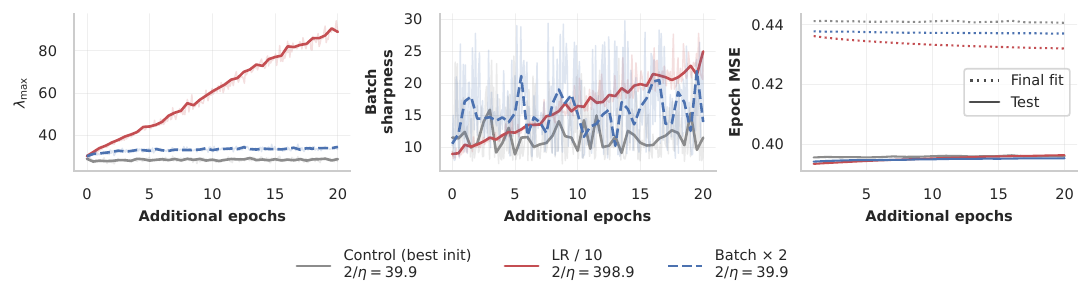}
\caption{\textbf{Releasing the curvature constraint.} MLP--SGD continuations from common fitted weights and optimizer state: unchanged control, LR/10 and batch$\times2$. Left to right: raw $\lambda_{\max}$, batch sharpness, and epoch final fit and test MSE over the same 20 additional epochs. The legend reports $2/\eta$ for each arm. Light curvature traces are individual measurements; dark curvature traces are half-epoch medians. Dotted loss curves show final fit; solid curves show test MSE. Curvature is measured on 4,096 observations; diagnostic batch size is 512, or 1,024 for batch$\times2$.}
\label{fig:stability_interventions}
\end{figure}

The SGD configurations answer clearly. For MLP--SGD (Figure~\ref{fig:stability_interventions}), cutting the learning rate lifts the median leading eigenvalue from approximately $29$ to $85$ over the last fifth of the common window, while the control holds its earlier level---and the loss keeps falling. Sharpening resumes, and the fitted solution was therefore not a minimum the optimizer had settled into: the step size itself was the binding constraint. The CNN diagnostic responds more strongly still. Adaptive optimizers behave differently, and the raw Hessian does not capture their effective update geometry \citep{cohen_adaptive_2022}; Appendix~\ref{app:mechanism} reports every configuration, including those that respond weakly or not at all. Update dynamics, not the loss surface alone, select the region a training run ends in.

\section{Behavioral Divergence in Trading Strategies}

\label{sec:result3}

Two MLP predictors with gross Sharpe ratios of $0.420$ and $0.411$ demand $16.3\%$ and $10.5\%$ daily turnover respectively: one requires roughly $55\%$ more trading than the other to express what is, on the leaderboard, the same forecast. Functional divergence becomes economic divergence as soon as forecasts drive decisions \citep{rudin2022interpretable,fisher2019all}. We use volatility-ranked portfolios to measure this decision multiplicity, with construction details and the Sharpe--turnover frontier in Appendix~\ref{app:vol_managed}.

Using the same representative predictors, we construct daily equal-weight volatility-quintile portfolios over May 2000--December 2024. This retrospective diagnostic includes fitting observations; Appendix~\ref{app:vol_managed} reports the held-out comparison. Gross risk-adjusted performance remains close, but daily turnover among neural predictors spans 14.8--20.8\% in the low-volatility quintile (Q1) and 10.5--16.6\% in the high-volatility quintile (Q5), versus 20.3\% and 15.0\% for OLS.

The ordering itself depends on architecture: SGD trades least for CNN and LSTM, Muon least for MLP. No optimizer is uniformly the cheaper one to run, which is precisely the point---the leaderboard cannot tell you which allocation rule you are buying, and similar gross performance does not make the induced rules interchangeable.

Turnover reflects changes in forecast-induced rankings, not response-surface smoothness alone, and under common proportional trading costs greater turnover incurs greater costs. These are gross comparisons, and we do not claim net profitability or capacity. What they establish is sharper than that: a predictor's error tells you almost nothing about the trading it will demand, so model evaluation must account for the decisions a predictor induces, not only the loss it attains.

\section{Conclusion}
\label{sec:conclusion}

\paragraph{Predictive equivalence is real and misleading.}

Financial volatility forecasting exhibits a \emph{Rashomon regime}: many architecture--optimizer comparisons have close mean errors, with equivalence supported within explicitly reported tolerances.
But similar NMSE does not imply similar predictions: held-out RMS distances quantify functional disagreement relative to variability across repeated fits.
Impulse-response profiles, optimizer-difference surfaces, and lag attributions reveal materially different input--output mappings, which translate into different stability and turnover once forecasts are embedded into trading rules.

\paragraph{Question 1 (interchangeability / ``should we use deep nets?'').}

\textbf{No}: tied predictors are not interchangeable, because ``same loss'' need not mean ``same behavior'' downstream.
\textbf{Yes}: deep networks still matter---not as leaderboard winners, but as flexible hypothesis classes that realize a \emph{menu of admissible signals} (stable \textit{vs}  reactive, smooth \textit{vs}  aggressive, short- \textit{vs}  long-memory).

\paragraph{Question 2 (``is the optimizer just engineering?'').}

\textbf{No}. When loss is indifferent, optimization is an \emph{implicit prior}:
holding data and architecture fixed, optimizer configurations can select \emph{which} admissible function is returned, often determining implementability (e.g., turnover) even when predictive accuracy is comparable.
Matched-checkpoint interventions trace this to the training dynamics themselves: in the SGD configurations, reducing the step size releases renewed sharpening and further fitting, so the fitted solution was a constraint of the optimizer rather than a minimum of the loss.
\textbf{Takeaway.} In underspecified time series, model selection is function selection.
When the leaderboard is a tie, choose the function that matches the downstream economic objective---or the optimizer will choose it for you.
In finance, the optimizer is part of the model.

\subsection*{AI use statement}
LLMs assisted with research discussions, implementation, evaluation scripts, literature checking, and manuscript revision, including the presentation of mathematical arguments. No synthetic observations were generated for empirical evaluation. The authors are responsible for reviewing the analyses, interpretations, and final content before submission.

\subsection*{Reproducibility statement}
Every result in this paper is generated from recorded artifacts rather than re-derived at write-up time. Appendix~\ref{app:dataset_analysis} gives the exact CRSP WRDS query defining the survivorship-bias-free panel, and Appendix~\ref{app:experimental_setup} records the preprocessing pipeline, the architectures, the validation-selected learning rates, weight decay and batch sizes for all twelve configurations, and the numerical settings used for every curvature measurement. Repeated-fit cohorts are frozen: all thirteen runs per configuration are retained, including loss outliers, and the accompanying reproducibility records identify each checkpoint and fitting budget. Prediction distances are computed on one shared panel of $564{,}795$ test observations with a single preprocessing and output scale, and the paired-initialization runs of Appendix~\ref{app:paired_initialization} verify initialization and data-order matching by hashing the loaded tensors and the data permutation.\label{endmain}

\bibliographystyle{iclr2027_conference}
\bibliography{references}

\newpage
\appendix
\makeatletter
\setlength{\@fptop}{0pt}
\setlength{\@fpsep}{16pt}
\makeatother

\section*{Appendix}

\section{Choosing a Dataset}
\label{app:choosing_dataset}

As a Gedankenexperiment, we postulate the ideal characteristics a dataset should possess:
\begin{itemize}
    \item \textbf{Reliability and provenance.} Data are free of obvious errors, typos, and measurement artifacts; missing values are minimal, explicitly flagged, and documented. Units, definitions, and formats are consistent across series and over the full time span. The origin and transformation pipeline are transparent and traceable to reputable sources.
    \item \textbf{Temporal and cross-sectional breadth.} Time series are sufficiently long to span multiple economic regimes and market cycles (e.g., bull and bear markets, recessions, high/low inflation). The cross-section is large (e.g., hundreds of securities rather than a handful), and each timestamp includes a rich feature set beyond prices alone.
    \item \textbf{Diversity with low average correlation.} The universe covers heterogeneous exposures to known risk factors (e.g., value \textit{vs}  growth, small- \textit{vs}  large-cap, high \textit{vs}  low volatility) and a broad range of industries (technology, healthcare, energy, financials, etc.), reducing concentration risk and cross-series redundancy.
    \item \textbf{Modeling readiness.} The provided variables, or standard transformations (e.g., returns), are reasonably stationary, or the dataset includes guidance to achieve stationarity. The predictive task is clearly defined, with one or more precomputed target variables. Standardized training/validation/test splits and a baseline training protocol are supplied for reproducibility and fair comparison.
\end{itemize}

\paragraph{Synthetic Data.}

\citet{LopezDePrado2018Synthetic,10.5555/3454287.3454781} propose using synthetic time series, respectively for backtesting/robustness and for training/evaluation via TSTR. While such data are useful for stress testing and mitigating overfitting, a benchmark’s purpose is to assess performance against real market outcomes. Hence, we contend that synthetic data are not an appropriate benchmarking substrate, for the following reasons:

\begin{enumerate}
    \item \textbf{Benchmarking aims at reality.} The primary purpose of a benchmark is to assess a model's performance against real market outcomes. Synthetic data are, by construction, a simplified and incomplete model of that reality.
    \item \textbf{Model-implied worlds.} A synthetic generator is calibrated to statistical regularities observed in historical data. It cannot, by definition, reproduce dynamics that are absent from (or misspecified in) its data-generating assumptions.
    \item \textbf{Validity is conditional on the generator.} Superior performance on synthetic data merely demonstrates skill at exploiting the rules of the generator, not the market. This invites specification-led overfitting.
    \item \textbf{Structural breaks are not authentically captured.} While one can program regime shifts by changing generator parameters, that does not replicate \emph{unforeseen} changes in market logic (i.e., true structural breaks). For instance, the post-2008 era of quantitative easing altered cross-asset correlations and the risk-free rate regime in ways not anticipated by pre-crisis models.
\end{enumerate}

In short, optimizing to a synthetic world risks tailoring models to the generator's quirks rather than to markets, undermining external validity as a benchmark.

\paragraph{The Closest Real Dataset to the Ideal Dataset.}

We propose a dataset constructed from the Center for Research in Security Prices (CRSP), focusing on daily observations of S\&P~500 constituents. The construction aligns with the ideal characteristics outlined above:

\begin{itemize}
  \item \textbf{Reliability and provenance.} The dataset is sourced exclusively from CRSP, the institutional standard for academic financial research in the United States, whose data undergo rigorous cleaning and validation. Our construction—joining the CRSP Daily Stock File (DSF) with the S\&P~500 constituents list (\texttt{msp500list})—is explicit and reproducible. By using historical membership windows, it correctly handles index composition changes and therefore avoids survivorship bias.

  \item \textbf{Temporal and cross-sectional breadth.} The sample spans from January~2000 to December~2024, covering multiple recessions (2001, 2008, 2020), the dot-com boom and bust, and a secular decline followed by a rise in interest rates. The cross-section comprises all securities during their tenure in the S\&P~500 (roughly 500 at any time). Each observation includes not only the close but also open, high, and low prices, trading volume, and the adjustment factors needed to reflect corporate actions. 

  \item \textbf{Diversity with low average correlation.} The S\&P~500 is a diversified benchmark spanning all major GICS sectors of the U.S.\ economy. Its inclusion criteria yield leading, highly liquid firms with heterogeneous exposures (growth \textit{vs}  value, differing risk profiles) and a large-cap tilt. Tracking constituent histories captures the evolving sectoral and factor composition of the U.S.\ market.

\end{itemize}

\newpage 

\section{Data Diagnostics and Descriptive Checks}
\label{app:dataset_analysis}
\begin{figure}[htbp]
    \centering
    \includegraphics[width=1\linewidth,trim=0 0 0 29bp,clip]{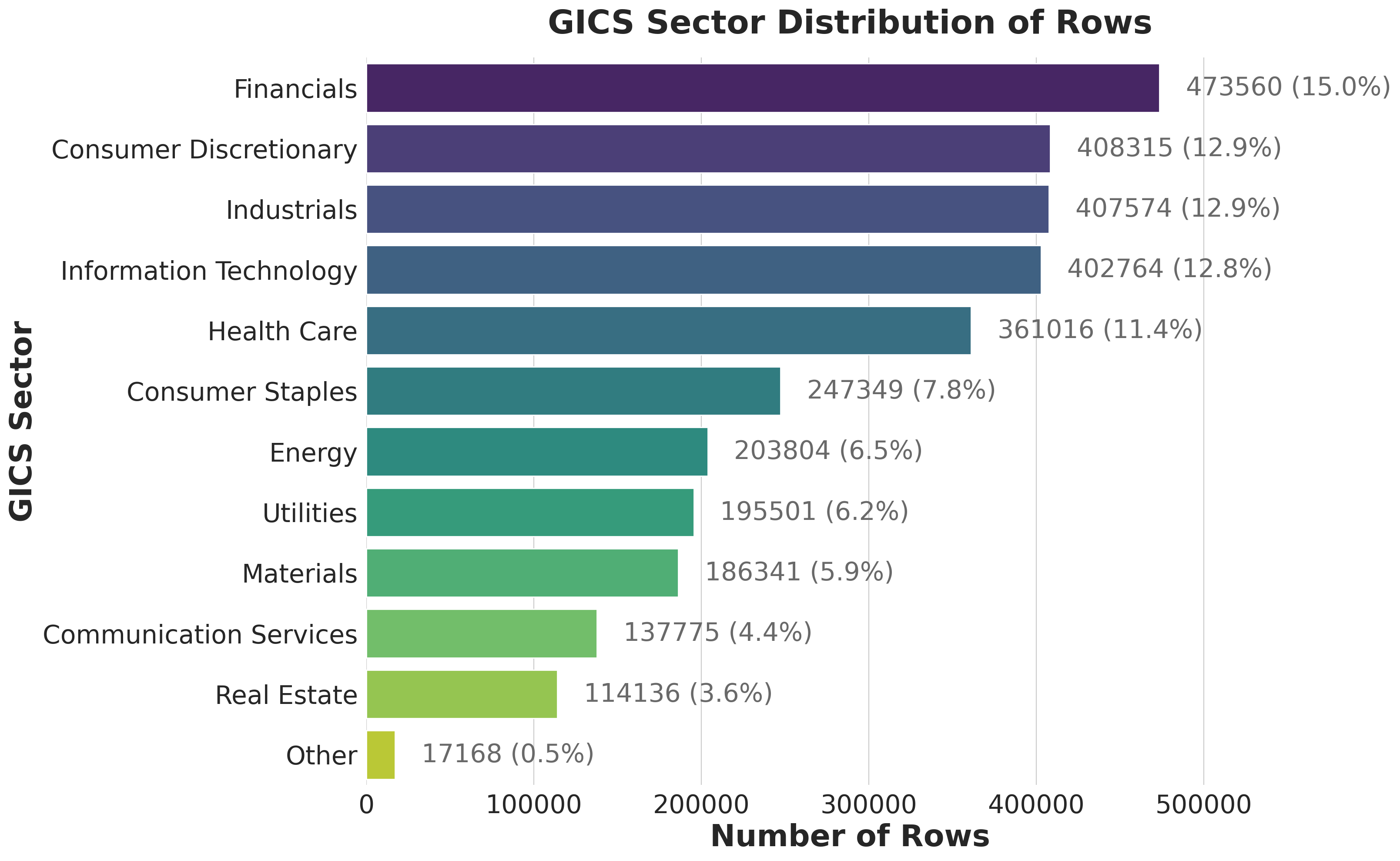}
    \caption{Distribution of observation counts by GICS sector.}
    \label{fig:sector}
\end{figure}

\begin{figure}[h!]
    \centering
    \includegraphics[width=1\linewidth,trim=0 0 0 40bp,clip]{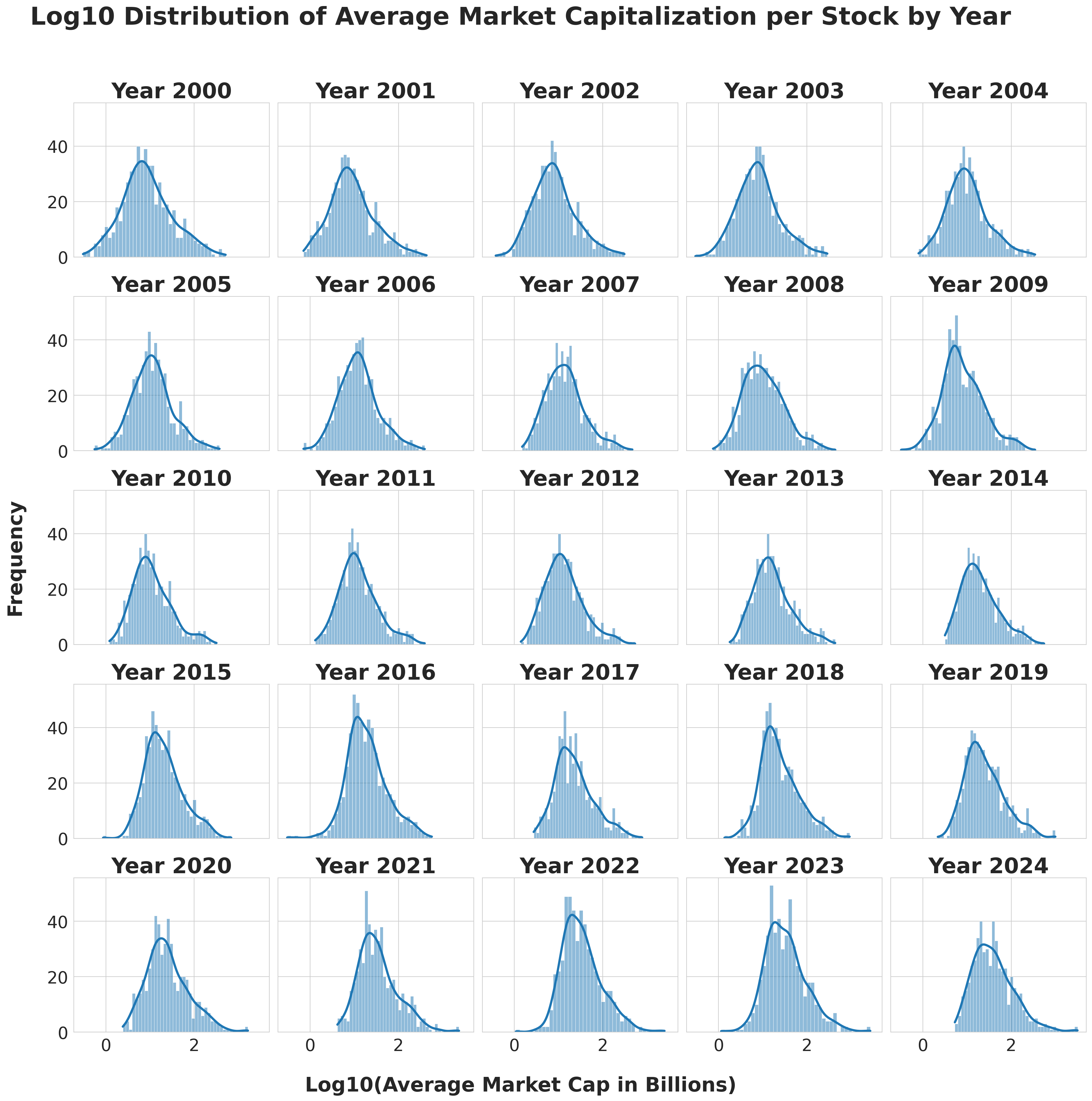}
    \caption{Yearly \(\log_{10}\) distribution of average market capitalization per stock.}
    \label{fig:mcaps}
\end{figure}

\begin{figure}[htbp]\centering
\includegraphics[width=\linewidth,trim=0 0 0 38bp,clip]{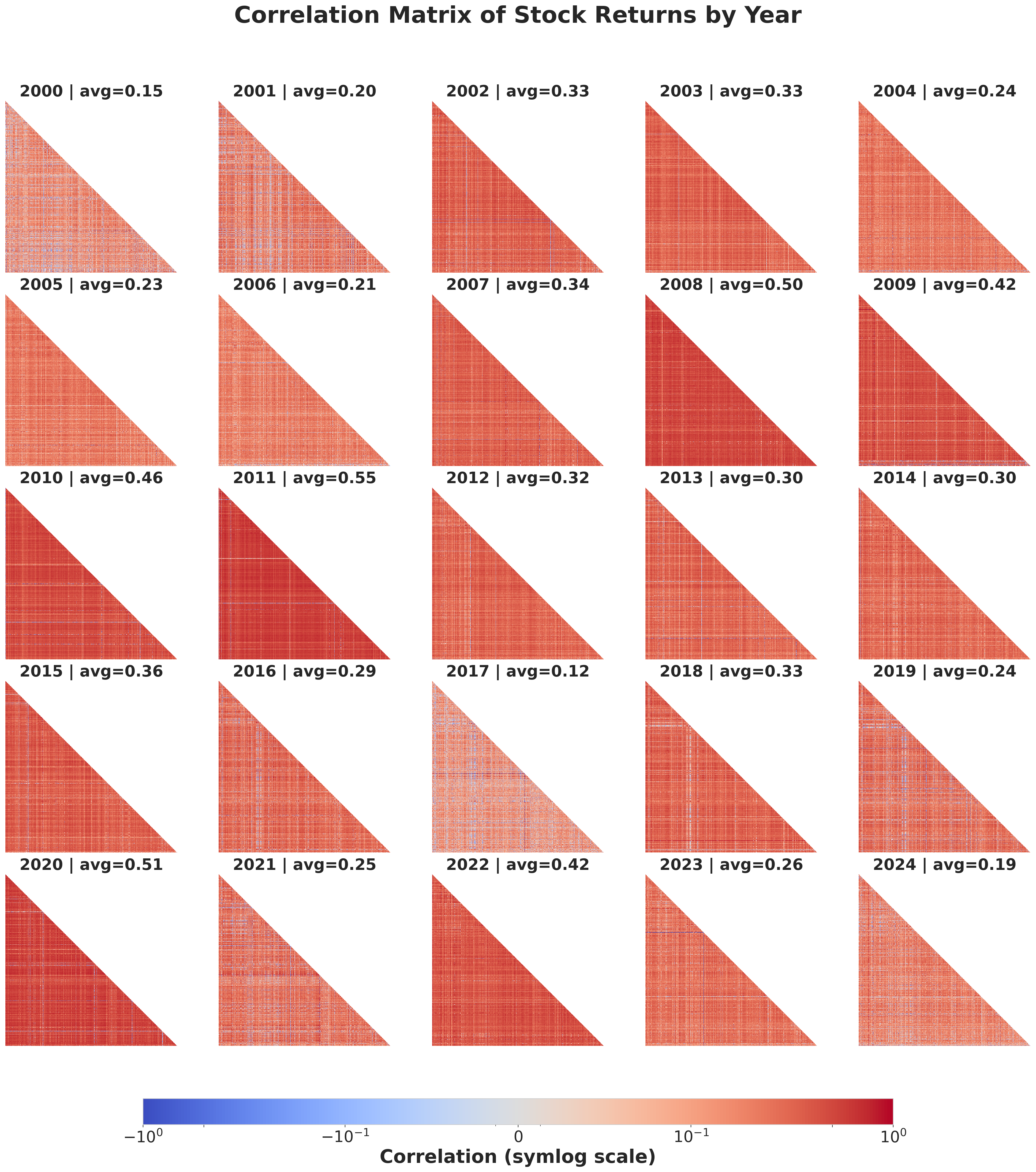}
\caption{Correlation matrices of daily returns by year.}\label{fig:corr}
\end{figure}

Our working panel contains 3{,}155{,}303 daily observations drawn from CRSP for S\&P~500 constituents between January~2000 and December~2024. Each row corresponds to a \texttt{PERMNO}--\texttt{date} pair and includes returns and standard OHLCV fields plus CRSP adjustment factors (Table~\ref{tab:variables}). Figures~\ref{fig:sector}--\ref{fig:corr} summarize sectoral coverage, firm size, and cross-sectional dependence.

\begin{table}[H]
    \centering
    \caption{Core variables in the analysis dataset}
    \label{tab:variables}
    \small
    \begin{tabular}{ll}
        \toprule
        \textbf{Variable} & \textbf{Description} \\
        \midrule
        \ttfamily PERMNO  & CRSP permanent security identifier \\
        \ttfamily date    & Trading day (NYSE calendar) \\
        \ttfamily ret     & Daily return (CRSP; split/dividend adjusted) \\
        \ttfamily open, high, low, close & Daily O/H/L/C prices \\
        \ttfamily vol     & Daily share volume \\
        \ttfamily cfacpr  & Price adjustment factor for corporate actions \\
        \bottomrule
    \end{tabular}
\end{table}

\noindent The sample is balanced: Financials, Consumer Discretionary, Industrials, Information Technology, and Health Care account for most rows, with no sector too concentrated, and all sectors well represented.

\subsection{Firm Size: Market Capitalization}

\noindent As expected for S\&P~500 constituents, capitalization levels are high and, within year, approximately normal on the log scale, with a right tail of very large firms in the latest years (Magnificent 7 effect). The panels show a stable center with modest cyclical shifts across regimes.

\subsection{Cross-Sectional Dependence}

\noindent Correlations are moderate and time-varying: higher in crisis/high-volatility periods (e.g., 2008, 2011, 2020) and lower in calmer years, consistent with standard co-movement dynamics—supporting models with time-varying dependence.

\subsection{Reproducibility Notes}

All series are pulled directly from CRSP using the WRDS Python API, using the following query: 

\begin{lstlisting}[]
SELECT
    s.permno, 
    s.date, 
    s.ret, 
    s.prc AS close, 
    s.vol, 
    s.openprc AS open, 
    s.askhi AS high, 
    s.bidlo AS low, 
    s.cfacpr 
FROM crsp.dsf AS s
JOIN crsp.msp500list AS m
  ON s.permno = m.permno
WHERE s.date BETWEEN GREATEST(m.start, DATE '2000-01-01')
                 AND LEAST(m.ending, DATE '2024-12-31')
ORDER BY s.permno, s.date;
\end{lstlisting}

\newpage

\section{Implementation and Experimental Setup}
\label{app:experimental_setup}

We employ the Garman--Klass estimator to compute a daily realized-variance proxy from OHLC data. We use the original chronological walk-forward construction with train/validation/test ratios 3:1:1. The evaluation includes 2,290,406 merged fitting observations and 564,795 test observations after the original filtering. 
To ensure numerical stability and mitigate the impact of extreme outliers (some exceeding 300 standard deviations), we apply the original per-asset clipping and log-standardization pipeline. During tuning, its fitting data are training-only; for the retained merged fits, preprocessing is refitted on train+validation. Here ``train'' in the formulas denotes the permitted fitting pool:
\[
x \leftarrow \min\!\big(\max(x,\, q_{0.1\%}^{\text{train}}),\, q_{99.9\%}^{\text{train}}\big).
\]
Subsequently, the values are annualized (scaled by 252) and standardized via z-score normalization using the mean and standard deviation of the training set:
\[
x \leftarrow \frac{x - \mu_{\text{train}}}{\sigma_{\text{train}}}.
\]
The final input vector comprises the processed log-variances $\mathbf{x}_t = (\widehat{\sigma}^{2}_{t-L+1}, \ldots, \widehat{\sigma}^{2}_{t})$ used to predict the next-day log-variance $y_t$. We use $L=100$.

\subsection{Model Architectures}
We evaluate four distinct deep learning architectures. All models share a consistent fine-tuning protocol focused on the learning rate and weight decay.

\begin{itemize}
    \item \textbf{LSTM:} We utilize a stacked LSTM with 2 layers of 256 hidden units each. We employ the hidden state of the last time step for prediction (\textit{readout: last}). To preserve signal integrity, we disable dropout and bidirectional processing.
    \item \textbf{CNN:} The convolutional network consists of three 1D-convolution blocks with channel sizes $[64, 128, 256]$, a kernel size of 8, and padding of 1. We use ReLU activation and Adaptive Average Pooling ($k=16$). The output is flattened and passed through an MLP head with sizes $[512, 256]$.
    \item \textbf{MLP:} A pure Multilayer Perceptron baseline consisting of four fully connected layers with sizes $[512, 256, 256, 128]$ and ReLU activations.
    \item \textbf{Transformer:} We use an encoder-only Transformer with $L=2$ layers, model dimension $d_{\text{model}}=128$, and $h=8$ attention heads. Inputs are projected to $d_{\text{model}}$ and combined with learned positional embeddings. The sequence representation is aggregated via mean pooling (\textit{readout: mean}) and passed to an MLP prediction head with sizes $[128, 64]$.
\end{itemize}

\subsection{Linear Baselines}
In addition to deep learning models, we evaluate three linear baselines.
\begin{itemize}
    \item \textbf{OLS:} We implement Ordinary Least Squares regression using \texttt{statsmodels}, computing the weights directly via the closed-form analytical solution.
    \item \textbf{LASSO:} We employ Least Absolute Shrinkage and Selection Operator regression via \texttt{scikit-learn}. The penalization parameter is tuned via grid search over $\alpha \in \{0.001, 0.01, 0.025, 0.05, 0.1, 0.25\}$. We select the $\alpha$ that minimizes the validation loss. Consistent with our general protocol, the final model is refitted on the combined training and validation sets using the optimal $\alpha$.
    \item \textbf{HAR:} We implement the Heterogeneous Autoregressive model. The model takes the form
    \begin{equation}
    \hat{\sigma}^2_{t+1} \;=\; \beta_0 \;+\; \sum_{h \in \mathcal{H}} \beta_h \, \overline{\sigma^2_{t,h}}, 
    \qquad \overline{\sigma^2_{t,h}} \;=\; \frac{1}{h}\sum_{i=0}^{h-1} \sigma^2_{t-i}.
    \end{equation}
    We perform a grid search over the horizon set $\mathcal{H}$, comparing the canonical specification $\{1, 5, 22\}$ (daily, weekly, monthly) of \citet{10.1093/jjfinec/nbp001} with the extended specification $\{5, 22, 66\}$ (weekly, monthly, quarterly), and select the latter as it achieves lower validation loss in our setting. Coefficients are estimated by Ordinary Least Squares on the training set and the model is refitted on the combined training and validation sets prior to test-set evaluation.
\end{itemize}

\subsection{Training, Selection, and Frozen Cohorts}

\label{app:configuration_records}
Hyperparameters are selected by validation loss, after which models are fitted on the combined training and validation data. We distinguish the representative predictor used for single-model diagnostics from the 13 repeated fits used to measure initialization variability. Reported training losses are therefore train+validation fitting losses. Table~\ref{tab:configuration} gives the selected learning rates, weight decay and batch sizes; the accompanying reproducibility records identify the exact checkpoints and fitting budgets.

\begin{table}[htbp]\centering\small
\caption{Selected optimizer settings. The reproducibility records retain full precision.}\label{tab:configuration}
\begin{tabular}{lrrr}\toprule
Configuration & Learning rate $\eta$ & Weight decay & Batch size \\
\midrule
MLP--Adam & 0.00019243323 & 0.01 & 512 \\
MLP--Muon & 0.00019243323 & 0.91 & 512 \\
MLP--SGD & 0.050134518 & 0.0018197008 & 512 \\
CNN--Adam & 0.00019243323 & 0.001 & 512 \\
CNN--Muon & $2.0145131\times10^{-5}$ & 0.0002978566 & 512 \\
CNN--SGD & 0.029985065 & 0.0034651588 & 512 \\
LSTM--Adam & 0.00019243323 & 0.003 & 512 \\
LSTM--Muon & 0.00019243323 & 0.42 & 512 \\
LSTM--SGD & 0.091933565 & 0.0031556215 & 512 \\
Transformer--Adam & $1.0719864\times10^{-5}$ & 0 & 512 \\
Transformer--Muon & $1.7073967\times10^{-5}$ & 0.086617615 & 512 \\
Transformer--SGD & 0.0030821083 & 0.00014953811 & 512 \\
\bottomrule\end{tabular}\end{table}

\paragraph{Optimizer construction and regularization.}
SGD uses the recorded momentum (zero for the selected SGD references). Adam is \texttt{torch.optim.Adam}, including \texttt{fused=True}; Muon updates two-dimensional parameters and uses a companion Adam optimizer for remaining parameters. One-dimensional parameters and names ending in \texttt{bias} are exempt from decay. All other parameters receive the recorded optimizer-specific decay. Adam is not replaced by AdamW; optimizer state and group-specific settings are preserved. Full parameter-group assignments are provided in the reproducibility records.

\paragraph{Reproducibility records.}
Table~\ref{tab:reference_errors} reports fitting and test errors for the representative predictors, together with seed means, standard deviations, and $t$-based $95\%$ intervals for mean error. Every designated run is retained, including loss outliers. These intervals describe variation across fits conditional on the observed data, not uncertainty across financial histories.
\begin{figure}[htbp] 
\centering
\includegraphics[width=.75\linewidth]{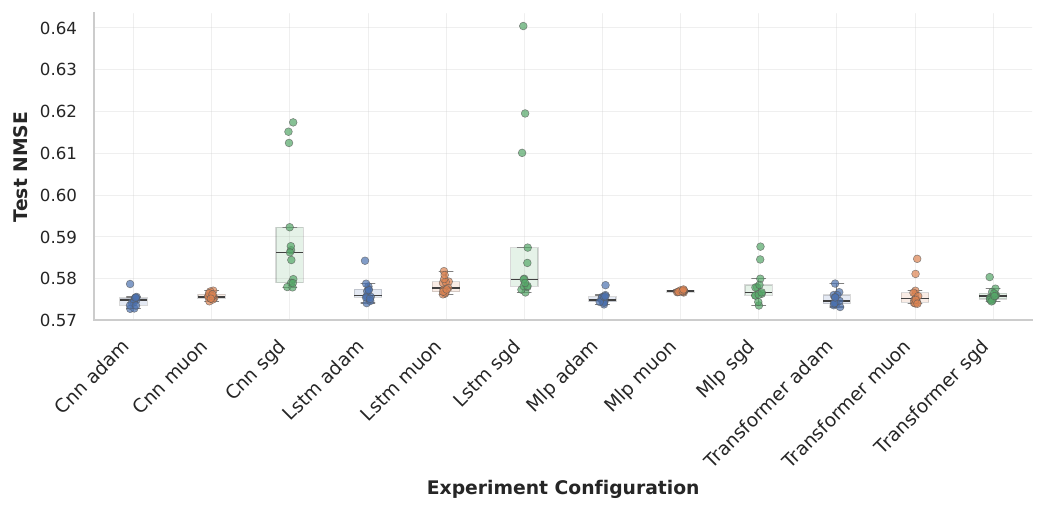}
\caption{\textbf{Repeated-fit error distributions.} All 13 runs per configuration are retained. CNN--SGD and LSTM--SGD exhibit greater run-to-run variability than their adaptive-optimizer counterparts.}
\label{fig:nmse_boxplot}
\end{figure}

\begin{table}[htbp]\centering\small
\caption{Representative fitting and test errors, with repeated-fit uncertainty. Fit MSE uses merged train+validation data. Seed summaries use the same 13 runs as Table~\ref{tab:main_results_nmse}; SD measures run-to-run variation, whereas the 95\% interval concerns the mean NMSE.}\label{tab:reference_errors}
\setlength{\tabcolsep}{3pt}
\begin{tabular}{lrrrrr}\toprule
Configuration & Fit MSE & Test MSE & Test NMSE & Seed mean $\pm$ SD & 95\% CI \\
\midrule
MLP--Adam & 0.4459 & 0.3927 & 0.5742 & $0.5752\pm0.0012$ & [0.5745, 0.5759] \\
MLP--Muon & 0.4501 & 0.3951 & 0.5778 & $0.5770\pm0.0002$ & [0.5769, 0.5771] \\
MLP--SGD & 0.4390 & 0.3943 & 0.5766 & $0.5781\pm0.0040$ & [0.5757, 0.5805] \\
CNN--Adam & 0.4428 & 0.3924 & 0.5738 & $0.5747\pm0.0015$ & [0.5738, 0.5757] \\
CNN--Muon & 0.4330 & 0.3980 & 0.5820 & $0.5758\pm0.0007$ & [0.5753, 0.5762] \\
CNN--SGD & 0.4539 & 0.4004 & 0.5855 & $0.5905\pm0.0147$ & [0.5816, 0.5993] \\
LSTM--Adam & 0.4459 & 0.3938 & 0.5758 & $0.5769\pm0.0026$ & [0.5753, 0.5784] \\
LSTM--Muon & 0.4387 & 0.3971 & 0.5807 & $0.5783\pm0.0018$ & [0.5773, 0.5794] \\
LSTM--SGD & 0.4630 & 0.4081 & 0.5967 & $0.5899\pm0.0203$ & [0.5776, 0.6021] \\
Tr.--Adam & 0.4424 & 0.3930 & 0.5747 & $0.5751\pm0.0016$ & [0.5742, 0.5760] \\
Tr.--Muon & 0.4434 & 0.3941 & 0.5763 & $0.5764\pm0.0032$ & [0.5745, 0.5783] \\
Tr.--SGD & 0.4454 & 0.3933 & 0.5751 & $0.5761\pm0.0015$ & [0.5751, 0.5770] \\
\bottomrule\end{tabular}\end{table}

\subsection{Compute Resources}
Training and evaluation use a GPU cluster, including NVIDIA A100 and H100 GPUs. Detailed hardware and runtime records accompany the reproducibility materials.
\subsection{Numerical Settings for Curvature Measurements}
\label{app:curvature_measurement_settings}
The measurements in Appendix~\ref{app:mechanism} use the recorded native edge-of-stochastic-stability implementation. LOBPCG estimates one leading eigenvalue from 4,096 sampled observations, with at most 100 iterations, relative tolerance 0.005, and the existing eigenvector cache. The directional-curvature diagnostic averages individual mini-batch Rayleigh quotients over at most 1,000 estimates, with minimum-estimate setting 20 and stopping tolerance 0.005. Actual estimate counts and Monte Carlo error summaries are retained in the reproducibility records; these describe diagnostic sampling variability, not variation across training seeds.

The diagnostic batches contain 512 observations for control and LR/10 and 1,024 for batch$\times2$. Measurement uses evaluation mode, including disabled Transformer dropout, and the data-MSE objective without weight decay. The continuation uses the saved optimizer state, recorded parameter groups, one reused shuffle and the final partial batch. Source records retain full-precision learning rates, weight-decay settings and measurement cadence. Faint curves show individual curvature measurements and dark curves show half-epoch medians. Loss panels report epoch final-fit and test MSE.

The displayed reference is $c/\eta$, with $c=2$ for SGD, the gradient-descent stability threshold, and $c=38$ for Adam, the adaptive threshold of \citet{cohen_adaptive_2022}. We plot the same $38/\eta$ line for Muon purely as a visual reference: no corresponding threshold has been established for it. Labels round this value to one decimal place. These reference labels do not rescale the raw-curvature axes. The adaptive stability analysis of \citet{cohen_adaptive_2022} concerns preconditioned curvature, so displaying the reference alongside a raw Hessian does not constitute that stability test. The distinction between the literature batch-sharpness definition and the recorded diagnostic is given in Appendix~\ref{app:spikes}.

\section{Robustness of the Leaderboard Tie}
\label{app:robustness-tie}
\label{app:welch_tests}

For each pair let $\delta=\mu_B-\mu_A$ be the difference in mean test NMSE across the two 13-run cohorts. Welch's two-sided test addresses $H_0:\delta=0$. We apply Holm correction to all 66 comparisons as one family. There are 35 raw rejections and seven corrected rejections at $0.05$; none of the corrected rejections involves SGD. Larger variability can reduce power to detect a mean difference, so nonrejection must not be interpreted as equivalence.

TOST instead tests $H_0:\delta\leq-\Delta\ \text{or}\ \delta\geq\Delta$ against $H_1:-\Delta<\delta<\Delta$ \citep{schuirmann1987}. With Welch standard error $s=\sqrt{s_A^2/n_A+s_B^2/n_B}$ and Welch--Satterthwaite degrees of freedom, the two one-sided statistics are $(\widehat\delta+\Delta)/s$ and $(\widehat\delta-\Delta)/s$. The pair-level $p$-value is the larger of the two one-sided $p$-values. Holm correction is then applied across 66 pair-level tests separately at each margin.

The five margins, $0.001,0.002,0.003,0.005,0.010$, are exploratory sensitivity values. The five families are not jointly corrected; selecting a successful margin after inspection is not a prespecified practical-equivalence conclusion. At these margins the corrected counts are respectively 0, 1, 10, 31, and 45. MLP Adam--Muon illustrates the distinction: equality is rejected, while equivalence passes at $0.005$ but not $0.003$ after correction. These tests follow the independent-cohort Welch analysis; they quantify run variability conditional on the observed test set, not sampling uncertainty over dependent dates/assets.

\begingroup\footnotesize
\begin{longtable}{llrrr}
\caption{All 66 Welch tests. Difference is mean(B) minus mean(A). Tr. denotes Transformer.}\label{tab:welch_full}\\
\toprule
A & B & $\widehat\delta$ & Raw $p$ & Holm $p$ \\
\midrule
\endfirsthead
\toprule
A & B & $\widehat\delta$ & Raw $p$ & Holm $p$ \\
\midrule
\endhead
\bottomrule
\endfoot
CNN--Adam & LSTM--Muon & +0.00362 & 0.0000 & 0.0007 \\
LSTM--Muon & MLP--Adam & -0.00312 & 0.0000 & 0.0020 \\
LSTM--Muon & Tr.--Adam & -0.00325 & 0.0000 & 0.0031 \\
CNN--Muon & MLP--Muon & +0.00122 & 0.0001 & 0.0037 \\
MLP--Adam & MLP--Muon & +0.00176 & 0.0001 & 0.0090 \\
CNN--Muon & LSTM--Muon & +0.00258 & 0.0002 & 0.0109 \\
CNN--Adam & MLP--Muon & +0.00225 & 0.0002 & 0.0112 \\
MLP--Muon & Tr.--Adam & -0.00189 & 0.0009 & 0.0538 \\
LSTM--Muon & Tr.--SGD & -0.00227 & 0.0020 & 0.1168 \\
CNN--Adam & CNN--SGD & +0.01574 & 0.0022 & 0.1276 \\
CNN--SGD & Tr.--Adam & -0.01537 & 0.0026 & 0.1477 \\
CNN--SGD & MLP--Adam & -0.01524 & 0.0028 & 0.1537 \\
CNN--Muon & CNN--SGD & +0.01470 & 0.0036 & 0.1923 \\
CNN--SGD & Tr.--SGD & -0.01439 & 0.0041 & 0.2184 \\
CNN--SGD & Tr.--Muon & -0.01409 & 0.0048 & 0.2516 \\
CNN--SGD & LSTM--Adam & -0.01360 & 0.0060 & 0.3057 \\
CNN--SGD & MLP--Muon & -0.01348 & 0.0062 & 0.3093 \\
CNN--Adam & MLP--SGD & +0.00339 & 0.0112 & 0.5477 \\
CNN--SGD & MLP--SGD & -0.01234 & 0.0112 & 0.5477 \\
CNN--SGD & LSTM--Muon & -0.01212 & 0.0117 & 0.5486 \\
LSTM--Muon & MLP--Muon & -0.00136 & 0.0172 & 0.7915 \\
CNN--Adam & LSTM--Adam & +0.00214 & 0.0196 & 0.8832 \\
CNN--Adam & LSTM--SGD & +0.01514 & 0.0199 & 0.8832 \\
MLP--SGD & Tr.--Adam & -0.00303 & 0.0211 & 0.9057 \\
LSTM--SGD & Tr.--Adam & -0.01477 & 0.0224 & 0.9398 \\
LSTM--SGD & MLP--Adam & -0.01464 & 0.0233 & 0.9550 \\
MLP--Adam & MLP--SGD & +0.00290 & 0.0240 & 0.9595 \\
LSTM--SGD & MLP--Muon & -0.01288 & 0.0411 & 1.0000 \\
LSTM--Muon & Tr.--Muon & -0.00197 & 0.0650 & 1.0000 \\
LSTM--SGD & Tr.--Muon & -0.01348 & 0.0348 & 1.0000 \\
LSTM--SGD & MLP--SGD & -0.01174 & 0.0616 & 1.0000 \\
CNN--Adam & CNN--Muon & +0.00103 & 0.0433 & 1.0000 \\
MLP--Adam & Tr.--Muon & +0.00116 & 0.2335 & 1.0000 \\
MLP--Adam & Tr.--SGD & +0.00085 & 0.1266 & 1.0000 \\
MLP--Muon & MLP--SGD & +0.00114 & 0.3206 & 1.0000 \\
MLP--Muon & Tr.--Muon & -0.00060 & 0.5052 & 1.0000 \\
MLP--Muon & Tr.--SGD & -0.00090 & 0.0579 & 1.0000 \\
MLP--SGD & Tr.--Muon & -0.00174 & 0.2275 & 1.0000 \\
MLP--SGD & Tr.--SGD & -0.00204 & 0.1029 & 1.0000 \\
Tr.--Adam & Tr.--Muon & +0.00129 & 0.2038 & 1.0000 \\
LSTM--SGD & Tr.--SGD & -0.01379 & 0.0309 & 1.0000 \\
MLP--Adam & Tr.--Adam & -0.00013 & 0.8116 & 1.0000 \\
LSTM--Adam & MLP--Adam & -0.00164 & 0.0543 & 1.0000 \\
LSTM--Muon & LSTM--SGD & +0.01152 & 0.0639 & 1.0000 \\
CNN--Adam & MLP--Adam & +0.00049 & 0.3660 & 1.0000 \\
CNN--Adam & Tr.--Adam & +0.00036 & 0.5547 & 1.0000 \\
CNN--Adam & Tr.--Muon & +0.00165 & 0.1073 & 1.0000 \\
CNN--Adam & Tr.--SGD & +0.00135 & 0.0355 & 1.0000 \\
CNN--Muon & LSTM--Adam & +0.00111 & 0.1647 & 1.0000 \\
CNN--Muon & LSTM--SGD & +0.01410 & 0.0277 & 1.0000 \\
CNN--Muon & MLP--Adam & -0.00054 & 0.1764 & 1.0000 \\
CNN--Muon & MLP--SGD & +0.00236 & 0.0551 & 1.0000 \\
CNN--Muon & Tr.--Adam & -0.00067 & 0.1805 & 1.0000 \\
LSTM--Muon & MLP--SGD & -0.00022 & 0.8544 & 1.0000 \\
CNN--Muon & Tr.--Muon & +0.00062 & 0.5033 & 1.0000 \\
CNN--SGD & LSTM--SGD & -0.00060 & 0.9317 & 1.0000 \\
LSTM--Adam & LSTM--Muon & +0.00148 & 0.1063 & 1.0000 \\
LSTM--Adam & LSTM--SGD & +0.01300 & 0.0403 & 1.0000 \\
Tr.--Adam & Tr.--SGD & +0.00098 & 0.1194 & 1.0000 \\
LSTM--Adam & MLP--Muon & +0.00011 & 0.8778 & 1.0000 \\
LSTM--Adam & MLP--SGD & +0.00125 & 0.3518 & 1.0000 \\
LSTM--Adam & Tr.--Adam & -0.00178 & 0.0487 & 1.0000 \\
LSTM--Adam & Tr.--Muon & -0.00049 & 0.6717 & 1.0000 \\
LSTM--Adam & Tr.--SGD & -0.00079 & 0.3597 & 1.0000 \\
CNN--Muon & Tr.--SGD & +0.00032 & 0.5171 & 1.0000 \\
Tr.--Muon & Tr.--SGD & -0.00030 & 0.7595 & 1.0000 \\
\end{longtable}
\endgroup
\begingroup\footnotesize
\begin{longtable}{llrrrrr}
\caption{All 66 TOST comparisons at the five exploratory margins: Holm-adjusted pair-level $p$-values. Values below .05 support equivalence for that margin. Unadjusted values and mean differences are retained in the complete accompanying CSV.}\label{tab:tost_full}\\
\toprule
A & B & .001 & .002 & .003 & .005 & .010 \\
\midrule
\endfirsthead
\toprule
A & B & .001 & .002 & .003 & .005 & .010 \\
\midrule
\endhead
\bottomrule
\endfoot
CNN--Adam & CNN--Muon & 1.0000 & 1.0000 & 0.0185 & 0.0000 & $<10^{-6}$ \\
CNN--Adam & CNN--SGD & 1.0000 & 1.0000 & 1.0000 & 1.0000 & 1.0000 \\
CNN--Adam & LSTM--Adam & 1.0000 & 1.0000 & 1.0000 & 0.0512 & $<10^{-6}$ \\
CNN--Adam & LSTM--Muon & 1.0000 & 1.0000 & 1.0000 & 0.5592 & $<10^{-6}$ \\
CNN--Adam & LSTM--SGD & 1.0000 & 1.0000 & 1.0000 & 1.0000 & 1.0000 \\
CNN--Adam & MLP--Adam & 1.0000 & 0.3150 & 0.0035 & $<10^{-6}$ & $<10^{-6}$ \\
CNN--Adam & MLP--Muon & 1.0000 & 1.0000 & 1.0000 & 0.0007 & $<10^{-6}$ \\
CNN--Adam & MLP--SGD & 1.0000 & 1.0000 & 1.0000 & 1.0000 & 0.0005 \\
CNN--Adam & Tr.--Adam & 1.0000 & 0.3895 & 0.0067 & 0.0000 & $<10^{-6}$ \\
CNN--Adam & Tr.--Muon & 1.0000 & 1.0000 & 1.0000 & 0.0514 & 0.0000 \\
CNN--Adam & Tr.--SGD & 1.0000 & 1.0000 & 0.3054 & 0.0001 & $<10^{-6}$ \\
CNN--Muon & CNN--SGD & 1.0000 & 1.0000 & 1.0000 & 1.0000 & 1.0000 \\
CNN--Muon & LSTM--Adam & 1.0000 & 1.0000 & 0.5846 & 0.0035 & $<10^{-6}$ \\
CNN--Muon & LSTM--Muon & 1.0000 & 1.0000 & 1.0000 & 0.0075 & $<10^{-6}$ \\
CNN--Muon & LSTM--SGD & 1.0000 & 1.0000 & 1.0000 & 1.0000 & 1.0000 \\
CNN--Muon & MLP--Adam & 1.0000 & 0.0365 & 0.0001 & $<10^{-6}$ & $<10^{-6}$ \\
CNN--Muon & MLP--Muon & 1.0000 & 0.0859 & 0.0000 & $<10^{-6}$ & $<10^{-6}$ \\
CNN--Muon & MLP--SGD & 1.0000 & 1.0000 & 1.0000 & 0.4514 & 0.0002 \\
CNN--Muon & Tr.--Adam & 1.0000 & 0.3895 & 0.0044 & 0.0000 & $<10^{-6}$ \\
CNN--Muon & Tr.--Muon & 1.0000 & 1.0000 & 0.4705 & 0.0064 & 0.0000 \\
CNN--Muon & Tr.--SGD & 1.0000 & 0.0795 & 0.0009 & $<10^{-6}$ & $<10^{-6}$ \\
CNN--SGD & LSTM--Adam & 1.0000 & 1.0000 & 1.0000 & 1.0000 & 1.0000 \\
CNN--SGD & LSTM--Muon & 1.0000 & 1.0000 & 1.0000 & 1.0000 & 1.0000 \\
CNN--SGD & LSTM--SGD & 1.0000 & 1.0000 & 1.0000 & 1.0000 & 1.0000 \\
CNN--SGD & MLP--Adam & 1.0000 & 1.0000 & 1.0000 & 1.0000 & 1.0000 \\
CNN--SGD & MLP--Muon & 1.0000 & 1.0000 & 1.0000 & 1.0000 & 1.0000 \\
CNN--SGD & MLP--SGD & 1.0000 & 1.0000 & 1.0000 & 1.0000 & 1.0000 \\
CNN--SGD & Tr.--Adam & 1.0000 & 1.0000 & 1.0000 & 1.0000 & 1.0000 \\
CNN--SGD & Tr.--Muon & 1.0000 & 1.0000 & 1.0000 & 1.0000 & 1.0000 \\
CNN--SGD & Tr.--SGD & 1.0000 & 1.0000 & 1.0000 & 1.0000 & 1.0000 \\
LSTM--Adam & LSTM--Muon & 1.0000 & 1.0000 & 1.0000 & 0.0124 & $<10^{-6}$ \\
LSTM--Adam & LSTM--SGD & 1.0000 & 1.0000 & 1.0000 & 1.0000 & 1.0000 \\
LSTM--Adam & MLP--Adam & 1.0000 & 1.0000 & 1.0000 & 0.0124 & $<10^{-6}$ \\
LSTM--Adam & MLP--Muon & 1.0000 & 0.6874 & 0.0510 & 0.0005 & $<10^{-6}$ \\
LSTM--Adam & MLP--SGD & 1.0000 & 1.0000 & 1.0000 & 0.1464 & 0.0000 \\
LSTM--Adam & Tr.--Adam & 1.0000 & 1.0000 & 1.0000 & 0.0204 & $<10^{-6}$ \\
LSTM--Adam & Tr.--Muon & 1.0000 & 1.0000 & 0.8150 & 0.0124 & $<10^{-6}$ \\
LSTM--Adam & Tr.--SGD & 1.0000 & 1.0000 & 0.4130 & 0.0018 & $<10^{-6}$ \\
LSTM--Muon & LSTM--SGD & 1.0000 & 1.0000 & 1.0000 & 1.0000 & 1.0000 \\
LSTM--Muon & MLP--Adam & 1.0000 & 1.0000 & 1.0000 & 0.0728 & $<10^{-6}$ \\
LSTM--Muon & MLP--Muon & 1.0000 & 1.0000 & 0.1634 & 0.0002 & $<10^{-6}$ \\
LSTM--Muon & MLP--SGD & 1.0000 & 1.0000 & 0.7739 & 0.0197 & 0.0000 \\
LSTM--Muon & Tr.--Adam & 1.0000 & 1.0000 & 1.0000 & 0.1992 & $<10^{-6}$ \\
LSTM--Muon & Tr.--Muon & 1.0000 & 1.0000 & 1.0000 & 0.1087 & 0.0000 \\
LSTM--Muon & Tr.--SGD & 1.0000 & 1.0000 & 1.0000 & 0.0075 & $<10^{-6}$ \\
LSTM--SGD & MLP--Adam & 1.0000 & 1.0000 & 1.0000 & 1.0000 & 1.0000 \\
LSTM--SGD & MLP--Muon & 1.0000 & 1.0000 & 1.0000 & 1.0000 & 1.0000 \\
LSTM--SGD & MLP--SGD & 1.0000 & 1.0000 & 1.0000 & 1.0000 & 1.0000 \\
LSTM--SGD & Tr.--Adam & 1.0000 & 1.0000 & 1.0000 & 1.0000 & 1.0000 \\
LSTM--SGD & Tr.--Muon & 1.0000 & 1.0000 & 1.0000 & 1.0000 & 1.0000 \\
LSTM--SGD & Tr.--SGD & 1.0000 & 1.0000 & 1.0000 & 1.0000 & 1.0000 \\
MLP--Adam & MLP--Muon & 1.0000 & 1.0000 & 0.0671 & 0.0000 & $<10^{-6}$ \\
MLP--Adam & MLP--SGD & 1.0000 & 1.0000 & 1.0000 & 1.0000 & 0.0003 \\
MLP--Adam & Tr.--Adam & 1.0000 & 0.0725 & 0.0008 & $<10^{-6}$ & $<10^{-6}$ \\
MLP--Adam & Tr.--Muon & 1.0000 & 1.0000 & 1.0000 & 0.0171 & 0.0000 \\
MLP--Adam & Tr.--SGD & 1.0000 & 1.0000 & 0.0177 & 0.0000 & $<10^{-6}$ \\
MLP--Muon & MLP--SGD & 1.0000 & 1.0000 & 1.0000 & 0.0708 & 0.0000 \\
MLP--Muon & Tr.--Adam & 1.0000 & 1.0000 & 0.5846 & 0.0003 & $<10^{-6}$ \\
MLP--Muon & Tr.--Muon & 1.0000 & 1.0000 & 0.4383 & 0.0066 & 0.0000 \\
MLP--Muon & Tr.--SGD & 1.0000 & 0.7499 & 0.0109 & 0.0000 & $<10^{-6}$ \\
MLP--SGD & Tr.--Adam & 1.0000 & 1.0000 & 1.0000 & 1.0000 & 0.0003 \\
MLP--SGD & Tr.--Muon & 1.0000 & 1.0000 & 1.0000 & 0.4000 & 0.0001 \\
MLP--SGD & Tr.--SGD & 1.0000 & 1.0000 & 1.0000 & 0.3330 & 0.0001 \\
Tr.--Adam & Tr.--Muon & 1.0000 & 1.0000 & 1.0000 & 0.0242 & 0.0000 \\
Tr.--Adam & Tr.--SGD & 1.0000 & 1.0000 & 0.0803 & 0.0000 & $<10^{-6}$ \\
Tr.--Muon & Tr.--SGD & 1.0000 & 1.0000 & 0.3287 & 0.0035 & $<10^{-6}$ \\
\end{longtable}
\endgroup

\section{Functional Divergence Analysis}
\label{app:functional_divergence_analysis}

\subsection{Distances, Coverage, and Uncertainty}
\label{app:functional_distance}
Both $D^2$ and $D$ in Eq.~\eqref{eq:function_distance} are saved for all $\binom{168}{2}=14{,}028$ model pairs, with representatives and repeated runs (13 seeds) identified separately. 
Matrices below summarize individual RMS distances computed on the test set.
Seed-matrix diagonals exclude self-pairs; off-diagonals include all $13\times13$ cross-configuration pairs.

For optimizer cohorts $A$ and $B$, define within-cohort means $W_A,W_B$ over distinct fitted runs and cross-cohort mean $C$ over all pairs. 
We display all three separately, together with the ratio $R$ of Eq.~\eqref{eq:distance_ratio} and the stricter $R^{\max}=C/\max(W_A,W_B)$. 

To estimate uncertainty, delete a fitted run, not an individual pairwise distance: many distances share the same checkpoint. 
Where seeds were shared, delete the shared seed block from both cohorts; partial overlaps retain shared blocks and singleton blocks. 
For independent cohorts, add the two delete-one jackknife variance contributions and use Welch--Satterthwaite degrees of freedom. 
Shared-block intervals use $K-1$ degrees of freedom, with $K$ the number of seed blocks. 
Ratio intervals are formed on the log scale. 
These pointwise intervals are exploratory, conditional on this test panel, and not corrected across comparisons.

\begin{table}[htbp]
\centering\small
\caption{Within-architecture repeated-fit distances. $R$ divides $C$ by $(W_A+W_B)/2$; $R^{\max}$ divides it by $\max(W_A,W_B)$ and is therefore the stricter comparison. Optimizer-specific fitting budgets are part of each recorded configuration. Full component intervals are in the accompanying table.}\label{tab:distance_uncertainty}
\begin{tabular}{lrrrrr}
\toprule
Cohorts A / B & $W_A$ & $W_B$ & $C$ & $R$ [95\% interval] & $R^{\max}$ [95\% interval] \\
\midrule
MLP--Adam / Muon & 0.0400 & 0.0119 & 0.0539 & $2.08\;[1.81,2.39]$ & $1.35\;[1.13,1.61]$ \\
MLP--Adam / SGD & 0.0400 & 0.0749 & 0.0764 & $1.33\;[1.23,1.44]$ & $1.02\;[0.87,1.20]$ \\
MLP--Muon / SGD & 0.0119 & 0.0749 & 0.0843 & $1.94\;[1.70,2.22]$ & $1.12\;[0.94,1.34]$ \\
CNN--Adam / Muon & 0.0475 & 0.0364 & 0.0580 & $1.38\;[1.29,1.48]$ & $1.22\;[1.05,1.42]$ \\
CNN--Adam / SGD & 0.0475 & 0.1032 & 0.1015 & $1.35\;[1.14,1.59]$ & $0.98\;[0.80,1.22]$ \\
CNN--Muon / SGD & 0.0364 & 0.1032 & 0.1067 & $1.53\;[1.27,1.84]$ & $1.03\;[0.84,1.27]$ \\
LSTM--Adam / Muon & 0.0477 & 0.0519 & 0.0693 & $1.39\;[1.24,1.56]$ & $1.33\;[1.18,1.51]$ \\
LSTM--Adam / SGD & 0.0477 & 0.1073 & 0.0899 & $1.16\;[0.92,1.46]$ & $0.84\;[0.60,1.18]$ \\
LSTM--Muon / SGD & 0.0519 & 0.1073 & 0.1007 & $1.26\;[1.08,1.47]$ & $0.94\;[0.67,1.32]$ \\
Tr.--Adam / Muon & 0.0461 & 0.0505 & 0.0545 & $1.13\;[1.04,1.22]$ & $1.08\;[0.89,1.31]$ \\
Tr.--Adam / SGD & 0.0461 & 0.0478 & 0.0622 & $1.33\;[1.18,1.49]$ & $1.30\;[1.09,1.56]$ \\
Tr.--Muon / SGD & 0.0505 & 0.0478 & 0.0567 & $1.15\;[1.06,1.26]$ & $1.12\;[0.99,1.27]$ \\
\bottomrule
\end{tabular}
\end{table}
\begin{figure}[htbp] 
\centering
\includegraphics[width=\linewidth]{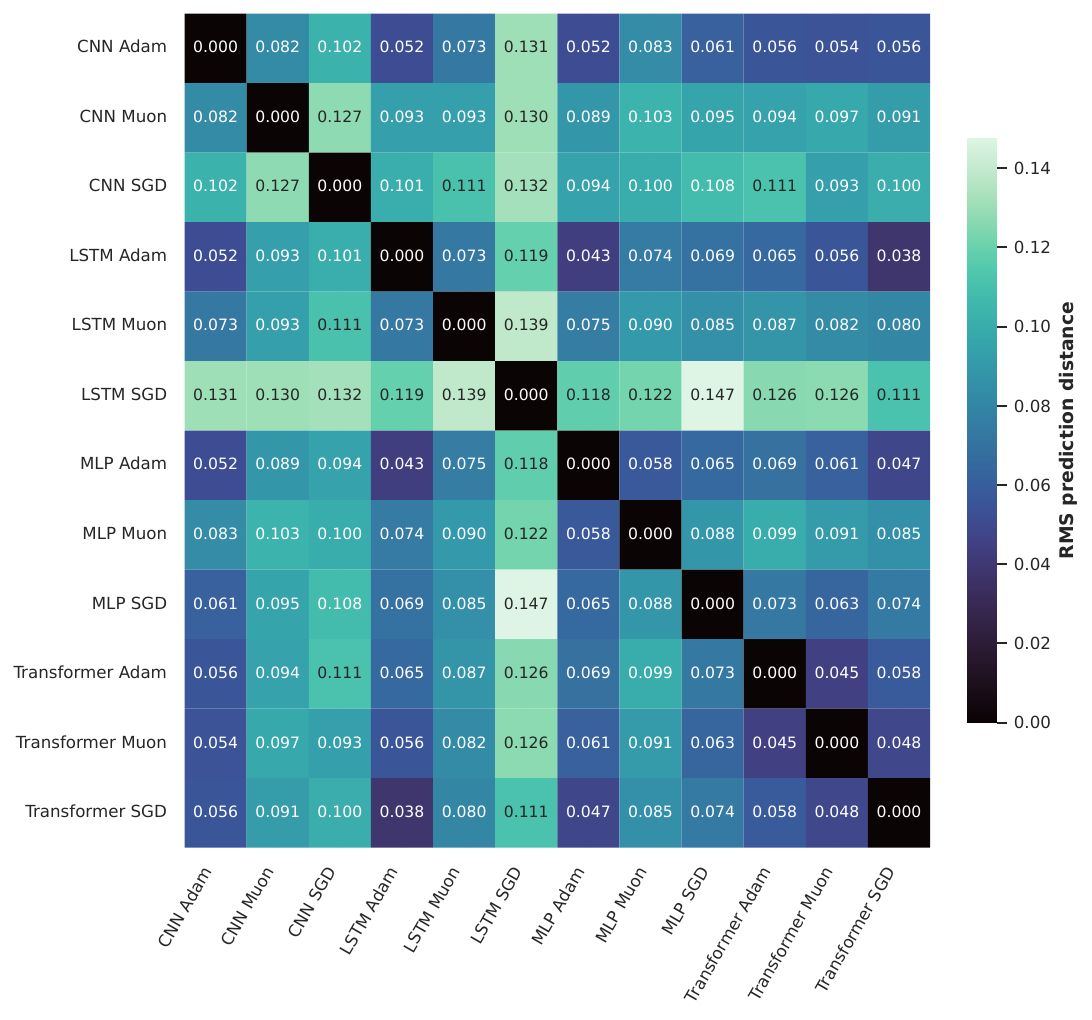}
\caption{Representative RMS prediction distances on all test observations. Diagonal self-distances are zero.}
\label{fig:distance_matrix}
\end{figure}
\begin{figure}[htbp] 
\centering
\includegraphics[width=\linewidth]{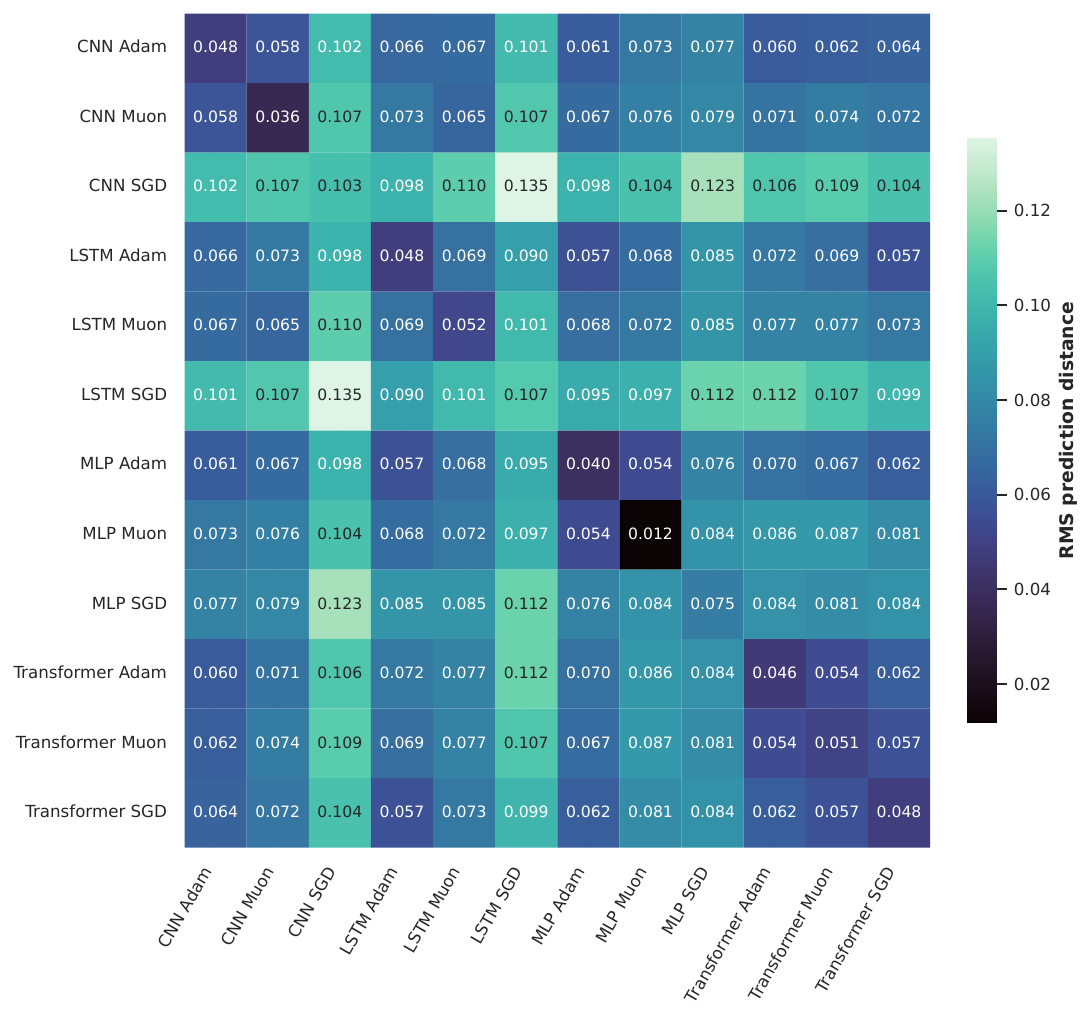}
\caption{Mean individual RMS distances among repeated-fit cohorts. Diagonals compare different runs of the same configuration; nonzero diagonal values are expected.}
\label{fig:seed_distance_matrix}
\end{figure}
\begin{figure}[htbp]\centering
\includegraphics[width=\linewidth]{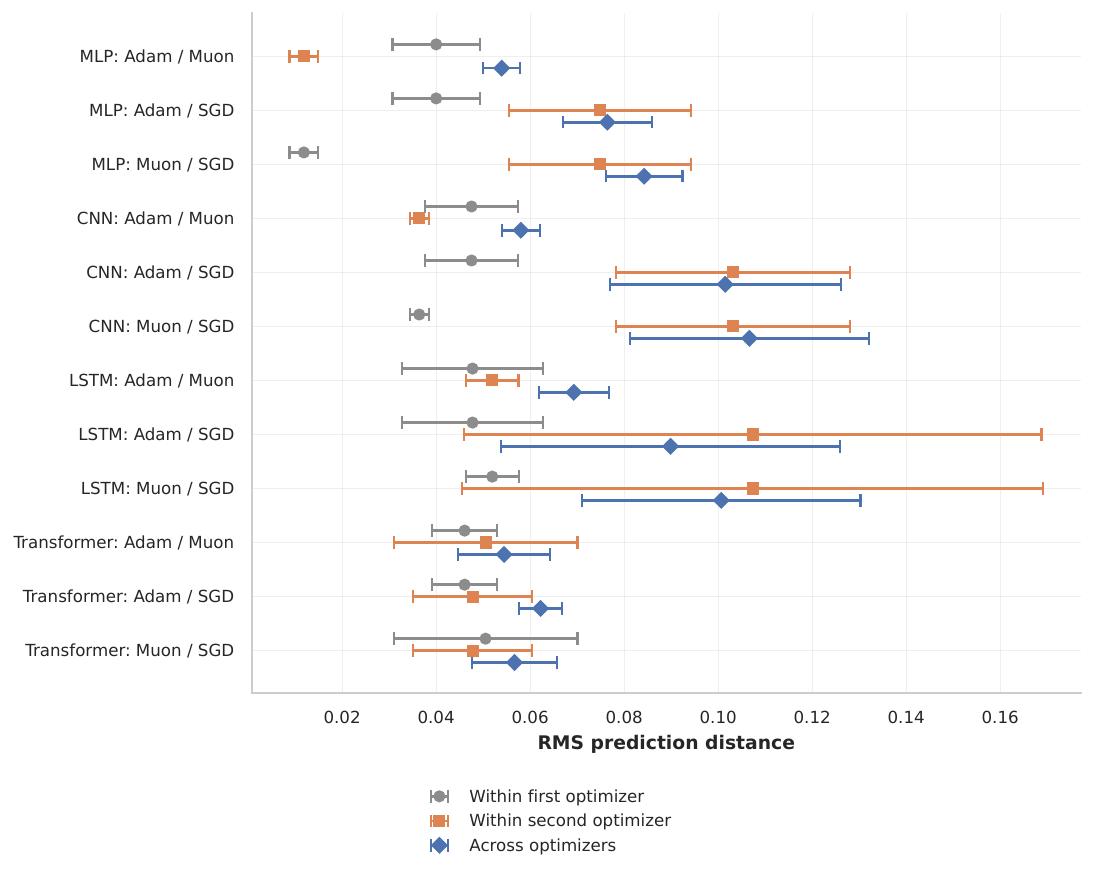}
\caption{\textbf{The components behind functional separation.} Both within-optimizer levels and the across-optimizer level are shown separately for all within-architecture comparisons. Bars are pointwise 95\% seed-level jackknife intervals. Across-optimizer disagreement need not exceed both within-optimizer levels.}
\label{fig:all_repeated_distances}\end{figure}

\begin{figure}[tbp]\centering
\begin{subfigure}{.32\linewidth}\centering
\includegraphics[width=\linewidth]{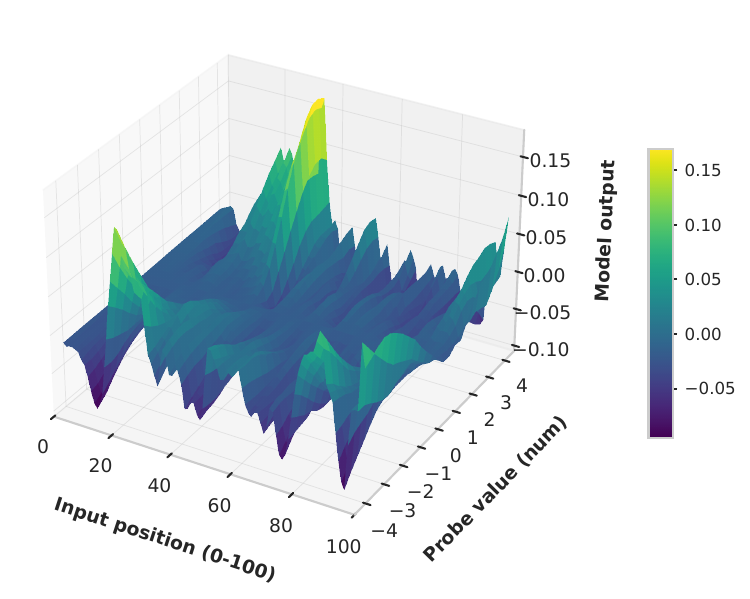}
\caption{CNN}\end{subfigure}\hfill
\begin{subfigure}{.32\linewidth}\centering
\includegraphics[width=\linewidth]{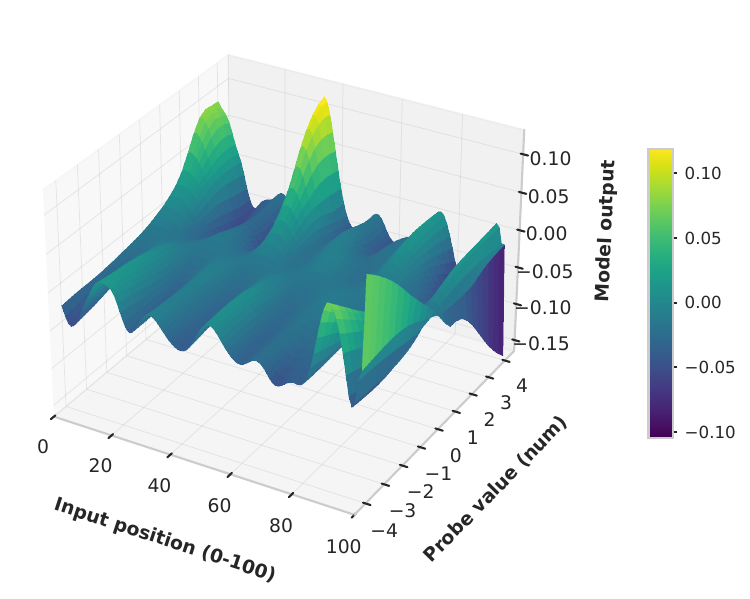}
\caption{LSTM}\end{subfigure}\hfill
\begin{subfigure}{.32\linewidth}\centering
\includegraphics[width=\linewidth]{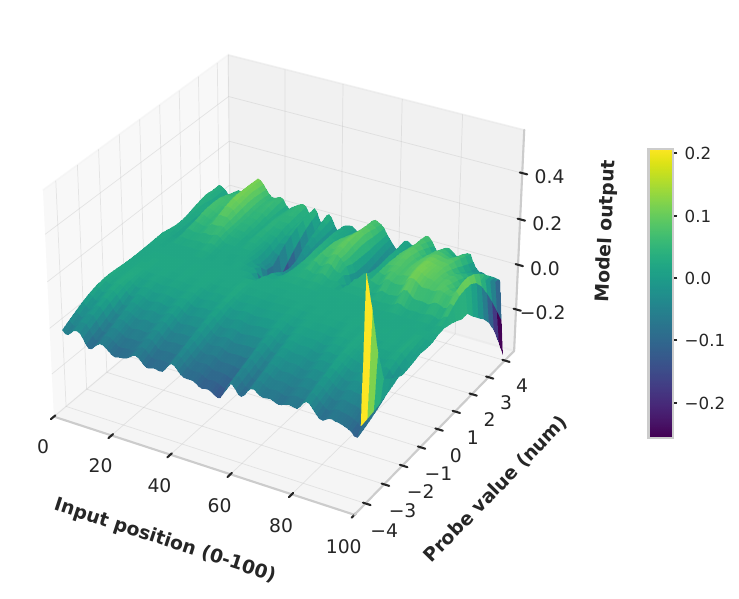}
\caption{Transformer}\label{fig:difference_adam_muon_transformer}\end{subfigure}\hfill
\caption{Muon minus Adam response surfaces for the selected configurations, on the common position/probe grid. Colorbars give standardized output differences.}\label{fig:all_difference_surfaces}\end{figure}

\FloatBarrier
\subsection{Paired Initialization and Data Order}
\label{app:paired_initialization}
We repeat each of the twelve architecture--optimizer configurations from three independently sampled initializations, pairing the initial weights across optimizers within each architecture. All 36 runs completed their intended budgets. Hashes of the loaded initial tensors and data permutation verify the matching. The data order is fixed across runs and reused across epochs; batch size is 512, including the final partial batch. Transformer dropout uses a paired random-number seed. All runs minimize MSE on the same merged training and validation observations; the loss function is not changed between optimizers. Test loss does not select a checkpoint.

Learning rates, weight decay and parameter groups match Table~\ref{tab:configuration} and remain optimizer-specific, so this experiment controls initialization and data order, not regularization or compute. Budgets are 50 epochs except the Transformer (25) and CNN--SGD (7). The LSTM--SGD run uses learning rate $0.0984626751$ and decay $0.0197855670$, a separate configuration from the seven-epoch LSTM--SGD reference of Table~\ref{tab:main_results_nmse}. These runs neither replace nor augment the 13-run cohorts.

\begin{figure}[tbp]\centering
\includegraphics[width=\linewidth]{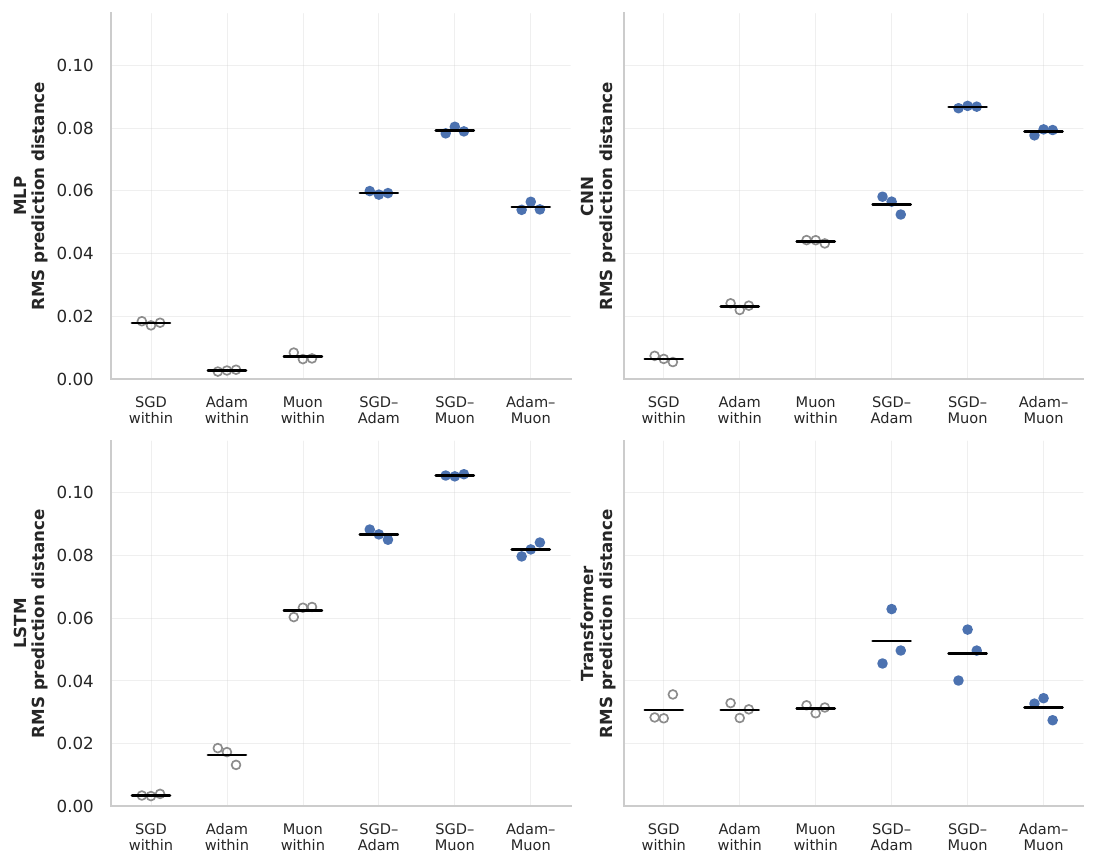}
\caption{\textbf{Functional differences under paired initialization.} Within each architecture, the first three groups compare different initializations under the same optimizer configuration (open points); the last three compare optimizers from identical initial weights (filled points). Each group contains three distances; black bars mark their means, not confidence intervals. Data order is common, while tuned optimizer settings and budgets are retained.}
\label{fig:paired_initialization}\end{figure}

For each optimizer pair within an architecture, we compare functions with the same initialization:
\begin{equation}
C^{\mathrm{paired}}_{AB}=\frac{1}{3}\sum_{s=1}^{3}D(f_{A,s},f_{B,s}).
\end{equation}
Within-optimizer distances average the three distinct initialization pairs. This paired quantity differs from $C_{AB}$ in Eq.~\eqref{eq:within_across}, which averages all cross-cohort pairs. Predictions use the same 564,795 test observations and common standardized log-variance scale. Figure~\ref{fig:paired_initialization} shows individual distances and their means; shared-model pairs are not independent observations, and three initializations do not provide a precise population uncertainty estimate.

Forecasting error and prediction disagreement become directly comparable once both are expressed as root-mean-square quantities on the common standardized log-variance scale. Writing $\mathrm{RMSE}=\sqrt{\mathrm{NMSE}\cdot\operatorname{Var}(y)}$ with $\operatorname{Var}(y)=0.6839$ on the test panel, every fitted model here has $\mathrm{RMSE}\approx0.627$. Table~\ref{tab:paired_scale} places the two quantities side by side: $|\Delta\mathrm{RMSE}|$ is how far apart two models' error magnitudes are, and $D$ is how far apart their predictions are, in the same units.

The separation is systematic. Across all twelve comparisons the paired cross-optimizer distance exceeds \emph{both} within-optimizer distances, not merely their average, by factors of $1.0$ to $7.8$; eleven of the twelve exceed them by at least $1.3$. The same strict comparison holds for nine of twelve seed-cohort pairs in Table~\ref{tab:distance_uncertainty} on point estimates, and for four once their seed-level intervals are required to exclude one; holding initialization and data order fixed is what sharpens it. Two models that start from identical weights, see identical data in identical order, and differ only in optimizer end up further apart than two runs of the same optimizer started from different weights.

The magnitudes make the same point on the error scale. Predictions differ by $5.0$--$16.7\%$ of the typical forecast error, while the models' error magnitudes differ by $0.003$--$0.62\%$ of it: $D$ exceeds $|\Delta\mathrm{RMSE}|$ by factors of $21$ to $2859$. MLP SGD--Adam is the extreme case, with error magnitudes matching to $2\times10^{-5}$ while predictions sit $0.059$ apart. Equivalently, the two fitted models' residuals correlate at $\rho=0.986$--$0.999$, and the residual $1-\rho$ is what the loss cannot see. Initialization noise does not account for this: it is the smaller effect in every comparison.

\begin{table}[tbp]\centering\small
\caption{\textbf{Loss gap versus functional gap, on one scale.} Paired-initialization runs; all quantities are RMS in standardized log-variance units, where the typical forecast error is $\mathrm{RMSE}\approx0.627$. $|\Delta\mathrm{RMSE}|$ compares the two models' accuracy; $D$ compares their predictions; $W^{\max}$ is the larger of the two within-optimizer distances from different initializations, i.e.\ the initialization-noise baseline.}\label{tab:paired_scale}
\begin{tabular}{llrrrrr}\toprule
Architecture & Pair & $|\Delta\mathrm{RMSE}|$ & $D$ & $W^{\max}$ & $D/|\Delta\mathrm{RMSE}|$ & $D/W^{\max}$\\\midrule
MLP & SGD/Adam & 0.00002 & 0.0593 & 0.0178 & 2859 & 3.34 \\
MLP & SGD/Muon & 0.00155 & 0.0792 & 0.0178 & 51 & 4.46 \\
MLP & Adam/Muon & 0.00153 & 0.0548 & 0.0070 & 36 & 7.77 \\
CNN & SGD/Adam & 0.00102 & 0.0556 & 0.0231 & 55 & 2.41 \\
CNN & SGD/Muon & 0.00282 & 0.0867 & 0.0439 & 31 & 1.98 \\
CNN & Adam/Muon & 0.00384 & 0.0788 & 0.0439 & 21 & 1.80 \\
LSTM & SGD/Adam & 0.00390 & 0.0866 & 0.0163 & 22 & 5.30 \\
LSTM & SGD/Muon & 0.00038 & 0.1055 & 0.0624 & 278 & 1.69 \\
LSTM & Adam/Muon & 0.00352 & 0.0819 & 0.0624 & 23 & 1.31 \\
Transf. & SGD/Adam & 0.00047 & 0.0527 & 0.0307 & 112 & 1.72 \\
Transf. & SGD/Muon & 0.00071 & 0.0487 & 0.0311 & 68 & 1.56 \\
Transf. & Adam/Muon & 0.00025 & 0.0315 & 0.0311 & 128 & 1.01 \\
\bottomrule\end{tabular}\end{table}

\begin{table}[tbp]\centering\small
\caption{Final fit MSE across the three paired initializations (mean $\pm$ sample SD). Test errors are summarized in Table~\ref{tab:paired_scale}.}\label{tab:paired_fit_summary}
\begin{tabular}{lrrr}\toprule Architecture & SGD & Adam & Muon\\\midrule
MLP & $0.438831\pm0.000100$ & $0.447133\pm0.000042$ & $0.450316\pm0.000340$ \\
CNN & $0.446792\pm0.000118$ & $0.444018\pm0.000220$ & $0.433788\pm0.000042$ \\
LSTM & $0.454815\pm0.000095$ & $0.446676\pm0.000270$ & $0.439183\pm0.000537$ \\
Transformer & $0.444912\pm0.000413$ & $0.442442\pm0.000323$ & $0.443362\pm0.000176$ \\
\bottomrule\end{tabular}\end{table}

\FloatBarrier
\FloatBarrier
\subsection{Attribution Estimators and Checks}
\label{app:shap_protocol}
All attribution comparisons use correctly merged-scaled inputs and fixed background/explanation samples shared within the comparison. We report each run's normalized mean absolute attribution, then take the median and IQR across the 13 runs at each lag. This preserves the distinction between seed spread and a confidence interval. A logarithmic vertical axis exposes smaller lag contributions. The sampling-based SHAP estimates can include exact zeros. Zero-valued medians and quartile bounds cannot be represented on this axis and are masked, producing gaps in the curves or shaded bands. These gaps are display artifacts, not missing observations; the stored attribution values are unchanged.

For MLP/CNN, the retained DeepSHAP protocol uses 300 background and 800 explanation observations for repeated fits, and 1,000/2,000 for representatives. LSTM uses KernelSHAP with 25 background centroids and 150 explanation observations for repeated fits, and 15/100 for representatives (250 coalition samples in the representative protocol). Settings and reconstruction checks are recorded per artifact. Transformer uses KernelSHAP throughout, with no mixture of DeepSHAP and KernelSHAP within its comparisons. The replacement protocol uses 15 weighted merged-training background centroids, 200 shared reproducibly sampled test observations, all 100 features, 2,048 coalition samples, and no feature-selection penalty.

Transformer representative estimates were repeated with an independent coalition-sampling seed and at 4,096 samples. The diagnostic targets were normalized-profile L1 difference at most 0.05 and at least eight shared top-ten features. All three passed at 2,048: the repeat/budget L1 differences were 0.0132/0.0340 (Adam), 0.0086/0.0146 (Muon), and 0.0026/0.0031 (SGD), with all ten top features shared. The same accepted budget is applied to each optimizer's complete 14-checkpoint group (one representative plus 13 repeats).

The earlier Transformer DeepSHAP evaluations failed prediction-reconstruction checks and emitted an unsupported-layer warning. This concerns the attribution decomposition, not the correctness of model predictions. KernelSHAP reconstruction is checked, but its near-exact reconstruction is partly enforced by the estimator and is insufficient on its own; the repeat/budget checks assess profile stability. These finite-sample checks support use of the displayed summaries without establishing a unique explanation under correlated lagged inputs.

\FloatBarrier
\begin{figure}[!htbp]\centering
\begin{subfigure}{.48\linewidth}\centering
\includegraphics[width=\linewidth]{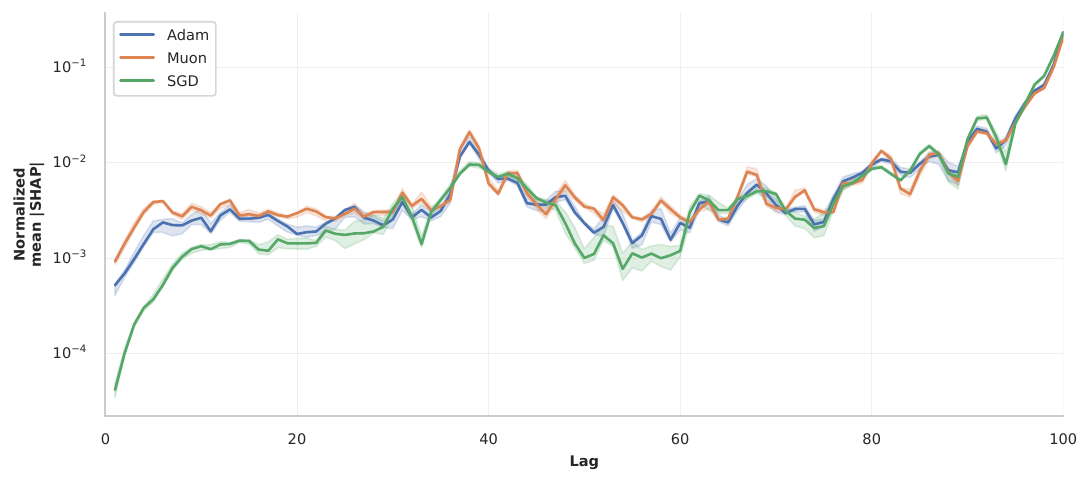}
\caption{CNN}\label{fig:shap_cnn_app}
\end{subfigure}\hfill
\begin{subfigure}{.48\linewidth}\centering
\includegraphics[width=\linewidth]{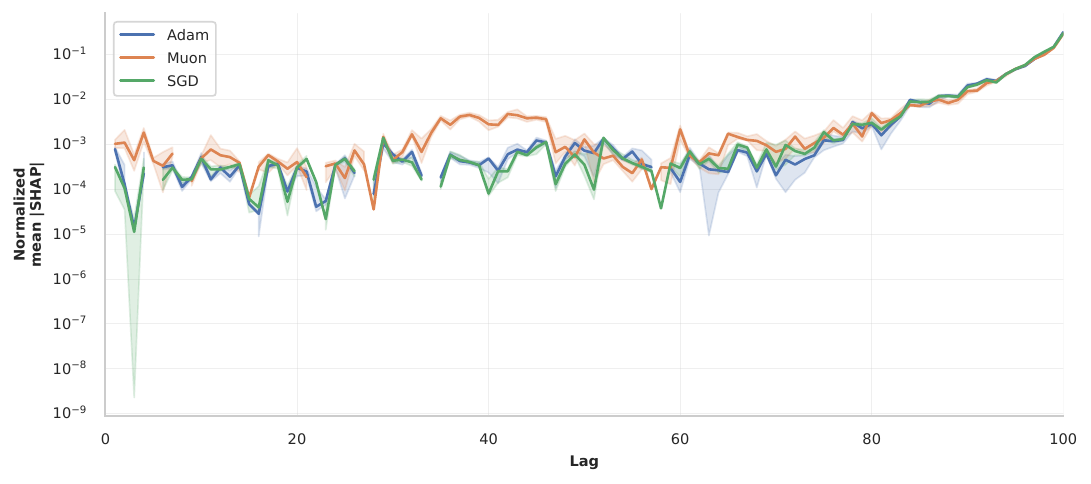}
\caption{LSTM}\label{fig:shap_lstm_app}
\end{subfigure}\par\smallskip
\begin{subfigure}{.48\linewidth}\centering
\includegraphics[width=\linewidth]{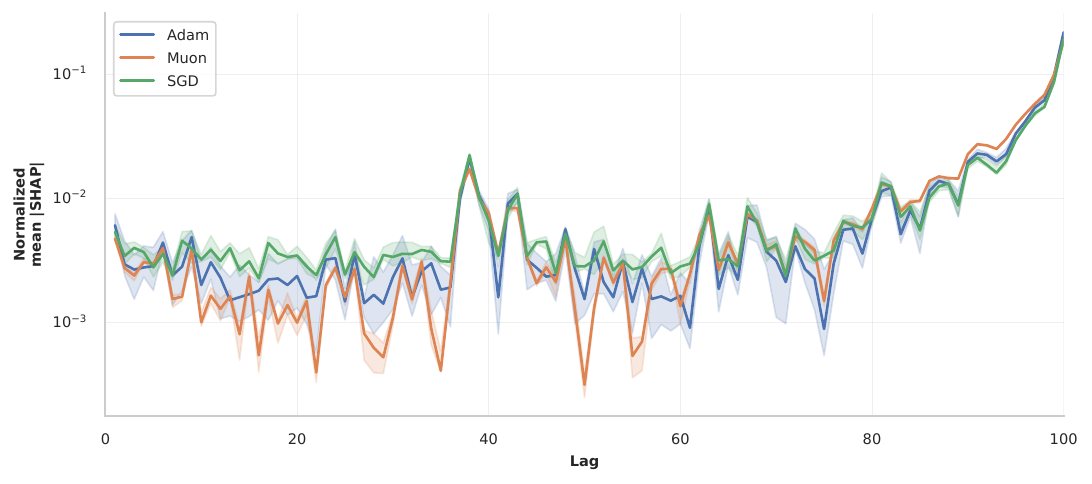}
\caption{MLP}\label{fig:shap_mlp_app}
\end{subfigure}\hfill
\begin{subfigure}{.48\linewidth}\centering
\includegraphics[width=\linewidth]{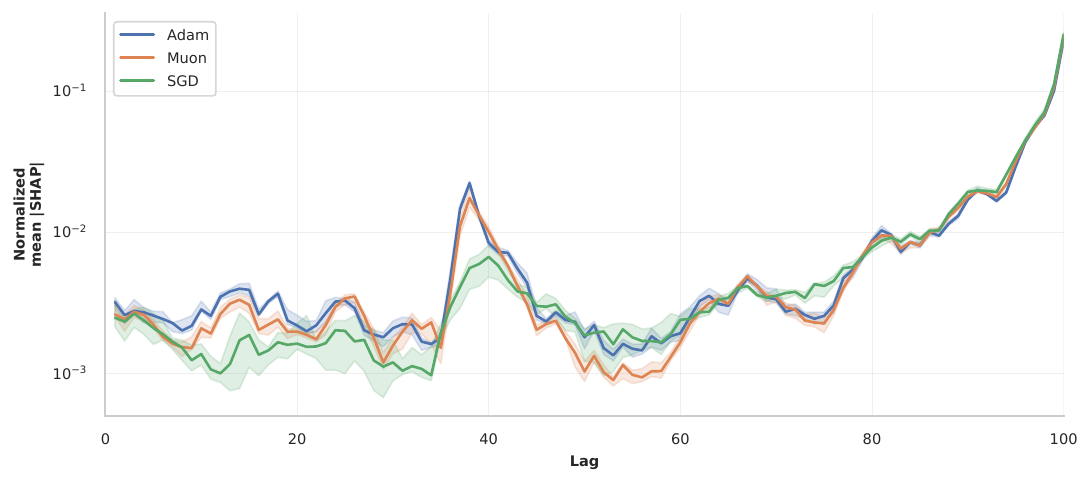}
\caption{Transformer}\label{fig:shap_transformer}
\end{subfigure}
\caption{\textbf{Optimizer-Driven Feature Attribution (SHAP).} Normalized mean |SHAP| across input lags over 13 runs. Solid lines show medians; shaded bands show the IQR (seed spread). The vertical axis is logarithmic. Positions run from 1 (oldest input) to 100 (most recent input).}
\label{fig:shap_comparison_grid}
\end{figure}

\begin{figure}[htbp]
    \centering
    \includegraphics[width=\linewidth,trim=0 0 0 29bp,clip]{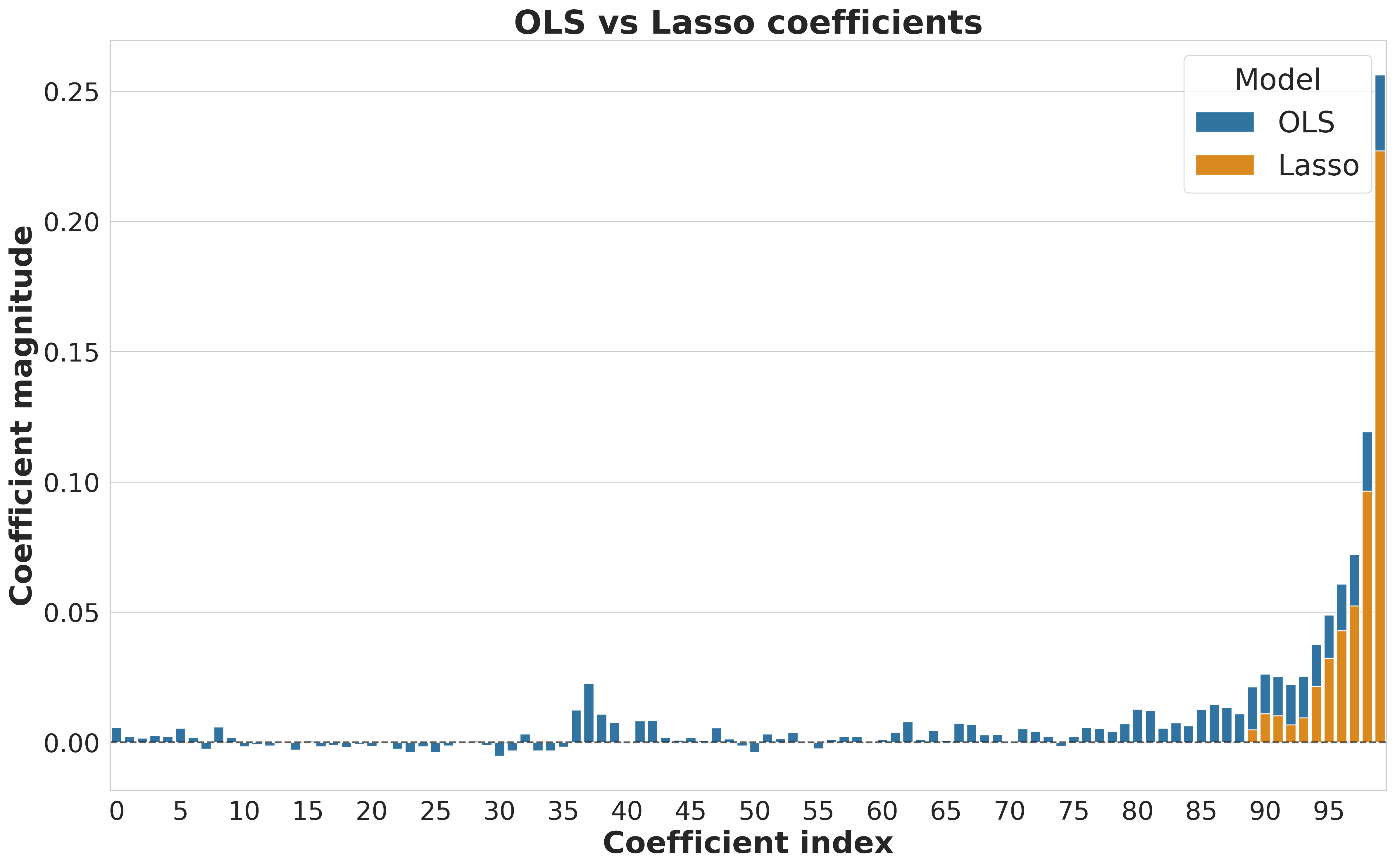}
    \caption{\textbf{Linear Baselines.} Coefficients of OLS and LASSO regressions for 100 lagged squared volatility terms ($\sigma^2_{t-99}$ through $\sigma^2_{t}$). Note the sparsity of LASSO versus the dense allocation of OLS.}
    \label{fig:ols_lasso_coefficients}
\end{figure}

\begin{figure}[tbp]\centering
\begin{subfigure}{.48\linewidth}

        \centering
        \includegraphics[width=\linewidth]{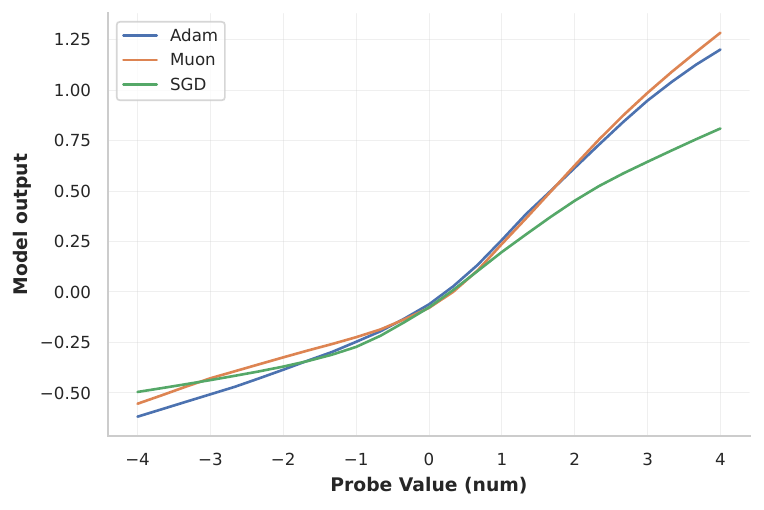}

        \caption{\textbf{CNN} Impulse Response}
        \label{fig:diff_cnn}

\end{subfigure}\hfill
\begin{subfigure}{.48\linewidth}

        \centering
        \includegraphics[width=\linewidth]{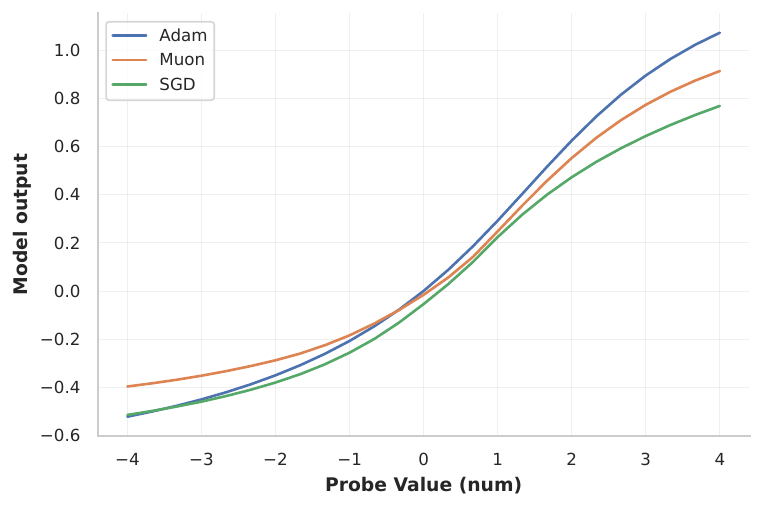}
        \caption{\textbf{LSTM} Impulse Response}
        \label{fig:diff_lstm}

\end{subfigure}\par\medskip
\begin{subfigure}{.48\linewidth}

        \centering
        \includegraphics[width=\linewidth]{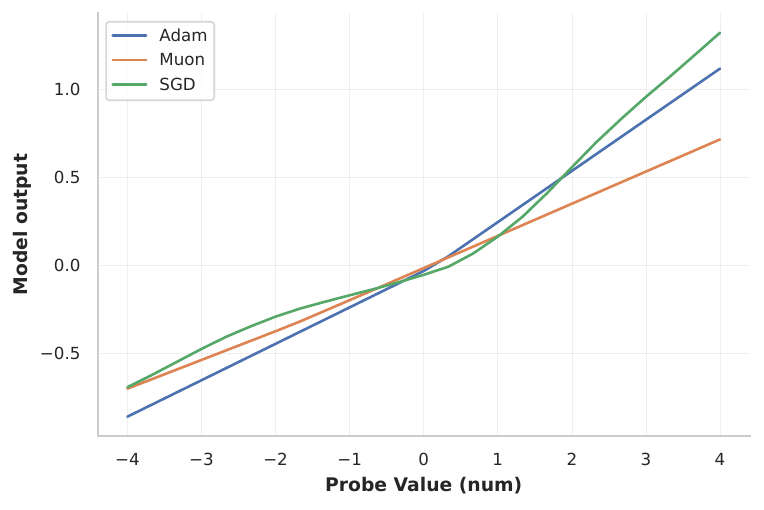}
        \caption{\textbf{MLP} Impulse Response}
        \label{fig:diff_mlp}

\end{subfigure}\hfill
\begin{subfigure}{.48\linewidth}

        \centering
        \includegraphics[width=\linewidth]{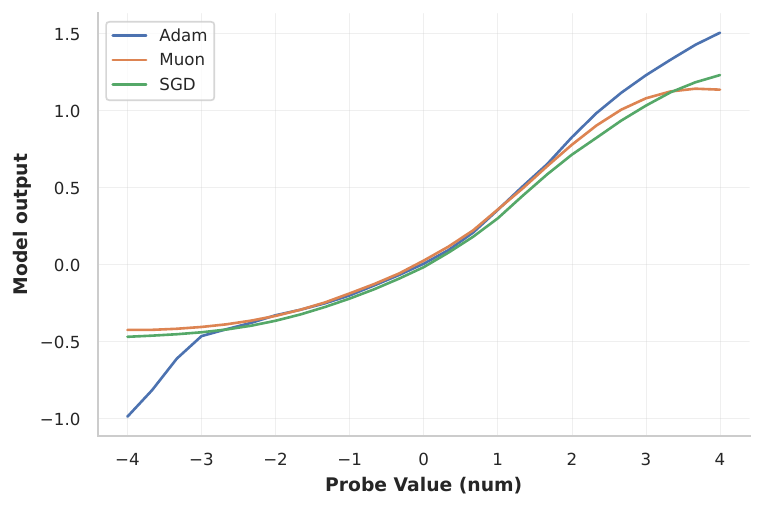}
        \caption{\textbf{Transformer} Impulse Response}
        \label{fig:diff_transformer}

\end{subfigure}\par\medskip
\caption{\textbf{Functional Divergence (Impulse Responses).} Comparison of model sensitivity at $t-1$ across optimizers.}
\label{fig:difference_profiles_grid}
\end{figure}

\begin{figure}[p]
\begin{minipage}{\linewidth}\centering
\begin{subfigure}{.32\linewidth}

        \centering
        \includegraphics[width=\linewidth]{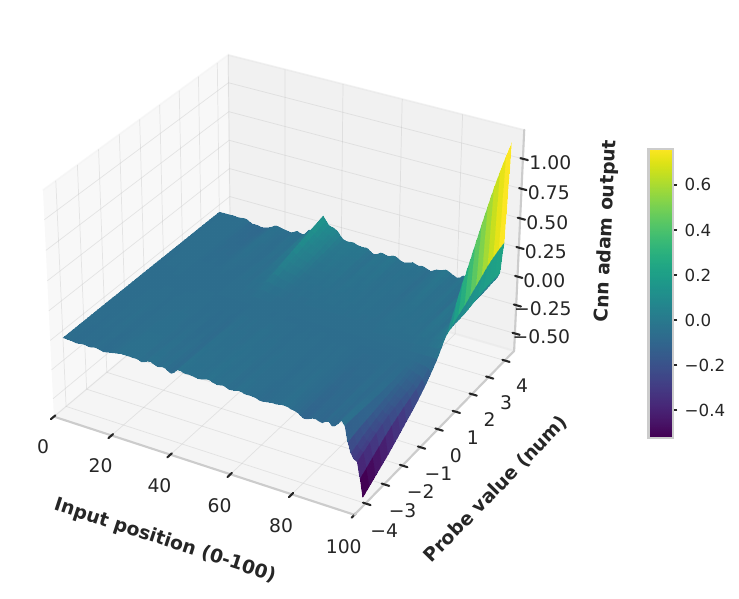}
        \caption{Adam}
        \label{fig:surf_cnn_adam}

\end{subfigure}\hfill
\begin{subfigure}{.32\linewidth}

        \centering
        \includegraphics[width=\linewidth]{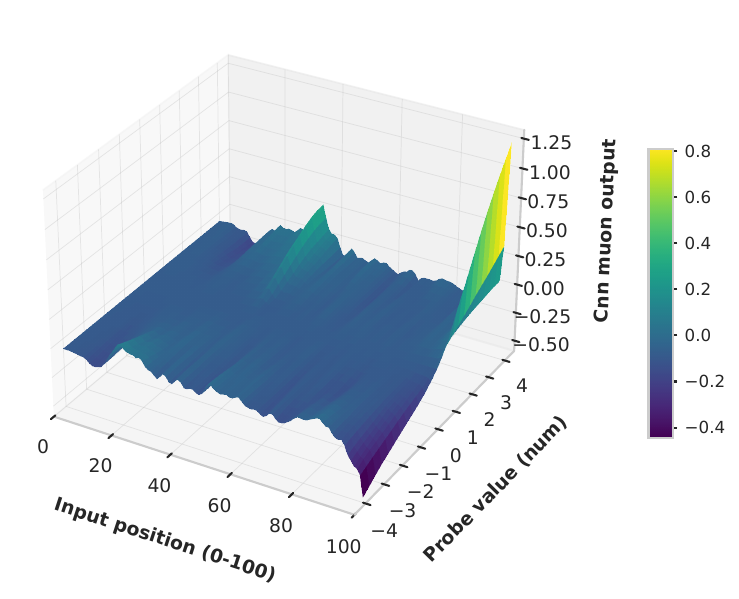}
        \caption{Muon}
        \label{fig:surf_cnn_muon}

\end{subfigure}\hfill
\begin{subfigure}{.32\linewidth}

        \centering
        \includegraphics[width=\linewidth]{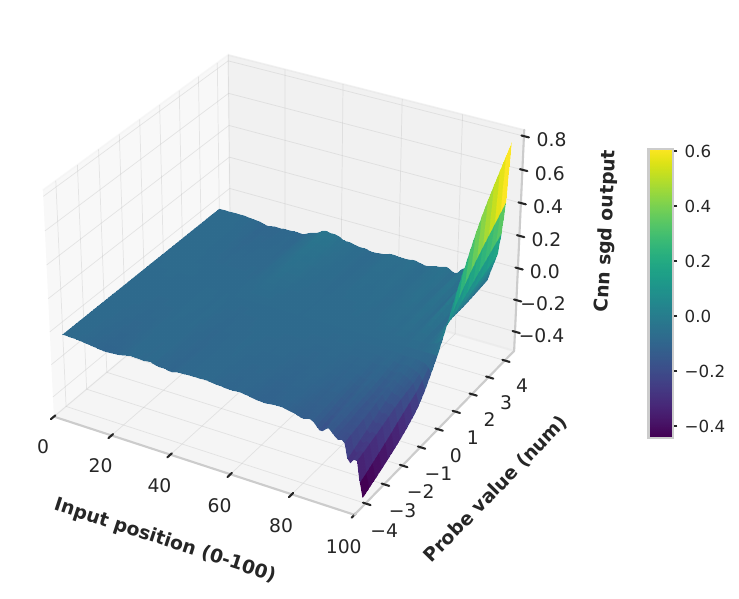}
        \caption{SGD}
        \label{fig:surf_cnn_sgd}

\end{subfigure}\par\smallskip
\caption{\textbf{CNN response surfaces.} Left to right: Adam, Muon, SGD.}
\label{fig:response_surfaces_cnn}
\end{minipage}
\par\medskip
\begin{minipage}{\linewidth}\centering
\begin{subfigure}{.32\linewidth}

        \centering
        \includegraphics[width=\linewidth]{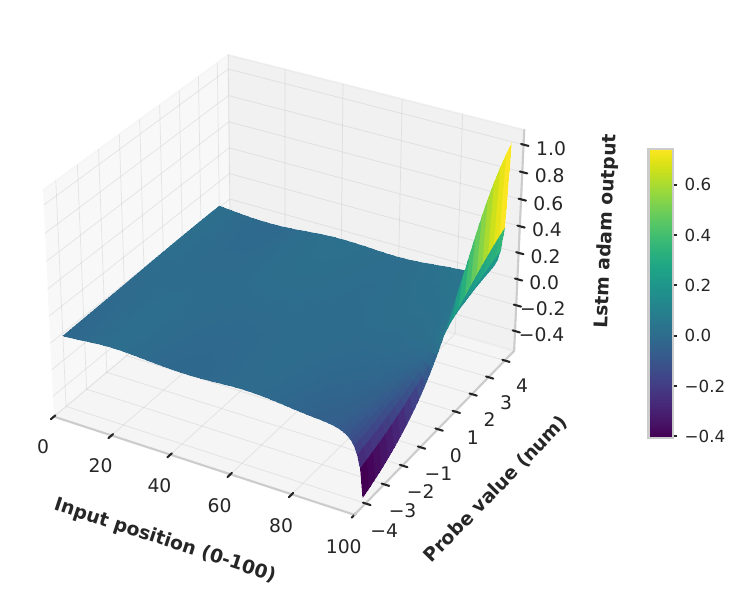}
        \caption{Adam}
        \label{fig:surf_lstm_adam}

\end{subfigure}\hfill
\begin{subfigure}{.32\linewidth}

        \centering
        \includegraphics[width=\linewidth]{revision_figs/response_surface_lstm_muon.pdf}
        \caption{Muon}
        \label{fig:surf_lstm_muon}

\end{subfigure}\hfill
\begin{subfigure}{.32\linewidth}

        \centering
        \includegraphics[width=\linewidth]{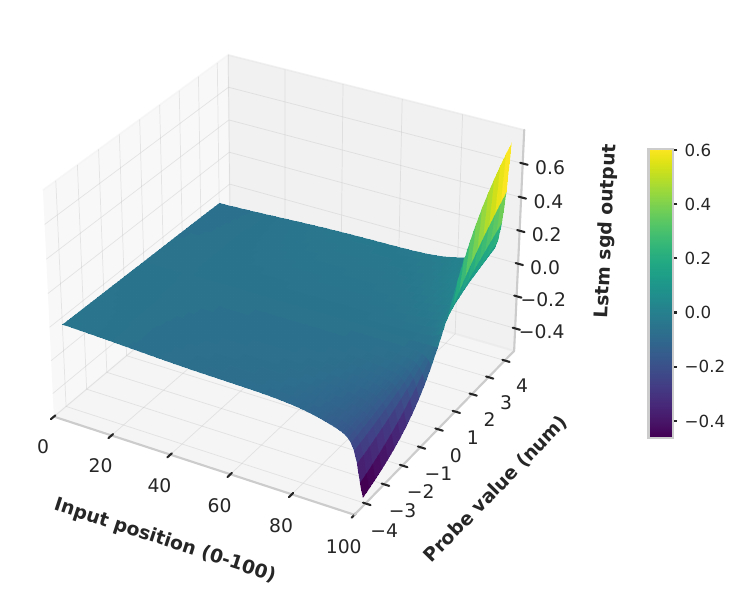}
        \caption{SGD}
        \label{fig:surf_lstm_sgd}

\end{subfigure}\par\smallskip
\caption{\textbf{LSTM response surfaces.} Left to right: Adam, Muon, SGD.}
\label{fig:response_surfaces_lstm}
\end{minipage}
\par\medskip
\begin{minipage}{\linewidth}\centering
\begin{subfigure}{.32\linewidth}

        \centering
        \includegraphics[width=\linewidth]{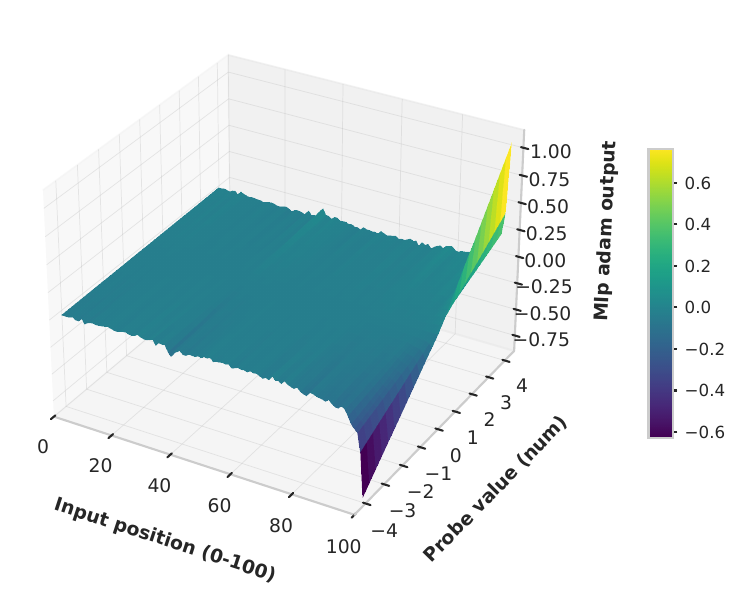}
        \caption{Adam}
        \label{fig:surf_mlp_adam}

\end{subfigure}\hfill
\begin{subfigure}{.32\linewidth}

        \centering
        \includegraphics[width=\linewidth]{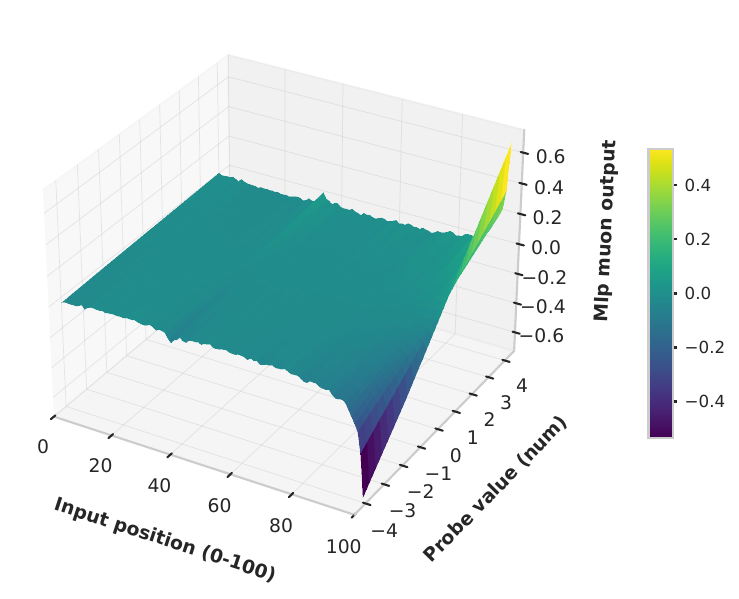}
        \caption{Muon}
        \label{fig:surf_mlp_muon}

\end{subfigure}\hfill
\begin{subfigure}{.32\linewidth}

        \centering
        \includegraphics[width=\linewidth]{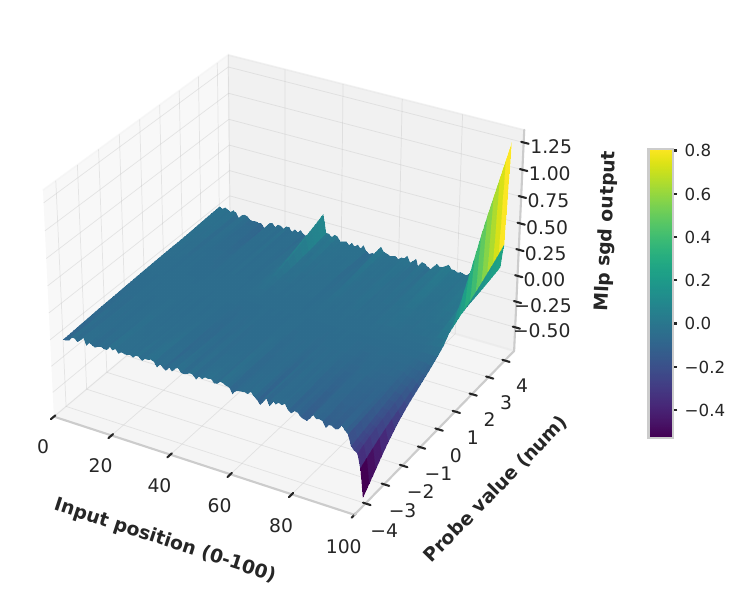}
        \caption{SGD}
        \label{fig:surf_mlp_sgd}

\end{subfigure}\par\smallskip
\caption{\textbf{MLP response surfaces.} Left to right: Adam, Muon, SGD.}
\label{fig:response_surfaces_mlp}
\end{minipage}
\par\medskip
\begin{minipage}{\linewidth}\centering
\begin{subfigure}{.32\linewidth}

        \centering
        \includegraphics[width=\linewidth]{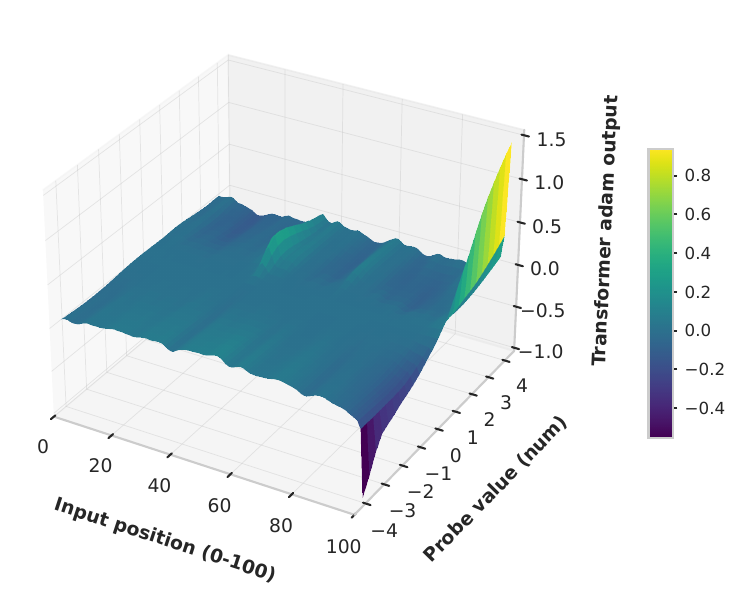}
        \caption{Adam}
        \label{fig:surf_transformer_adam}

\end{subfigure}\hfill
\begin{subfigure}{.32\linewidth}

        \centering
        \includegraphics[width=\linewidth]{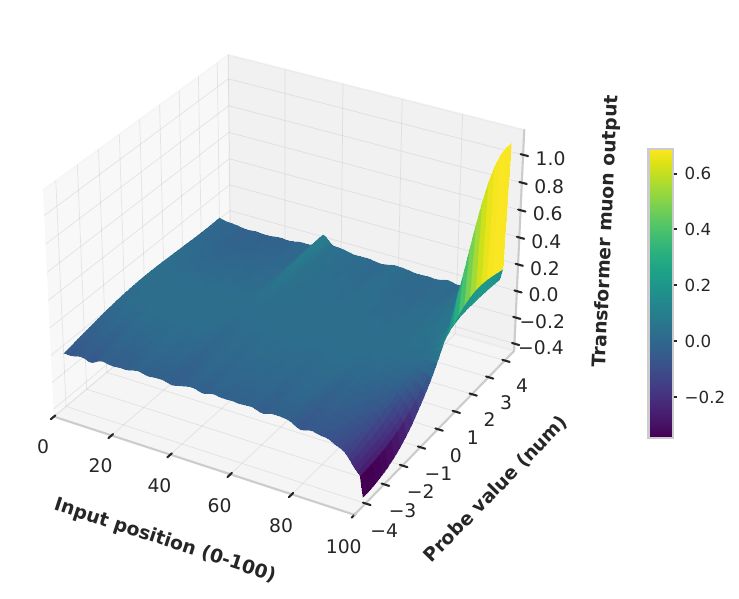}
        \caption{Muon}
        \label{fig:surf_transformer_muon}

\end{subfigure}\hfill
\begin{subfigure}{.32\linewidth}

        \centering
        \includegraphics[width=\linewidth]{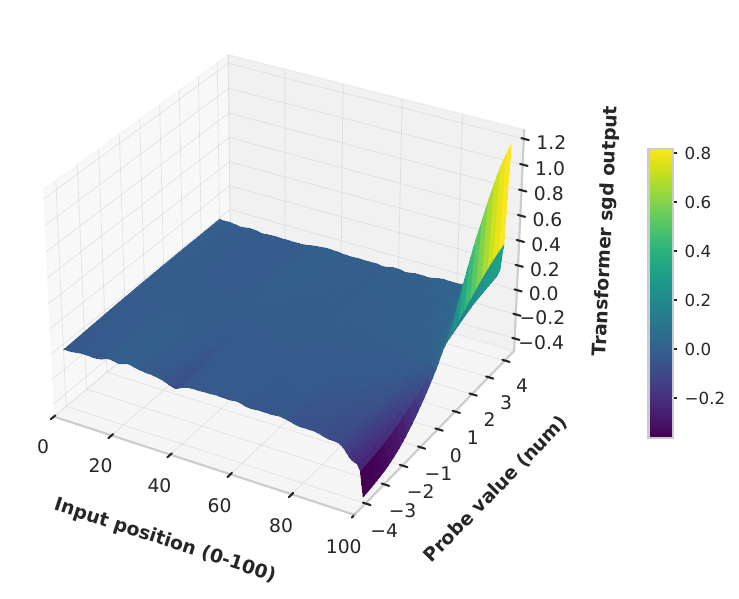}
        \caption{SGD}
        \label{fig:surf_transformer_sgd}

\end{subfigure}\par\smallskip
\caption{\textbf{Transformer response surfaces.} Left to right: Adam, Muon, SGD.}
\label{fig:response_surfaces_transformer}
\end{minipage}
\end{figure}

\FloatBarrier

\section{Search for a Mechanism}
\label{app:mechanism}

\subsection{Stability Constraints and Function Selection}
\label{app:eos}
\label{app:weight_decay_stability}
How do training procedures with similar forecasting errors select different functions? The stability viewpoint offers a candidate mechanism: the region reached by training depends on the constraints imposed by its update dynamics. Self-stabilization can therefore contribute to function selection, rather than merely determine the speed of approach to a solution \citep{damian_self-stabilization_2023}.

In full-batch gradient descent, the edge of stability describes continued training near the local quadratic boundary $2/\eta$, with short-time oscillations and nonlinear regulation of curvature \citep{cohen2022gradientdescentneuralnetworks}. Central flow formalizes a related constraint on the time-averaged trajectory \citep{cohen_understanding_2024}; stochastic training additionally couples curvature to mini-batch gradients \citep{andreyev2025edgestochasticstabilityrevisiting}. These results motivate a direct experiment: change the dynamics at a fitted solution and ask whether sharpening and further fitting resume.

Around a fixed positive-definite quadratic minimum, reducing the learning rate contracts the stationary fluctuations of SGD without changing the Hessian \citep{mandt2017}. If, instead, the apparent endpoint is maintained by a dynamical curvature constraint, smaller steps can permit continued motion into sharper regions. \emph{Reducing the learning rate releases renewed progressive sharpening, indicating that the preceding curvature plateau reflects a constraint of the training dynamics rather than merely fluctuations around a fixed local quadratic minimum.} This is the interpretation tested below: the interventions support a dynamical explanation, but do not prove the absence of a local minimum, since contraction within a nonquadratic neighborhood can also change curvature.

The numerical height of the plateau is not the sole criterion. Weight decay can alter progressive sharpening and sustain subcritical curvature levels in regularized full-batch dynamics \citep{saether2026weightdecay}. This supplies a reason to examine the response to interventions rather than infer unconstrained convergence from $\lambda_{\max}<2/\eta$ alone. Our minibatch experiments test that response; they do not identify weight decay as its unique cause.

\subsection{Measurements and Intervention Design}
\label{app:spikes}
\label{app:intervention_experiments}
We record the leading data-loss Hessian eigenvalue $\lambda_{\max}$ and directional mini-batch curvature. For mini-batch gradient $g_B$ and Hessian $H_B$, the plotted diagnostic is
\begin{equation}
\mathrm{BS}_{\mathrm{diag}}=\mathbb{E}_B\!\left[\frac{g_B^\top H_Bg_B}{\|g_B\|^2}\right].
\label{eq:batch_sharpness_diagnostic}
\end{equation}
It averages Rayleigh quotients; the literature definition instead uses $\mathbb{E}_B[g_B^\top H_Bg_B]/\mathbb{E}_B[\|g_B\|^2]$, weighting batches by gradient norm \citep{andreyev2025edgestochasticstabilityrevisiting}. The two need not coincide.

Each intervention starts from common fitted weights and saved optimizer state: unchanged control, LR/10, or batch$\times2$. Decay settings are preserved. We compare equal additional data passes, so batch doubling halves the update count and can change accumulated regularization. Curvature is measured in evaluation mode, excluding decay from the Hessian. Diagnostic batch size is 512 for control and LR/10, and 1,024 for batch$\times2$; the latter changes both training and the measured statistic. Appendix~\ref{app:curvature_measurement_settings} records numerical settings.

Training panels are separate realizations, not the histories leading to the continuation checkpoints. CNN--SGD and LSTM--SGD use the historical diagnostic settings. Transformer--SGD training uses $\eta=0.0028016$, while its continuation uses the selected configuration with $\eta=0.0030821$; these rows are contextual comparisons, not one trajectory. The remaining pairs use the corresponding predictive settings. Recorded horizons appear on the axes.

\subsection{Training and Intervention Evidence}
\label{app:training_curvature_evolution}
\label{app:continuation_atlas}
\begin{figure}[!htbp]\centering
\includegraphics[width=.95\linewidth]{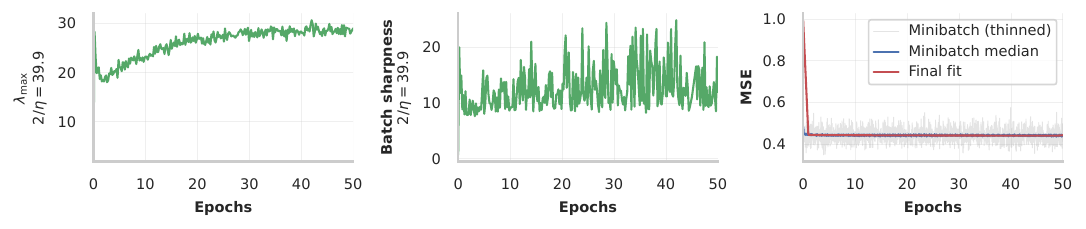}\par\smallskip
\includegraphics[width=.95\linewidth]{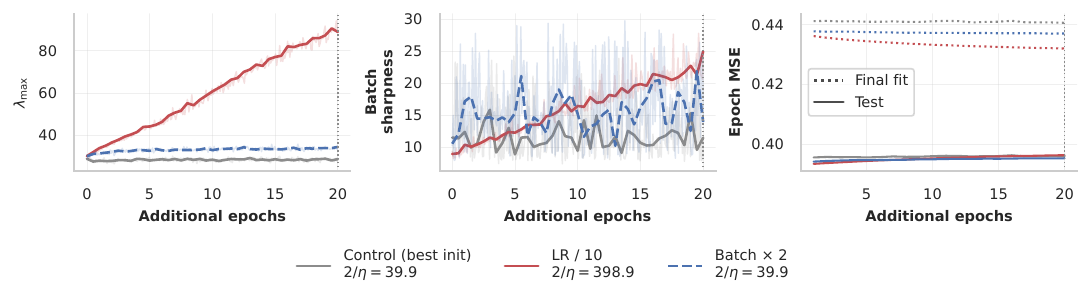}
\caption{\textbf{MLP--SGD.} Training: sharpening then plateau. LR/10: renewed sharpening in both measures and lower final-fit loss; batch doubling gives a smaller response.}\label{fig:paired_mlp_sgd}\label{fig:training_evolution_mlp}\label{fig:atlas_mlp_curvature}\label{fig:atlas_mlp_loss}
\end{figure}

\paragraph{SGD: releasing the curvature constraint.}
MLP provides the clearest illustration (Figure~\ref{fig:paired_mlp_sgd}). Training sharpens before settling into a fluctuating range. From the fitted solution, LR/10 releases a sustained increase in both curvature measures, accompanied by further loss reduction. Over the final fifth of the common 20-epoch window, median $\lambda_{\max}$ is approximately $29$ under control, $85$ under LR/10 and $34$ under batch$\times2$; final-fit MSE at epoch 20 is $0.4406$, $0.4320$ and $0.4371$, respectively. The intervention thus exposes further fitting beyond the preceding plateau.

\begin{figure}[!htbp]\centering
\includegraphics[width=.95\linewidth]{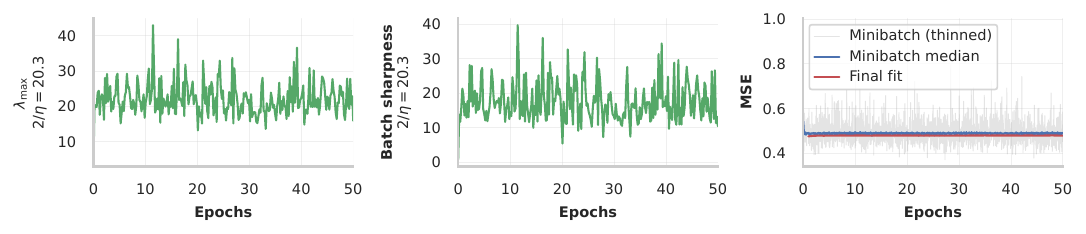}\par\smallskip
\includegraphics[width=.95\linewidth]{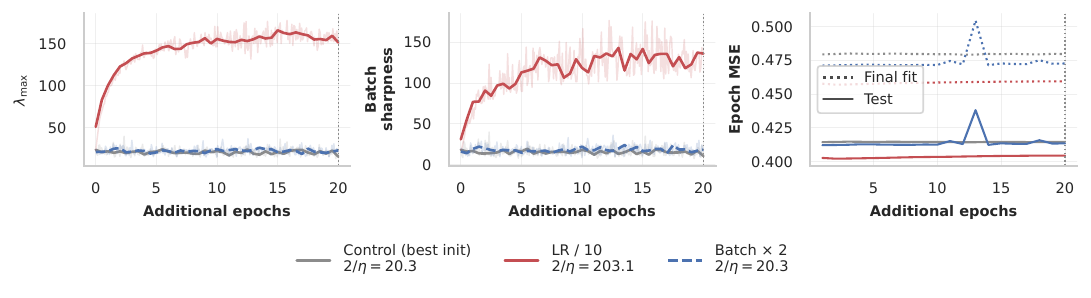}
\caption{\textbf{CNN--SGD.} Training: fluctuating curvature plateau. LR/10: strong renewed sharpening and a higher level in both measures; batch doubling shows a transient loss spike.}\label{fig:paired_cnn_sgd}\label{fig:training_evolution_cnn}\label{fig:atlas_cnn_curvature}\label{fig:atlas_cnn_loss}
\end{figure}

\begin{figure}[!htbp]\centering
\includegraphics[width=.95\linewidth]{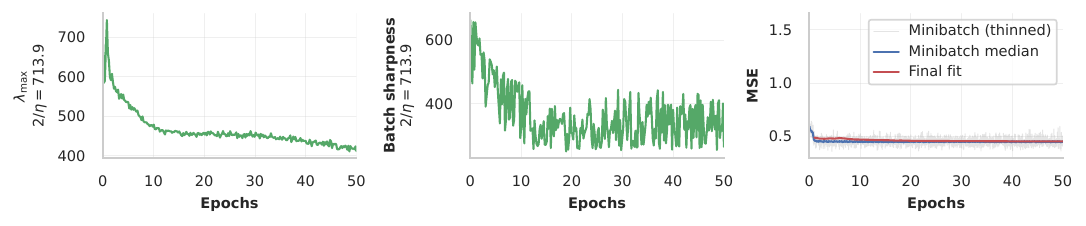}\par\smallskip
\includegraphics[width=.95\linewidth]{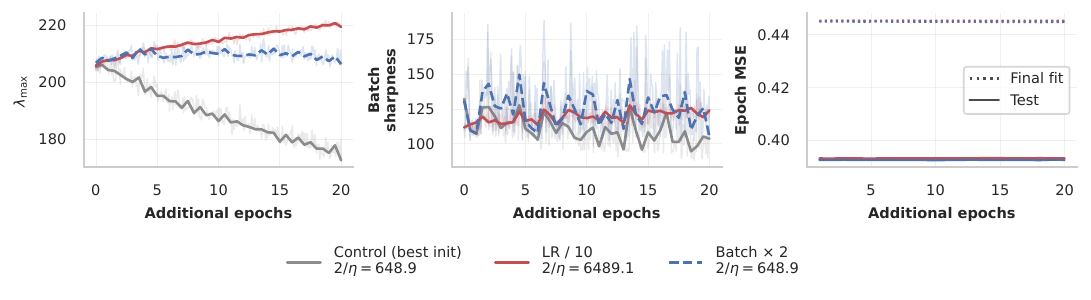}
\caption{\textbf{Transformer--SGD.} Training: declining curvature in the historical configuration. Selected-configuration continuation: modest eigenvalue increase under LR/10, with noisy batch sharpness; the two rows are not a continuous run.}\label{fig:paired_transformer_sgd}\label{fig:training_evolution_transformer}\label{fig:atlas_transformer_curvature}\label{fig:atlas_transformer_loss}
\end{figure}

The CNN diagnostic shows a stronger response: LR/10 raises both curvature measures and approaches a higher level, whereas control remains near its previous range (Figure~\ref{fig:paired_cnn_sgd}). The late-window eigenvalue is about $7.6$ times control. Transformer--SGD shows a weaker response: its continuation eigenvalue rises modestly under LR/10 while control declines, and batch sharpness remains noisy (Figure~\ref{fig:paired_transformer_sgd}). Late-window eigenvalues are approximately $178$, $219$ and $209$ for control, LR/10 and batch$\times2$. Its separately initialized training trace decreases rather than progressively sharpens.

\begin{figure}[!htbp]\centering
\includegraphics[width=.95\linewidth]{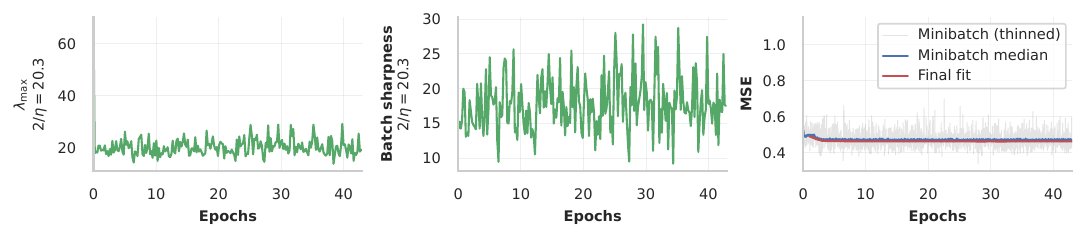}\par\smallskip
\includegraphics[width=.95\linewidth]{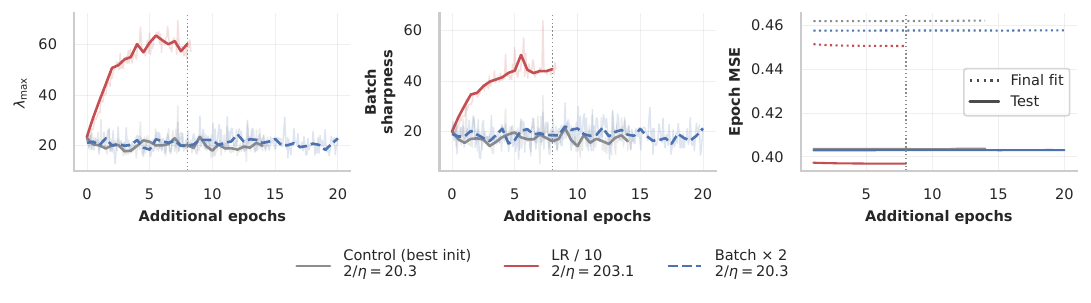}
\caption{\textbf{LSTM--SGD.} Training: sharpening toward a fluctuating level. LR/10: renewed sharpening in both measures over the common eight-epoch window; the intervention arms have different completed horizons. Training above; interventions below.}\label{fig:paired_lstm_sgd}\label{fig:training_evolution_lstm}\label{fig:atlas_lstm_curvature}\label{fig:atlas_lstm_loss}
\end{figure}

The LSTM--SGD diagnostic also shows renewed sharpening after LR/10 (Figure~\ref{fig:paired_lstm_sgd}). Over the common eight-epoch window, its late-window eigenvalue is about $2.8$ times control. The LR/10 arm ends earlier than the other arms; this supports a response comparison rather than a shared convergence claim.

Together, the SGD interventions support stability-constrained region selection, most clearly for MLP and CNN. Renewed sharpening, stabilization at a higher level and inverse-learning-rate scaling are distinct observations: the first does not establish the other two. The weaker batch response may be consistent with the large-batch regime discussed by \citet[Section~5.2]{andreyev2025edgestochasticstabilityrevisiting}, but these experiments do not locate its crossover.

\paragraph{Reading the figures.}
Each figure pairs training (top) and interventions (bottom); columns show $\lambda_{\max}$, batch sharpness and loss. Training curvature is green for SGD, blue for Adam and orange for Muon. Training-loss traces are grey for minibatch measurements, blue for bin medians and red for final-fit MSE. In continuation rows, grey denotes control, red LR/10 and dashed blue batch$\times2$; dotted loss lines show final fit and solid lines show test MSE. Faint/dark curvature traces show measurements/half-epoch medians, and vertical dotted lines mark the common completed horizon. The axes or legends report $2/\eta$ for SGD and the adaptive threshold $38/\eta$ of \citet{cohen_adaptive_2022} for Adam; raw curvature is not normalized. The same line is drawn for Muon purely as a reference, no threshold having been established for it.

\FloatBarrier

\begin{figure}[!htbp]\centering
\includegraphics[width=.95\linewidth]{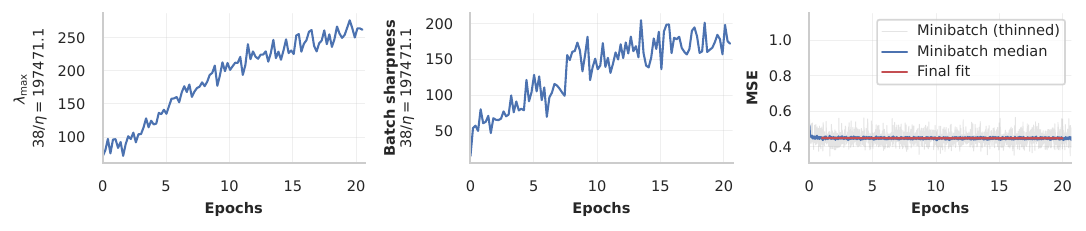}\par\smallskip
\includegraphics[width=.95\linewidth]{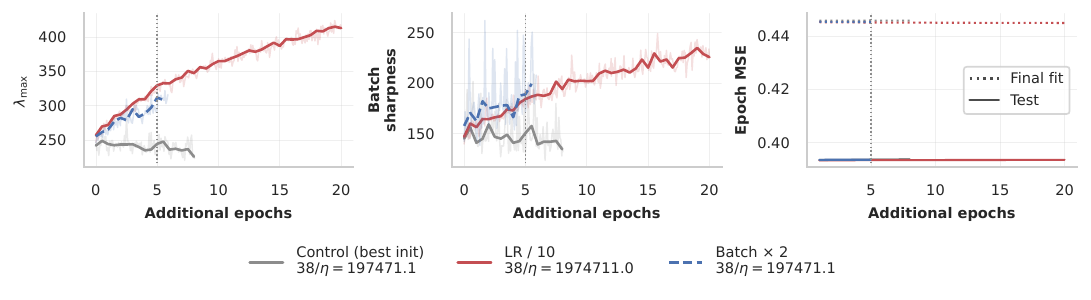}
\caption{\textbf{LSTM--Adam.} Training: continued sharpening. LR/10 raises both curvature measures relative to control over the shorter common observation window; the arms have different completed horizons. Training above; interventions below.}\label{fig:paired_lstm_adam}
\end{figure}

\paragraph{Adam: modest or persistent curvature differences.}
MLP--Adam continues sharpening during training, but the continuation arms show little separation. CNN--Adam also sharpens during training; LR/10 produces a modestly higher continuation eigenvalue, with much less separation in batch sharpness. LSTM--Adam shows upward responses in both curvature measures over its shorter common window (Figure~\ref{fig:paired_lstm_adam}). Transformer--Adam behaves differently: training curvature declines, and LR/10 largely maintains its starting eigenvalue while control decreases. Remaining above control here is not renewed progressive sharpening. Adaptive stability depends on effective update geometry, including optimizer state \citep{cohen_adaptive_2022}; the raw Hessian and batch diagnostic do not measure that complete geometry. We therefore describe Adam and Muon empirically rather than apply the SGD mechanism unchanged.

\begin{figure}[!htbp]\centering
\includegraphics[width=.95\linewidth]{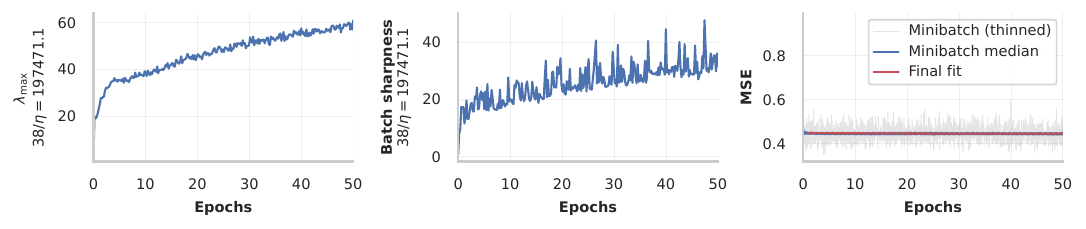}\par\smallskip
\includegraphics[width=.95\linewidth]{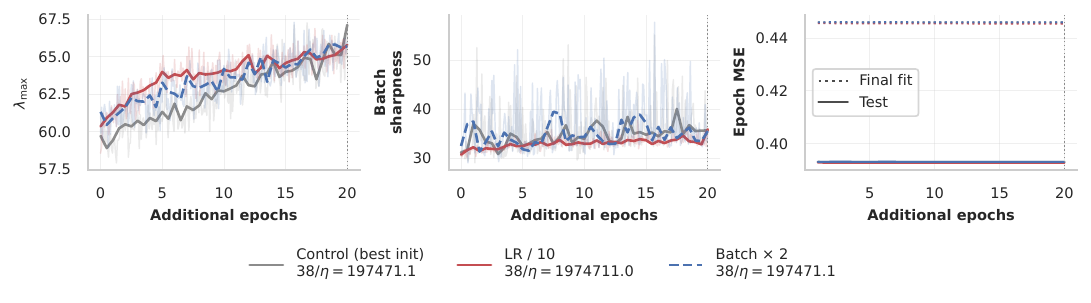}
\caption{\textbf{MLP--Adam.} Training: continued sharpening. Interventions: weak curvature separation and little visible loss separation.}\label{fig:paired_mlp_adam}
\end{figure}

\begin{figure}[!htbp]\centering
\includegraphics[width=.95\linewidth]{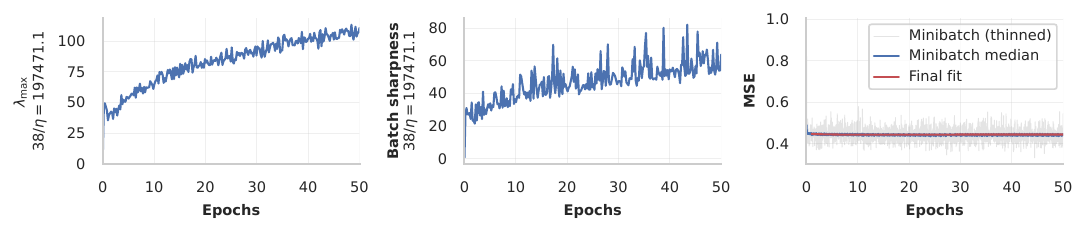}\par\smallskip
\includegraphics[width=.95\linewidth]{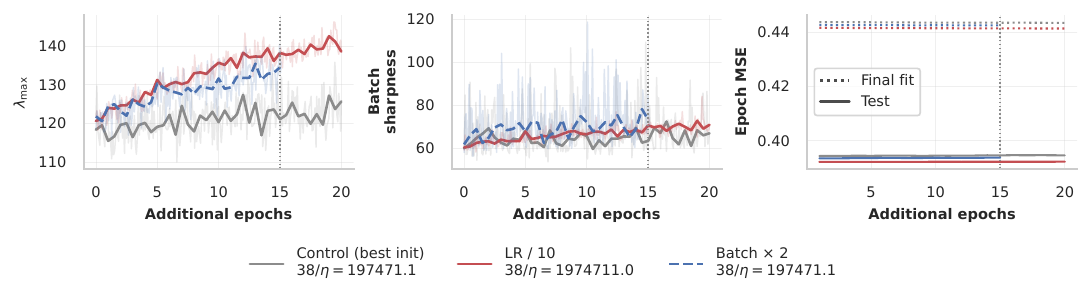}
\caption{\textbf{CNN--Adam.} Training: continued sharpening. LR/10: modest eigenvalue separation from control, with much weaker separation in batch sharpness.}\label{fig:paired_cnn_adam}
\end{figure}

\begin{figure}[!htbp]\centering
\includegraphics[width=.95\linewidth]{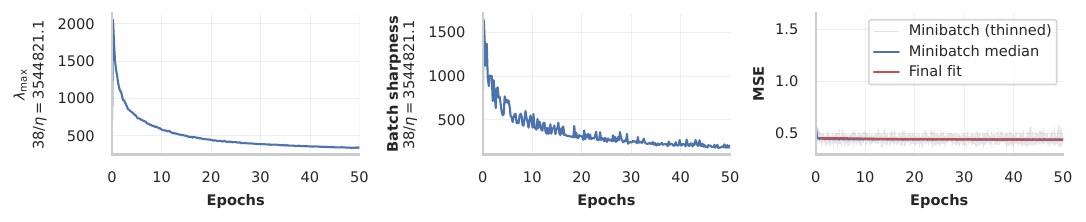}\par\smallskip
\includegraphics[width=.95\linewidth]{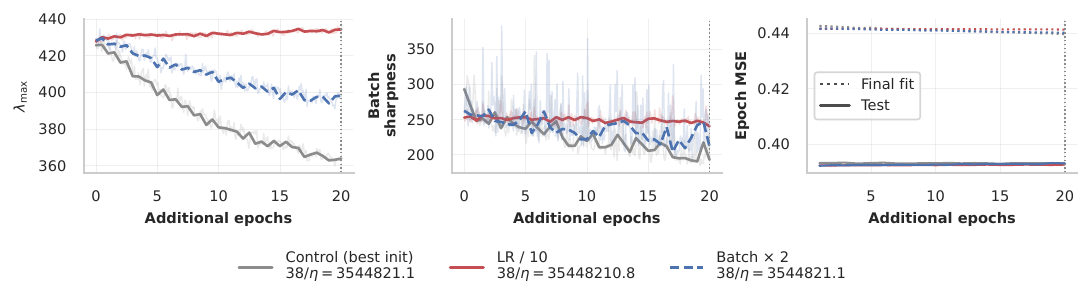}
\caption{\textbf{Transformer--Adam.} Training: declining curvature. LR/10 largely maintains the eigenvalue while control declines; batch sharpness is noisy and generally declines. This is not renewed sharpening.}\label{fig:paired_transformer_adam}
\end{figure}

\FloatBarrier

\begin{figure}[!htbp]\centering
\includegraphics[width=.95\linewidth]{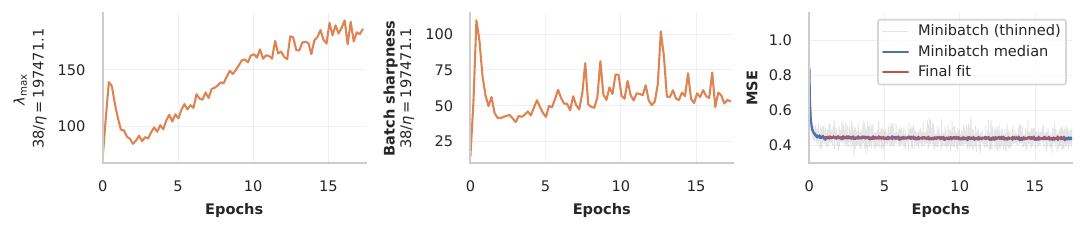}\par\smallskip
\includegraphics[width=.95\linewidth]{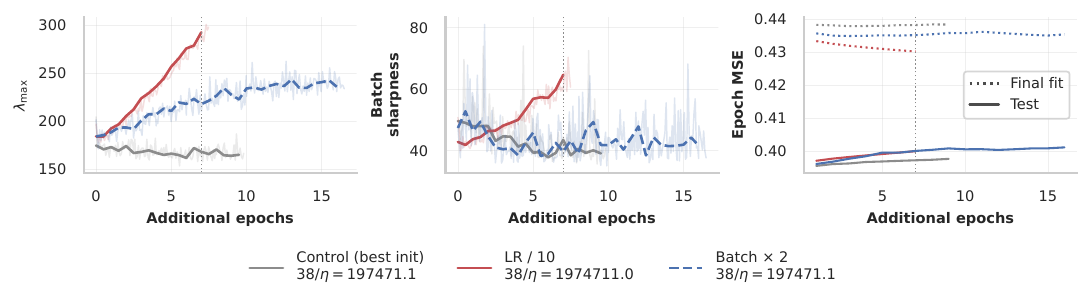}
\caption{\textbf{LSTM--Muon.} Training: continued sharpening. LR/10 raises both curvature measures over its recorded window, unlike the MLP and CNN responses; the arms have different completed horizons. Training above; interventions below.}\label{fig:paired_lstm_muon}
\end{figure}

\paragraph{Muon: heterogeneous responses.}
MLP--Muon sharpens during training, yet LR/10 ends with lower curvature than control. CNN--Muon shows continued training sharpening, while LR/10 slows subsequent growth in both curvature measures relative to control. LSTM--Muon instead shows an upward LR/10 response in both measures over its shorter recorded window (Figure~\ref{fig:paired_lstm_muon}). Transformer--Muon training curvature declines; LR/10 approximately maintains its eigenvalue and batch sharpness as control decreases. These responses differ from the strongest SGD examples. The stability mechanism for Muon is less understood, and the plotted reference does not turn these raw-curvature observations into a stability test.

\begin{figure}[!htbp]\centering
\includegraphics[width=.95\linewidth]{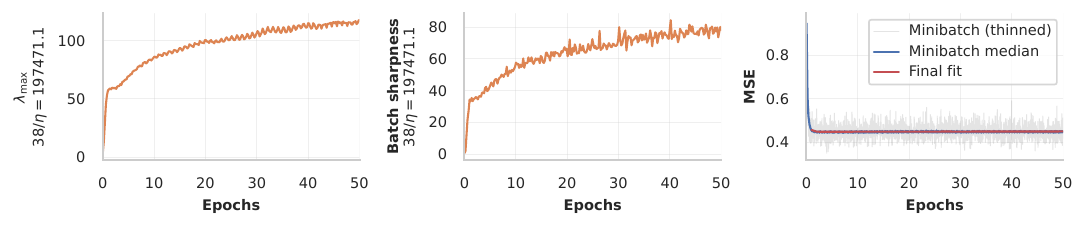}\par\smallskip
\includegraphics[width=.95\linewidth]{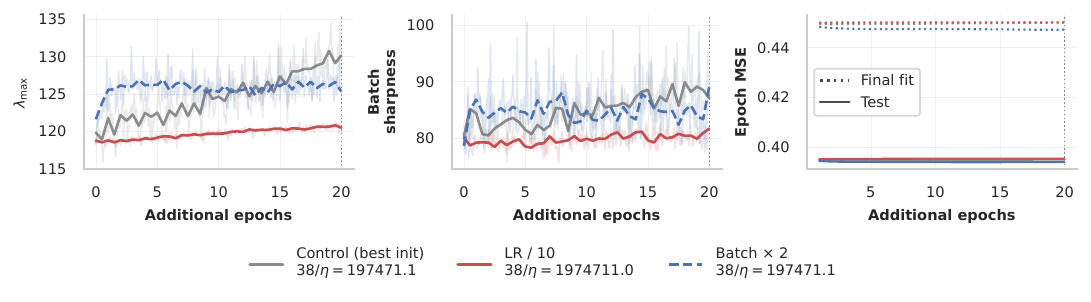}
\caption{\textbf{MLP--Muon.} Training: continued sharpening. LR/10 produces lower late-window eigenvalue and batch sharpness than control.}\label{fig:paired_mlp_muon}
\end{figure}

\begin{figure}[!htbp]\centering
\includegraphics[width=.95\linewidth]{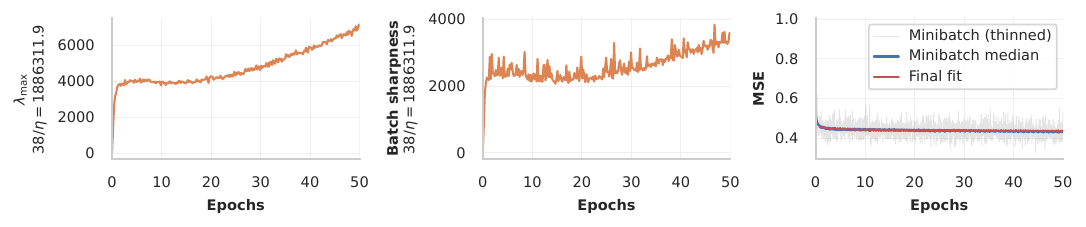}\par\smallskip
\includegraphics[width=.95\linewidth]{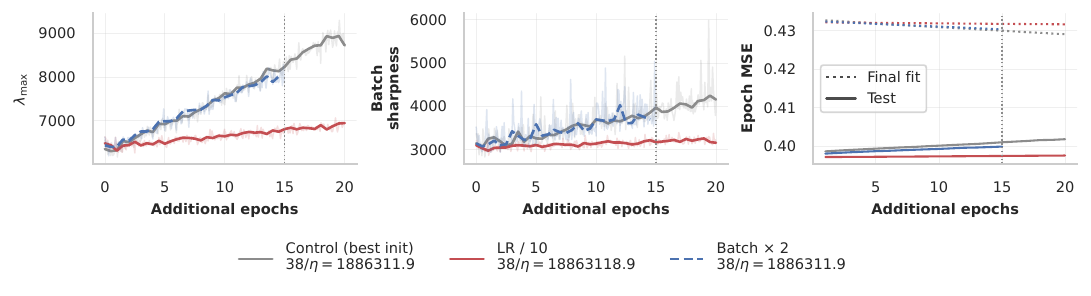}
\caption{\textbf{CNN--Muon.} Training: continued sharpening. LR/10 slows further growth in both curvature measures relative to control.}\label{fig:paired_cnn_muon}
\end{figure}

\begin{figure}[!htbp]\centering
\includegraphics[width=.95\linewidth]{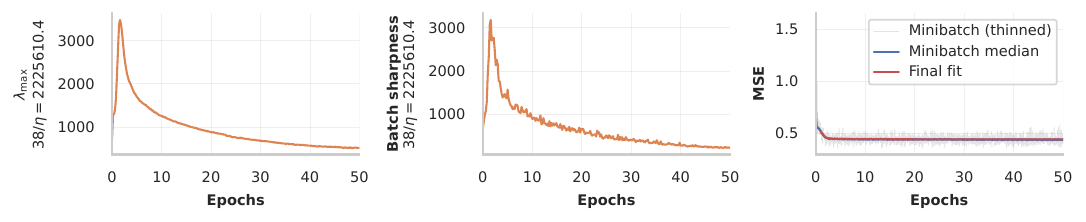}\par\smallskip
\includegraphics[width=.95\linewidth]{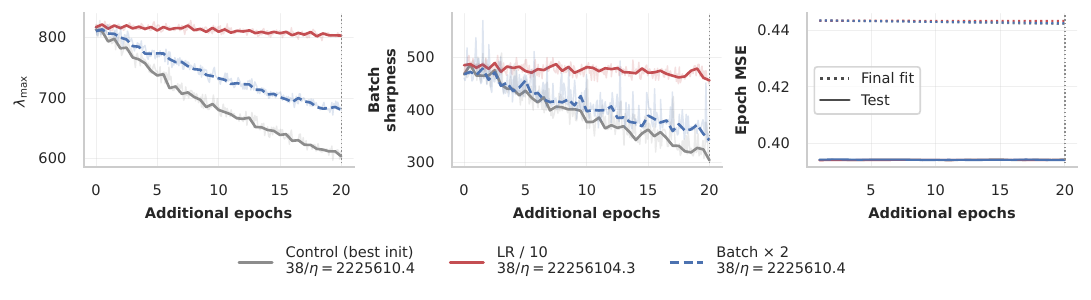}
\caption{\textbf{Transformer--Muon.} Training: declining curvature. LR/10 approximately maintains both measures while control declines; the difference reflects persistence rather than renewed sharpening.}\label{fig:paired_transformer_muon}
\end{figure}

\FloatBarrier

\paragraph{Movement in prediction space.}
\label{app:intervention_function_distance}
The available MLP--SGD checkpoints also connect the interventions to function selection. At the shared 20-epoch endpoint, prediction distances from the starting solution are $0.04328$ for control, $0.05101$ for LR/10 and $0.03810$ for batch$\times2$. Relative to control, LR/10 moves predictions by $0.05547$ RMS and batch$\times2$ by $0.02433$; the two intervention endpoints differ by $0.03909$. These use the same test observations and distance definition as Eq.~\eqref{eq:function_distance}. The interventions therefore change the learned predictions as well as curvature; this does not isolate curvature as the unique cause of functional movement.

\section{Volatility-Managed Portfolios}
\label{app:vol_managed}

\begin{figure}[t]
\centering
\includegraphics[width=.94\linewidth]{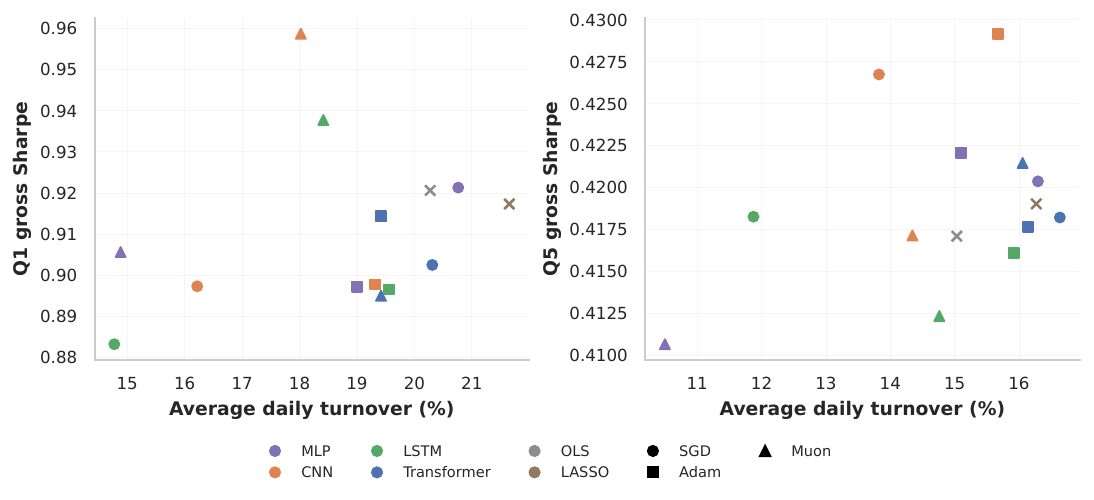}
\caption{\textbf{Sharpe--Turnover Frontier.} Selected representative models over the available May 2000--December 2024 history, including fitting observations. Q1 (left) and Q5 (right) are the lowest- and highest-predicted-volatility quintiles. Gross Sharpe ratios occupy a narrow range while average daily target-weight turnover differs across fitted predictors.}
\label{fig:sharpe_turnover_q1q5}
\end{figure}

This appendix reports additional details for the downstream portfolio experiment used to illustrate how optimizer-induced functional differences in volatility forecasts translate into differences in \emph{decision stability}. The analysis is not intended as a performance benchmark, but as a diagnostic of the implementability of metrically equivalent predictors.

\subsection{Cross-sectional implementation}
On each recorded trading day, assets are sorted into quintiles using predicted volatility. Within each quintile, portfolios are equal-weighted and reconstituted daily. The full-history diagnostics use 2,854,980 asset--date observations from 25 May 2000 to 24 December 2024 and the selected representative checkpoints. They include 2,290,185 fitting observations through 26 December 2019 and 564,795 held-out observations from 21 May 2020 onward. The full-history results therefore describe retrospective decision behavior, not a continuous out-of-sample trading backtest. Table~\ref{tab:vol_sorted_test_check} separately reports the held-out results.

Prediction rows are aligned by asset and observation identity, and the common merged-training transformation is inverted before portfolio construction. Portfolios are not averaged across training seeds. The original return/date convention is retained; executable signal timing is not independently established by this diagnostic.

\subsection{Turnover and performance metrics}
Portfolio turnover is computed as $\frac{1}{2}\sum_i|w_{i,t}-w_{i,t-1}|$ and averaged across successive recorded dates, where both weight vectors are equal-weight \emph{target} allocations. This measures changes in portfolio membership and target weights; it does not adjust previous weights for intervening asset returns. Annualized return is the mean daily return multiplied by 252, annualized volatility is daily standard deviation multiplied by $\sqrt{252}$, and the reported gross Sharpe ratio is their ratio, without risk-free subtraction. Maximum drawdown is computed from cumulative gross returns. No transaction costs are deducted, and no confidence interval or equality test for Sharpe ratios is asserted.

\begin{table}[htbp]
\centering\small
\caption{\textbf{Volatility-sorted portfolios over the full available history.} Selected representatives, May 2000--December 2024, including fitting observations. Q1 and Q5 denote lowest and highest predicted volatility. Return, volatility, drawdown and daily turnover are fractions.}
\label{tab:vol_sorted_summary}
\setlength{\tabcolsep}{4pt}
\begin{tabular}{lrrrrr}
\toprule
Model & Ann. Ret. & Ann. Vol. & Sharpe & Max DD & Turnover \\
\midrule
\multicolumn{6}{l}{\textbf{Q1: Low Volatility}} \\
OLS & 0.120 & 0.130 & 0.921 & -0.397 & 0.203 \\
LASSO & 0.119 & 0.130 & 0.917 & -0.402 & 0.217 \\
CNN (SGD) & 0.117 & 0.130 & 0.897 & -0.406 & 0.162 \\
CNN (Muon) & 0.124 & 0.129 & 0.959 & -0.389 & 0.180 \\
CNN (Adam) & 0.117 & 0.130 & 0.898 & -0.398 & 0.193 \\
LSTM (SGD) & 0.114 & 0.129 & 0.883 & -0.403 & 0.148 \\
LSTM (Muon) & 0.121 & 0.129 & 0.938 & -0.398 & 0.184 \\
LSTM (Adam) & 0.116 & 0.130 & 0.896 & -0.395 & 0.196 \\
MLP (SGD) & 0.120 & 0.130 & 0.921 & -0.397 & 0.208 \\
MLP (Muon) & 0.117 & 0.129 & 0.906 & -0.390 & 0.149 \\
MLP (Adam) & 0.117 & 0.130 & 0.897 & -0.397 & 0.190 \\
Transformer (SGD) & 0.117 & 0.130 & 0.902 & -0.395 & 0.203 \\
Transformer (Muon) & 0.116 & 0.130 & 0.895 & -0.392 & 0.194 \\
Transformer (Adam) & 0.119 & 0.130 & 0.914 & -0.396 & 0.194 \\
\midrule
\multicolumn{6}{l}{\textbf{Q5: High Volatility}} \\
OLS & 0.124 & 0.297 & 0.417 & -0.718 & 0.150 \\
LASSO & 0.124 & 0.296 & 0.419 & -0.717 & 0.163 \\
CNN (SGD) & 0.127 & 0.297 & 0.427 & -0.701 & 0.138 \\
CNN (Muon) & 0.124 & 0.297 & 0.417 & -0.684 & 0.143 \\
CNN (Adam) & 0.127 & 0.296 & 0.429 & -0.703 & 0.157 \\
LSTM (SGD) & 0.125 & 0.298 & 0.418 & -0.712 & 0.119 \\
LSTM (Muon) & 0.122 & 0.297 & 0.412 & -0.706 & 0.148 \\
LSTM (Adam) & 0.123 & 0.296 & 0.416 & -0.723 & 0.159 \\
MLP (SGD) & 0.124 & 0.296 & 0.420 & -0.709 & 0.163 \\
MLP (Muon) & 0.123 & 0.299 & 0.411 & -0.719 & 0.105 \\
MLP (Adam) & 0.125 & 0.297 & 0.422 & -0.712 & 0.151 \\
Transformer (SGD) & 0.124 & 0.295 & 0.418 & -0.717 & 0.166 \\
Transformer (Muon) & 0.125 & 0.295 & 0.421 & -0.710 & 0.161 \\
Transformer (Adam) & 0.124 & 0.296 & 0.418 & -0.716 & 0.161 \\
\bottomrule\end{tabular}\end{table}

\subsection{Summary results}
Across neural predictors, gross Sharpe ratios range from 0.883 to 0.959 in Q1 and from 0.411 to 0.429 in Q5. Corresponding annualized volatilities are 12.92--13.00\% and 29.54--29.90\%. These are descriptive comparisons over the observed period, not statistical equivalence claims about portfolio returns.

Turnover varies more substantially: the maximum-to-minimum ratio among neural predictors is 1.41 in Q1 and 1.58 in Q5. CNN--SGD and LSTM--SGD trade less than their Adam and Muon counterparts, but MLP--Muon trades less than MLP--SGD, and Transformer turnover is relatively similar across optimizers. Thus the economic consequence is configuration-dependent decision multiplicity, rather than a uniform SGD advantage.

\begin{figure}[htbp]\centering
\includegraphics[width=.8\linewidth]{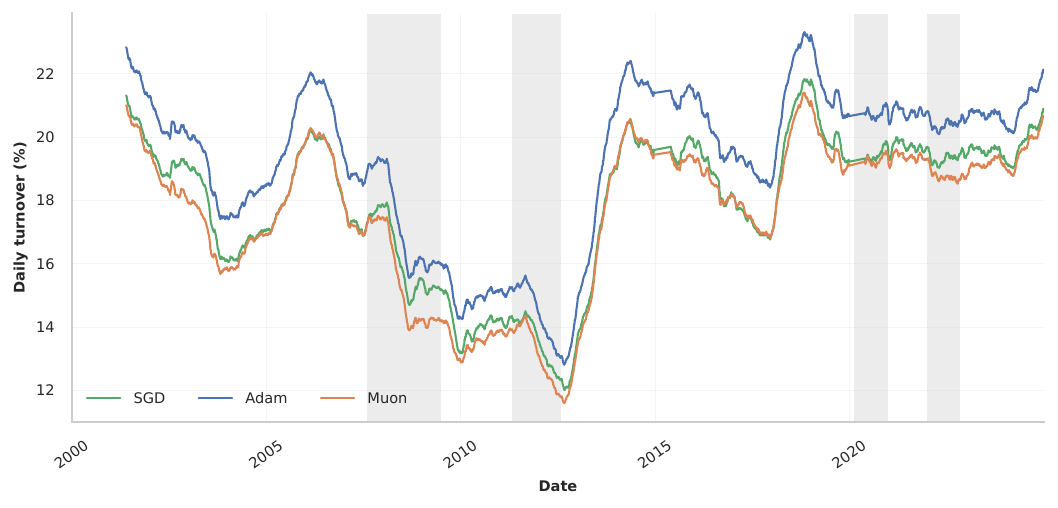}
\caption{\textbf{Rolling target-weight turnover, Q1.} One-year rolling averages over the available 2000--2024 history, with neural architectures equally represented within each optimizer average. The average does not imply the same ordering in every architecture.}
\label{fig:turnover_avg_q1}
\end{figure}

\subsection{Time-series evidence: turnover dynamics}
\begin{figure}[htbp]\centering
\includegraphics[width=\linewidth]{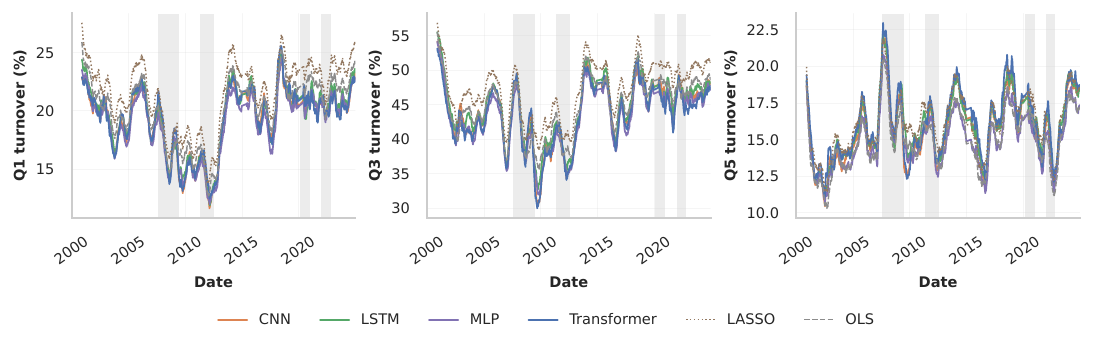}
\caption{\textbf{Rolling turnover by volatility quintile (Adam).} Six-month rolling target-weight turnover over the available 2000--2024 history for Q1 (low volatility), Q3 and Q5 (high volatility), from left to right. Vertical scales differ across panels. Grey bands mark the market-stress windows defined above.}\label{fig:turnover_panels_adam}\end{figure}

\begin{figure}[htbp]\centering
\includegraphics[width=\linewidth]{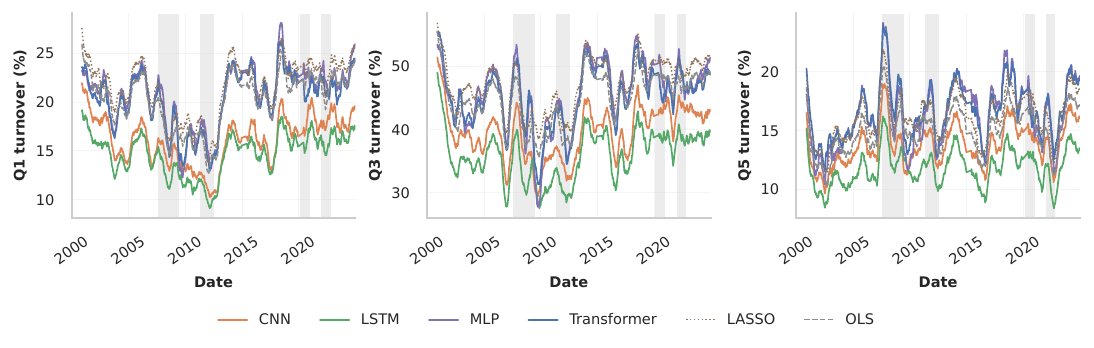}
\caption{\textbf{Rolling turnover by volatility quintile (SGD).} Six-month rolling target-weight turnover over the available 2000--2024 history for Q1 (low volatility), Q3 and Q5 (high volatility), from left to right. Vertical scales differ across panels. Grey bands mark the market-stress windows defined above.}\label{fig:turnover_panels_sgd}\end{figure}

\begin{figure}[htbp]\centering
\includegraphics[width=\linewidth]{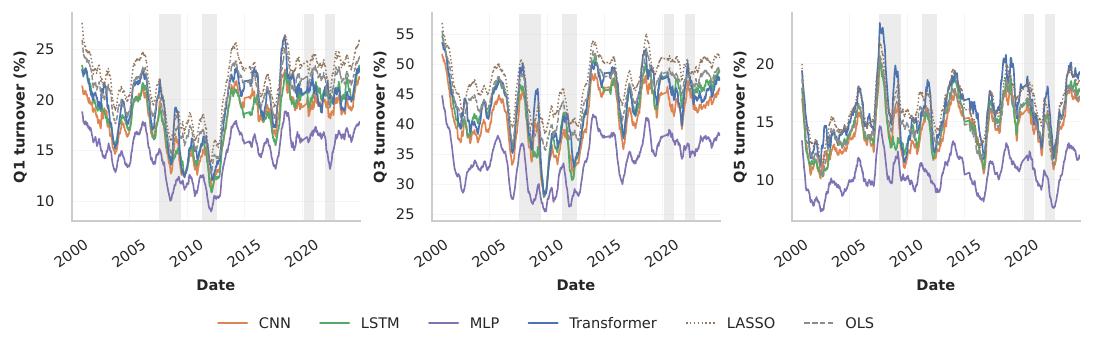}
\caption{\textbf{Rolling turnover by volatility quintile (Muon).} Six-month rolling target-weight turnover over the available 2000--2024 history for Q1 (low volatility), Q3 and Q5 (high volatility), from left to right. Vertical scales differ across panels. Grey bands mark the market-stress windows defined above.}\label{fig:turnover_panels_muon}\end{figure}

Figure~\ref{fig:turnover_avg_q1} summarizes optimizer averages; Figures~\ref{fig:turnover_panels_adam}--\ref{fig:turnover_panels_muon} retain the architecture-specific paths for Q1, Q3 and Q5. Rolling averages describe variation over time and should be read alongside the complete-period table. The disaggregated plots are essential because optimizer averaging hides the different MLP and CNN/LSTM orderings. No universal ordering or amplification during every stress period is implied. Grey bands mark the global financial crisis (August 2007--June 2009), European debt crisis (May 2011--July 2012), COVID period (15 February--December 2020), and inflation shock (January--October 2022). These are contextual stress windows, not formal recession dates.  Each quintile has its own labeled vertical scale.

Together, these results show that similar forecasting errors and gross portfolio performance can coexist with materially different allocation changes. The financial experiment supplies a downstream illustration of the learned functions' differences, rather than a test of which optimizer is universally preferable.


\begin{table}[htbp]
\centering\small
\caption{\textbf{Held-out portfolio comparison, May 2020--December 2024.} The same representatives, restricted to test observations. Turnover is a daily fraction; Sharpe is gross.}
\label{tab:vol_sorted_test_check}
\begin{tabular}{lrrrr}
\toprule
& \multicolumn{2}{c}{Q1: Low volatility} & \multicolumn{2}{c}{Q5: High volatility} \\
Model & Turnover & Sharpe & Turnover & Sharpe \\
\midrule
OLS & 0.223 & 1.142 & 0.153 & 0.859 \\
LASSO & 0.237 & 1.149 & 0.166 & 0.843 \\
CNN (SGD) & 0.184 & 1.158 & 0.144 & 0.823 \\
CNN (Muon) & 0.198 & 1.150 & 0.150 & 0.813 \\
CNN (Adam) & 0.210 & 1.122 & 0.161 & 0.853 \\
LSTM (SGD) & 0.164 & 1.116 & 0.121 & 0.851 \\
LSTM (Muon) & 0.209 & 1.182 & 0.157 & 0.841 \\
LSTM (Adam) & 0.212 & 1.152 & 0.163 & 0.834 \\
MLP (SGD) & 0.228 & 1.133 & 0.165 & 0.849 \\
MLP (Muon) & 0.164 & 1.144 & 0.107 & 0.850 \\
MLP (Adam) & 0.208 & 1.110 & 0.154 & 0.836 \\
Transformer (SGD) & 0.218 & 1.136 & 0.168 & 0.829 \\
Transformer (Muon) & 0.208 & 1.105 & 0.164 & 0.840 \\
Transformer (Adam) & 0.208 & 1.150 & 0.164 & 0.835 \\
\bottomrule\end{tabular}\end{table}
\FloatBarrier

\section{Further Related Work}
\label{app:further_work}

Our work relates to several strands of literature on underspecification, optimization-induced inductive bias, interpretability, and empirical benchmarking in financial machine learning. While each of these literatures documents important facets of modern learning systems, they are rarely studied jointly in low signal-to-noise time-series environments.

\subsection{Underspecification and Predictive Multiplicity}
When the available data are only \emph{weakly informative} about the target relationship, empirical risk minimization is best viewed as searching a landscape with a \emph{broad, shallow basin}: many parameter settings sit on (nearly) the same ``loss contour,'' so the training objective does not single out a canonical predictor. This phenomenon has been formalized as \emph{underspecification} \citep{damour2020underspecification}: an ML \emph{pipeline} (data, preprocessing, hypothesis class, optimizer, validation protocol) is underspecified when it can output many \emph{distinct} predictors that are indistinguishable under in-distribution validation yet behave differently in deployment. Underspecification is closely related to Breiman's \emph{Rashomon effect} \citep{breiman2001statistical}, named after Kurosawa's film in which multiple coherent narratives coexist without a single definitive truth \citep{kurosawa1950rashomon}. The key methodological point is not merely that multiple models achieve similar error, but that \emph{the data do not refute many qualitatively different functional explanations}.

\paragraph{Rashomon sets as a geometric object.}
Fix a function class $\mathcal{F}$ and loss $\ell$. Writing $\widehat R_n(f)\!=\!\frac{1}{n}\sum_{i=1}^n \ell(f(x_i),y_i)$ and $\widehat R_n^\star\!=\!\inf_{f\in\mathcal{F}}\widehat R_n(f)$, the (empirical) Rashomon set at tolerance $\varepsilon\!\ge\!0$ is
\[
\mathcal{R}_\varepsilon(\mathcal{F}) \;=\; \Bigl\{ f\in\mathcal{F} \;:\; \widehat R_n(f)\le \widehat R_n^\star+\varepsilon \Bigr\}.
\]
In high-noise or intrinsically stochastic domains, $\mathcal{R}_\varepsilon(\mathcal{F})$ can be \emph{large} even for small $\varepsilon$: recent results isolate regimes where noisy label generation provably induces large collections of near-optimal models \citep{semenova2023noise}, and perspective work emphasizes that such multiplicity is common in high-stakes tabular settings (e.g., credit decisions, healthcare, criminal justice), precisely because the outcome mechanism is nondeterministic \citep{rudin2024}. Moreover, a large Rashomon set is empirically and theoretically linked to the existence of \emph{simple-yet-accurate} predictors \citep{semenova2022rashomon}, contributing to the broader evidence that, on many tabular problems, high accuracy does not require black-box complexity \citep{holte1993very,lou2013accurate,angelino2018corels,mctavish2022fast,liu2022fastsparse,mcelfresh2023when}.

\paragraph{Near-optimal risk need not mean functional equivalence.}
A central complication is that proximity in scalar risk does not control pointwise agreement. Two predictors can both lie in $\mathcal{R}_\varepsilon(\mathcal{F})$ and yet diverge sharply on individual instances, which is precisely the operational content of \emph{predictive multiplicity}. In classification, one natural disagreement functional is
\[
\mathrm{Dis}(f,g) \;=\; \mathbb{P}_{X\sim \mathcal{D}_X}\!\bigl[f(X)\neq g(X)\bigr],
\qquad f,g\in \mathcal{R}_\varepsilon(\mathcal{F}),
\]
and analogues for regression can track sign flips or threshold exceedances (e.g., $\mathbb{P}[\,|f(X)-g(X)|>\tau\,]$), which are often more decision-relevant than MSE. Predictive multiplicity has been explicitly quantified in classification \citep{marx2020predictivemultiplicityclassification} and broadened to procedural/justificatory concerns in policy-facing pipelines \citep{black2022multiplicity}. Importantly, the underlying observation that ``many models fit about equally well'' appears in older statistical and econometric discussions of model specification and competing adequate descriptions \citep{mountain1989combined,mccullagh1989glm}; what is comparatively less emphasized there is the downstream implication Breiman highlighted: if multiple distinct predictors induce different narratives (feature attributions, mechanisms, causal stories), then \emph{any single-model explanation is, without further structure, underdetermined by the data} \citep{breiman2001statistical,marx2020predictivemultiplicityclassification}.

\paragraph{Why we need behavioral evaluation beyond a single score.}
These observations motivate evaluation protocols that treat a benchmark metric as \emph{necessary but not sufficient}. Rather than interpreting a one-number gap (say, a tiny $\Delta$ in validation loss) as evidence of substantive superiority, one should ask how much \emph{behavioral latitude} remains within $\mathcal{R}_\varepsilon(\mathcal{F})$:
\begin{itemize}\setlength{\itemsep}{1pt}\setlength{\parskip}{0pt}\setlength{\parsep}{0pt}
  \item \emph{Stability of decisions:} how large can $\mathrm{Dis}(f,g)$ be among near-optimal $f,g$? \citep{marx2020predictivemultiplicityclassification,black2022multiplicity}
  \item \emph{Stability of boundary geometry:} how much can decision regions/threshold crossings move while preserving $\widehat R_n$? \citep{rudin2024}
  \item \emph{Constraint feasibility:} does the Rashomon set contain models satisfying domain constraints (monotonicity, fairness, sparsity) \emph{without} sacrificing accuracy? \citep{rudin2024}
\end{itemize}
Recent algorithms now make this viewpoint operational by explicitly representing or approximating Rashomon sets for nontrivial classes---including sparse decision trees, sparse GAMs, and risk scores---so that one can \emph{query} the set for models that satisfy user preferences and examine variable importance \emph{across all good models}, rather than conditioning on a single trained predictor \citep{xin2022rashomon,zhong2023goodgams,liu2022fasterrisk,zhu2023mortality,rudin2024}.

\paragraph{Implications for financial forecasting.}
Financial prediction problems are prototypical ``noisy-label'' regimes: outcomes are affected by latent factors and stochastic shocks, so it is natural to expect large Rashomon sets even when the modeling class is heavily restricted \citep{rudin2024,semenova2023noise}. Yet evaluation in finance often reduces model selection to a single scalar score (e.g., a loss, a correlation, or a portfolio-level summary), effectively projecting a high-dimensional behavioral object $\mathcal{R}_\varepsilon(\mathcal{F})$ onto one axis. The econometrics literature often responds to model uncertainty by combining forecasts \citep{RePEc:eee:ecofch:1-04}; in contrast, we focus on a different question: when we \emph{must} select a single predictor, which \emph{inductive biases} (regularization, architecture, optimization, data processing choices) decide \emph{where} we land within $\mathcal{R}_\varepsilon(\mathcal{F})$, and how that choice governs the stability of instance-level predictions, decision thresholds, and downstream actions under underspecification.

\subsection{Implicit Bias of Optimization Trajectories}
\label{app:implicit-bias-optimization}

\paragraph{Changing optimizer or hyperparameters: same objective, different solutions.}
A substantial literature documents that changing the optimizer (or its noise scale) can move learning toward different regions of parameter space with different curvature, geometry, and generalization behavior. 
Empirically, SGD-like methods are often associated with flatter solutions and better test performance than adaptive methods in high-signal domains \citep{wilson2017marginal, keskar2017largebatch}, while adaptive optimizers can exhibit distinct ``effective regularization'' behavior unless properly regularized (notably via decoupled weight decay) \citep{loschilov2019decoupled, zou2021adam}. 
These observations connect to a longer thread linking curvature/sharpness to generalization \citep{hochreiter1997flat, keskar2017largebatch, dinh2017sharpminima, jastrcatastrophic2021, foret2021sam}, though the sharpness--generalization relationship is subtle and depends on parametrization and invariances \citep{dinh2017sharpminima}. 
From the present paper's perspective, the key point is methodological: much of the optimizer-comparison literature focuses on regimes where optimizer choice changes \emph{test error}. In contrast, in low-signal settings (the regime of our experiments), optimizer choice can be consequential even when test losses are statistically indistinguishable, because the optimizer determines \emph{which} near-optimal function is selected.

\paragraph{SGD \textit{vs} Adam.}
A useful mental model is that the \emph{loss} draws the landscape, but the \emph{optimizer} chooses the
vector field that actually drives the dynamics; in that sense, optimizing the same $L$ with different
methods can amount to (approximately) following gradient flow on different ``effective'' functionals.
\begin{itemize}[leftmargin=1em,itemsep=0.6em,topsep=0.15em,parsep=0pt]
\item \textbf{Empirical signatures of the gap.}
In language modeling, heavy-tailed (Zipf) class frequencies create a regime where GD/SGD makes
little progress on rare classes, while Adam and sign-based methods remain comparatively insensitive
\citep{kunstner2024heavy}. In transformers, the Hessian spectra vary dramatically across
parameter blocks (``block heterogeneity''), so a single global learning rate makes SGD ill-suited,
whereas Adam's coordinate-wise scaling mitigates this heterogeneity \citep{zhang2024transformers}.
More broadly, adaptive methods can select qualitatively different solutions (and generalize
differently) than SGD on problems with many minimizers \citep{wilson2017marginal}.
\item \textbf{Modified-loss viewpoint for SGD.}
For small but finite step size, the mean SGD iterate (under random shuffling) stays close to gradient
flow on a \emph{modified loss}---the original loss plus an implicit regularizer penalizing minibatch-gradient
norms---with correction scale proportional to $\eta/B$ \citep{smith2021origin}.
In the more realistic without-replacement regime, SGD is locally equivalent to an \emph{additional}
step on a novel regularizer, which effectively regularizes the trace of the gradient-noise covariance
(and often a weighted Fisher) along flat directions \citep{beneventano2023trajectories}.
Related symmetry-based analyses show that minibatch noise can select ``noise-balanced'' solutions
and shape the stationary distribution of the stochastic gradient flow \citep{ziyin2025noise}.
At a high level, these results can be summarized as a first-order perturbation
\[
\tilde L_{\mathrm{SGD}}(\theta)=L(\theta)+\frac{\eta}{B}\,R(\theta)+o(\eta/B),
\qquad
\dot\theta \approx -\nabla \tilde L_{\mathrm{SGD}}(\theta).
\]
\item \textbf{Adam differs already at leading order.}
For Adam/RMSProp, backward-error analysis yields hyperparameter-dependent drift terms that can
penalize a (perturbed) $\ell_1$-type norm of $\nabla L$ or even \emph{impede} its reduction (a typical
regime), rather than inducing the same $\ell_2$-type regularization as GD \citep{cattaneo2023implicit,cattaneo2025memory,cattaneo2026effect}.
Moreover, taking $\eta\to0$ while keeping momentum parameters fixed, Adam approaches a sign-gradient flow,
indicating that the limiting vector field can differ from $-\nabla L$ already at \emph{zeroth} order in $\eta$
\citep{ma2022qualitative}. This discrepancy is reflected in convergence theory: Adam can track an ``Adam
vector field'' whose zeros generally differ from critical points of $L$ \citep{dereich2024convergence},
and classic convex counterexamples show that Adam may even fail to converge to the minimizer \citep{reddi2019convergence}.
\end{itemize}

\paragraph{Classical implicit bias: minimum-norm and maximum-margin principles.}
Beyond comparisons between optimizers, a complementary line of work asks \emph{which} predictor is selected by a given optimization trajectory when the empirical objective admits many global minimizers.
For linearly separable classification with exponential-tail losses (e.g., logistic / cross-entropy), gradient descent can drive $\|w_t\|\to\infty$ while the \emph{direction} $w_t/\|w_t\|$ converges to a maximum-margin separator, thereby making the ``implicit regularizer'' explicit \citep{soudry2018implicitbias,ji2019nonseparable,lyu2020margin,gunasekar2018optgeometry}.
Concretely, in the linear case the limiting direction aligns with a solution of the hard-margin SVM problem,
\[
\hat w \in \arg\min_{w}\ \|w\|_2^2
\quad\text{s.t.}\quad
y_i\langle w,x_i\rangle \ge 1\ \ \forall i,
\]
even though the training objective contains no explicit norm penalty \citep{soudry2018implicitbias,ji2019nonseparable}.
In deep \emph{homogeneous} models trained with exponential-tail losses, analogous margin-maximizing behavior emerges (in appropriate normalizations/limits), with the normalized margin increasing and converging to KKT points of a natural max-margin problem \citep{lyu2020margin}.
A key nuance is that the induced notion of ``simplicity'' depends not only on the loss but also on the \emph{parameterization} and the \emph{optimization geometry}: reparameterizations realizing the same predictor class can yield different implicit regularizers \citep{gunasekar2018optgeometry}.
For instance, linear convolutional parameterizations bias toward sparsity in the frequency domain via an $\ell_{2/L}$-type penalty, in contrast to fully connected linear parameterizations whose bias is $\ell_2$ max-margin \citep{gunasekar2018conv}; and factorized parameterizations in underdetermined quadratic objectives can bias toward low-complexity (e.g., low nuclear-norm) solutions \citep{gunasekar2017mf}.

In regression and ``lazy'' (linearized) regimes, the same selection phenomenon appears as \emph{minimum-norm interpolation} in an induced function space.
In the infinite-width limit, training dynamics of wide networks are governed by kernel gradient descent with the neural tangent kernel (NTK) \citep{jacot2018ntk,lee2019wide,chizat2019lazy}, so that (for square loss, and in the interpolating setting) gradient-based training selects the minimum-RKHS-norm interpolant:
\[
f_\star \in \arg\min_{f\in\mathcal H_{\mathrm{NTK}}}\ \|f\|_{\mathcal H_{\mathrm{NTK}}}
\quad\text{s.t.}\quad
f(x_i)=y_i\ \ \forall i.
\]
Taken together, these works support a unifying picture: when many interpolating solutions exist, the optimizer--parameterization pair often induces a geometry in which training selects a canonical low-complexity solution (max-margin for separable classification; minimum-norm in an induced hypothesis space for regression), despite the absence of any explicit complexity penalty \citep{gunasekar2018optgeometry,jacot2018ntk,lee2019wide}.

\paragraph{Biases along the trajectory: learning order, signal selection, and terminal geometry.}
Implicit bias is not only a statement about the final predictor, but also about \emph{which components are learned first} and \emph{which signals dominate} under the training dynamics.
A prominent example is \emph{spectral bias}: neural networks tend to fit low-frequency components before high-frequency components, shaping intermediate functions along training and sometimes constraining what is practically learned under finite-time optimization \citep{rahaman2019spectralbias}.
(Complementarily, NTK-based analyses make explicit how learning rates depend on the kernel spectrum, which can be interpreted as a frequency- or eigendirection-dependent ordering of fit \citep{jacot2018ntk,lee2019wide}.)
At the representation level, late-phase training can exhibit structured terminal geometry, including within-class collapse and simplex-like class arrangements (``neural collapse'') \citep{papyan2020neuralcollapse}.
Alongside these geometric phenomena, modern evidence emphasizes that standard ERM/SGD pipelines can preferentially exploit ``simple'' or ``shortcut'' features---highly predictive yet brittle signals---depending on the data/optimization geometry \citep{ilyas2019adversarial,geirhos2020shortcut,shah2020simplicitybias,sagawa2020spurious}.
This line also connects to methods designed to \emph{avoid} such shortcut reliance by enforcing invariances across environments (e.g., invariant risk minimization) \citep{arjovsky2019irm}.
Most recently, theory suggests SGD-like dynamics can even \emph{recover support} by suppressing irrelevant features, yielding sparse \emph{effective} solutions without explicit sparsity regularization \citep{beneventano2024neuralnetworkslearnsupport}.
Overall, these results reinforce that optimizer-induced preferences can manifest as differences in (i) \emph{temporal learning order} (e.g., low-to-high frequency), (ii) \emph{feature reliance} (robust \textit{vs}  spurious/shortcut), and (iii) \emph{terminal representation geometry}, even when the scalar training loss is unchanged.

\paragraph{Edge of Stability.}
A growing literature connects optimizer hyperparameters to \emph{curvature-regulated} training dynamics, framing optimization as a trajectory constrained by stability considerations rather than a purely “loss-minimizing” procedure. In full-batch (or sufficiently large-batch) training, this viewpoint typically decomposes learning into two regimes: an early phase of \emph{progressive sharpening}—where the dominant curvature scale increases rapidly—and a later oscillatory phase where training hovers near the boundary of instability, i.e., the \emph{Edge of Stability} (EoS).

Concretely, prior work documents that the leading Hessian eigenvalue $\lambda_{\max}$ often grows throughout early training (sometimes after a brief initial decrease), reflecting that the trajectory moves into progressively sharper regions of the landscape \citep{jastrzebski_relation_2019,jastrzebski_break-even_2020,cohen2022gradientdescentneuralnetworks}. This growth does not continue indefinitely: \citet{jastrzebski_break-even_2020} highlight a relatively abrupt \emph{phase transition} (“break-even”) that marks the end of the progressive sharpening regime. Importantly, this transition is typically attributed to \emph{algorithmic} stability limits (i.e., the update rule becoming marginally unstable), rather than a static property of the loss surface. Consistent with that interpretation, the location of the transition depends on the optimization method and its hyperparameters even when the underlying task and model are held fixed \citep{jastrzebski_relation_2019,jastrzebski_break-even_2020,cohen2022gradientdescentneuralnetworks}. In the EoS regime itself, full-batch GD exhibits the canonical signature $\lambda_{\max}\approx 2/\eta$ (the quadratic stability threshold), with $\lambda_{\max}$ fluctuating around that boundary during training \citep{cohen2022gradientdescentneuralnetworks}. Empirically, a substantial fraction of optimization—especially under MSE-like objectives—often occurs in this marginally stable regime, which can strongly influence the curvature (and hence geometry) of the final solution \citep{cohen2022gradientdescentneuralnetworks}. Subsequent analyses explain why $\lambda_{\max}$ can slightly exceed $2/\eta$ in practice: higher-order nonlinearities shift the effective stability condition away from the quadratic approximation \citep{chen_beyond_2023}. Mechanistically, \citet{damian_self-stabilization_2023} propose a “self-stabilization” picture for GD at the edge, where third-order effects can act as a stabilizing force under empirically observed alignment conditions; however, extending this explanation cleanly to mini-batch training remains nontrivial.

\paragraph{Adaptive and Stochastic Edge of Stability.}
Recent work extends the “edge” framework to adaptive optimizers. \citet{cohen_adaptive_2022} show that full-batch Adam does not generally sit at an edge defined by the \emph{raw} Hessian. Instead, it operates near an \emph{Adaptive Edge of Stability} (AEoS): the stability boundary is governed by the spectrum of a \emph{preconditioned} curvature matrix (e.g., the maximum eigenvalue of $P^{-1}H$, where $P$ is Adam’s second-moment preconditioner). In this view, the relevant instability threshold scales like $\lambda_{\max}(P^{-1}H)\approx c/\eta$ (with $c$ depending on Adam’s hyperparameters, e.g., $\beta_1$), while the unpreconditioned Hessian can continue to grow because the preconditioner dynamically rescales directions of high curvature. This produces a qualitative distinction from GD/SGD: non-adaptive methods are “pushed away” from high-curvature regions by step-size instability, whereas adaptive methods can traverse into higher raw-curvature regions while maintaining stability via preconditioning.

Finally, bringing the edge narrative to \emph{mini-batch} SGD is delicate because oscillations and noise are not, by themselves, certificates of instability. \citet{andreyev2025edgestochasticstabilityrevisiting} propose the \emph{Edge of Stochastic Stability} (EoSS), which reframes the edge in terms of computable \emph{instability certificates} for the local quadratic model under stochastic gradients. A key empirical and conceptual contribution is to replace full-batch $\lambda_{\max}$ with an SGD-relevant curvature statistic—\emph{Batch Sharpness}—that captures direction-aware curvature at the mini-batch level. Empirically, Batch Sharpness sharpens and then saturates near a stability boundary proportional to $2/\eta$, and threshold crossings coincide with catapult-like excursions and loss spikes under targeted $(\eta,b)$ interventions. We use this line of work to motivate the matched-checkpoint learning-rate and batch-size interventions. Their raw-curvature responses support a stability-constraint interpretation in identified configurations; the measured statistic and stabilizing feedback need not coincide with the canonical EoSS criterion.

\paragraph{What is (still) poorly understood in tied-loss regimes.}
Most implicit-bias results are derived in regimes where either (i) the data are separable/high-signal and the optimizer selects a margin- or norm-optimal classifier \citep{soudry2018implicitbias, lyu2020margin}, or (ii) optimization choices drive measurable test-error differences \citep{wilson2017marginal, keskar2017largebatch}. 
By contrast, the regime emphasized in this paper---\emph{predictive equivalence with functional divergence}---asks for a different kind of characterization: not ``which optimizer generalizes better,'' but ``which function is selected when generalization is a tie''. 
Our empirical findings can be viewed as evidence that, in low-signal time series, optimizer-dependent stability constraints and update geometry instantiate a consequential implicit prior over admissible predictors, making the choice of optimizer an integral part of the modeling assumption rather than a tool to achieve better performance.

\subsection{Interpretability and SHAP}

\paragraph{Interpretability and signal selection.}
A parallel literature shows that identical test error can conceal qualitatively different feature reliance and decision logic. Explanatory or interpretability multiplicity arises when near-optimal predictors admit conflicting post-hoc explanations for the same inputs \citep{brunet2022indeterminacy}. Related work in vision demonstrates that models trained with standard empirical risk minimization may exploit shortcut or non-robust features that are predictive but semantically misaligned \citep{ilyas2019adversarial}. Closest to our focus, studies of optimization bias show that learning dynamics can prioritize particular signals even when richer alternatives exist within the hypothesis space \citep{shah2020simplicitybias, lampinen2024featurebias}. These findings suggest that optimization pathways shape \emph{which} signal is learned, not merely how well it fits the data. In the financial domain, \citet{Lo03062023} similarly emphasize that high predictive performance is insufficient; understanding the internal mechanics and ``interpretability'' of deep models is a prerequisite for their deployment in risk premia forecasting.

\subsection{Neural Networks in Finance}

\paragraph{Benchmarking in financial machine learning.}
Benchmark studies in finance report mixed evidence: deep models can help with rich covariates or cross-series pooling, but well-specified linear baselines remain competitive in low-SNR settings \citep{gu2020empirical, chen2024deep, VORTELINOS2017824, BRANCO2024101524}. To demonstrate gains, many studies introduce additional features, alternative targets, or increasingly complex architectures \citep{heaton2017deep, lim2021time, 10.1093/jjfinec/nbac020}. While valuable, this focus leaves open a more fundamental question: when performance ties persist, what distinguishes the learned predictors themselves? Our work departs from the benchmark race by holding data and objectives fixed and instead interrogating the functional consequences of training choices.

\paragraph{Volatility forecast with classical ML.}
Volatility forecasting has a long history in econometrics. Classical models like ARCH/GARCH processes \citep{engle1982autoregressive, BOLLERSLEV1986307} and their extensions treat volatility as a latent process and have achieved considerable success. The availability of high-frequency data has further refined these targets, allowing for non-parametric ex-post measurement of volatility via realized variance and bipower variation \citep{10.1093/jjfinec/nbh001}, which now serve as standard ground-truth proxies. The Heterogeneous Autoregressive (HAR) model \citep{10.1093/jjfinec/nbp001} further improved daily volatility forecasts by incorporating multi-horizon averages (e.g., daily, weekly, monthly volatility) in a simple linear framework. These parsimonious models often perform robustly across datasets and remain strong benchmarks \citep{VORTELINOS2017824, BRANCO2024101524}.

With the rise of data mining, researchers began exploring machine learning (ML) algorithms for volatility prediction. Tree-based ensembles and kernel methods (e.g., random forests, boosting, SVMs) have been applied to capture nonlinear patterns in volatility dynamics. However, early studies found that when using only past returns or volatilities as features, complex ML models often provided little to no improvement over well-specified linear models \citep{BRANCO2024101524, VORTELINOS2017824, SoutoMoradi2023_FT_LSTM_RV, MaHongSong2024_RRMIDAS_CNNLSTM}. For example, \citet{BRANCO2024101524} find that no ML method consistently beats a HAR model when forecasting realized volatility out-of-sample, especially at longer horizons. These results echo other comparisons showing that traditional models can remain competitive in feature-constrained settings \citep{VORTELINOS2017824, BRANCO2024101524}. In practice, the performance gap narrows without additional data: adding macro-financial features or high-frequency inputs is often helpful for ML models to outperform simple baselines. This underscores that in low signal-to-noise regimes, many models achieve similar predictive accuracy, setting the stage for potential underspecification.

\paragraph{Deep Neural Networks for Volatility Forecasting.}

The application of deep learning to volatility forecasting has attracted significant attention in recent years \citep{Leushuis2026}. Early neural network approaches, including multilayer perceptrons and recurrent networks, were explored as alternatives to GARCH. For instance, \citet{DONALDSON199717} proposed one of the first ANN-GARCH hybrid models and found that a neural network could approximate the volatility process similarly to GARCH. More recently, \citet{DiGiorgi2025} formalize this connection by showing that deep recurrent neural networks can replicate the recursive structure of GARCH-type models while retaining the flexibility of data-driven learning. Deep architectures have also been deployed at scale \citep[e.g.,][]{bucci2020}: LSTMs, in particular, have been a popular choice due to their ability to capture long-term dependencies in financial time series \citep{hochreiter1997long}. An empirical study by \citet{SoutoMoradi2023_FT_LSTM_RV} reports that LSTM-based models can outperform HAR in some settings;
improvements vary by horizon and features.  Recent work has attempted to extend these architectures by incorporating attention mechanisms or hybridizing them with classical error-correction terms \citep{Leushuis2026, KumarThenmozhi2021_CNN_LSTM_RV}. 

Beyond RNNs, researchers have explored CNNs and attention-based models \citep{reisenhofer2022harnetconvolutionalneuralnetwork, BetzHeilPeter2023_CNNVolatilityImages}. Temporal convolutional networks and Transformers have been evaluated on volatility prediction tasks, sometimes as components of larger frameworks. For instance, \citet{zhang2023volatilityforecastingmachinelearning} leverage cross-asset intraday data and show that a simple feed-forward network pooling many stocks’ volatilities outperforms linear and tree models, attributing the gain to the network’s ability to capture latent “commonality” in volatility. Moreover, it is demonstrated that, given rich feature sets or additional signals (like macro variables, order flow), deep networks can significantly outperform traditional models \citep{b5bfaa97d7b6466ead4246df4ca0afdc, zhang2023volatilityforecastingmachinelearning}. On the other hand, in purely univariate settings with only past realizations as inputs, even sophisticated networks often end up ``tied'' with linear models on standard error metrics \citep{gu2020empirical, chen2024deep, bucci2020, BetzHeilPeter2023_CNNVolatilityImages}. This phenomenon is evident in many recent financial ML benchmarks, where increasing model complexity yields diminishing returns \citep{hu2025fintsbcomprehensivepracticalbenchmark}. As a result, many studies report that volatility forecasting is a tough low-signal task: simple and complex models often achieve statistically indistinguishable accuracy. This backdrop of frequent leaderboard ties motivates going beyond aggregate metrics to examine what models are actually learning.

\paragraph{Foundation Models and Recent Advances in Time-Series Forecasting.}

Inspired by the success of large-scale models in NLP and vision, researchers have begun developing foundation models for time series \citep{bommasani2022opportunitiesrisksfoundationmodels}. These are extensive pre-trained models (often Transformer-based) that aim to capture universal time-series patterns and enable zero-shot or few-shot forecasting across diverse tasks. For example, Google’s TimesFM is a pre-trained Transformer that demonstrated accurate zero-shot forecasts across many time series without task-specific training \citep{das2024decoderonlyfoundationmodeltimeseries}. Similarly, Salesforce’s MOIRAI and other large time-series models have been released, alongside open-source initiatives like MOMENT \citep{woo2024unifiedtraininguniversaltime, goswami2024momentfamilyopentimeseries}. These models leverage massive datasets and novel architectures to approach time-series forecasting as a generalizable transfer-learning problem \citep{zhou2023tsf}.
Despite their promise, current foundation time-series models still face the classic pitfalls seen in earlier studies. A recent survey by \citet{Liang_2024} highlights that carefully tuned lightweight models can rival or even beat foundation models on many benchmarks, especially when the latter are not fine-tuned. Indeed, a NeurIPS 2025 workshop \citep{berts_workshop_2025} noted that time-series foundation models have yet to achieve a definitive “BERT moment”---in other words, they do not strictly dominate dedicated models in all cases. In volatility forecasting specifically, even the most advanced deep models (e.g., Transformer encoders or hybrid LSTM-GARCH schemes) often yield performance on par with simpler methods \citep{DAMATO2022127158, VORTELINOS2017824, 10.1093/jjfinec/nbac020}. Moreover, these large models typically rely on adaptive optimizers (e.g., Adam) during training, treating the optimizer as a default setting rather than a design choice. Our work raises the point that even in such state-of-the-art systems, if multiple training runs or optimization strategies achieve similar loss, they could embed different inductive biases. As foundation models push predictive performance limits, understanding the role of the training procedure becomes crucial to ensure consistency and reliability of the learned patterns across deployments.

\paragraph{Positioning of Our Contribution.}
In light of the above, our work sits at the intersection of these threads. We focus on a domain---daily equity volatility forecasting---where deep learning models have struggled to decisively outperform parsimonious econometric models \citep{BRANCO2024101524, VORTELINOS2017824, 10.1093/jjfinec/nbac020}. Rather than introduce another architecture or exogenous feature, we interrogate the underspecification that arises in this low signal-to-noise regime \citep{damour2020underspecification}. Prior literature has observed that many ML predictors are observationally equivalent in such settings, typically choosing between them based on minor differences in validation loss or aesthetic preferences. We extend this discussion by showing that when several models tie in predictive accuracy, the choice of optimizer can be a decisive factor in which functional form is learned. While others have noted the implicit biases of optimization in high-impact domains \citep{wilson2017marginal, zou2021adam}, we demonstrate this effect in financial time series, where it has been less systematically studied. In doing so, we bridge the gap between benchmarking studies (emphasizing which model yields lower error) and interpretability or decision-focused analyses (emphasizing how the model’s behavior matters for users). Our findings suggest that financial ML researchers should evaluate models not just by Sharpe ratios or MSE, but also by the stability and economic plausibility of the patterns they capture. By revealing that different architecture–optimizer pairs can encode substantively different volatility dynamics despite equal performance, we provide a new perspective on model selection: when performance is tied, selecting the model with the right inductive bias (e.g., smoother responses, longer memory, lower turnover) becomes paramount for downstream applications. This nuanced approach to evaluation complements existing work on interpretable ML \citep{rudin2022interpretable}
in finance and calls for broader adoption of diagnostics that account for functional and decision-level differences, not just error metrics.

\newpage
\end{document}